\documentclass[]{article}
\usepackage[utf8]{inputenc} 
\usepackage[T1]{fontenc}    
\usepackage{url}
\usepackage{amsfonts}
\usepackage{nicefrac}
\usepackage{microtype}
\usepackage{amsmath}
\usepackage{mathtools}
\usepackage[english]{babel}
\usepackage{subcaption}
\usepackage{ragged2e}
\usepackage{caption}
\usepackage{booktabs}
\usepackage{tikz}

\usepackage{etoolbox}
\usepackage{xspace}
\usepackage{xpatch}
\usepackage{xstring}
\usepackage{enumerate}
\usepackage[square,numbers]{natbib}
\usepackage{setspace}
\usepackage{tabularx}
\usepackage{wrapfig}
\usepackage[useregional=numeric]{datetime2}
\usepackage{tikzscale}
\usepackage{afterpage}
\usepackage{multicol}
\usepackage{bm}
\usepackage{multirow}
\usepackage{float}
\usepackage{xcolor}
\usepackage{algorithm}
\usepackage{algorithmic}
\usepackage{pifont}
\usepackage{caption}
\usepackage{lipsum}
\usepackage{pdfpages}

\usetikzlibrary{positioning,arrows.meta,fit,calc,backgrounds,decorations.pathmorphing,bayesnet}

\definecolor{mydarkblue}{rgb}{0,0.08,0.45}
\definecolor{myblue}{RGB}{103,169,207}
\definecolor{myred}{RGB}{239,138,98}

\usepackage{hyperref}
\hypersetup{%
colorlinks=true,
linkcolor=mydarkblue,
citecolor=mydarkblue,
filecolor=mydarkblue,
urlcolor=mydarkblue}

\DeclareRobustCommand{\note}[1]{}

\DeclarePairedDelimiterX{\SquareBrackets}[1]{[}{]}{#1}
\DeclarePairedDelimiterX{\RoundBrackets}[1]{(}{)}{#1}
\DeclarePairedDelimiterX{\DivergenceBrackets}[2]{[}{]}{#1\;\delimsize\|\;#2}

\NewDocumentCommand{\pr}{ O{p} r() }{
  \def\prArg{#2}\patchcmd{\prArg}{|}{\mid}{}{}#1\RoundBrackets{\prArg}}
\NewDocumentCommand{\p}{ r() }{\pr[p](#1)}
\NewDocumentCommand{\q}{ r() }{\pr[q](#1)}
\NewDocumentCommand{\prm}{ r() }{\pr[\mathrm{p}](#1)}
\NewDocumentCommand{\Normal}{ r() }{\pr[\operatorname{Normal}](#1)}
\NewDocumentCommand{\Cat}{ r() }{\pr[\operatorname{Cat}](#1)}
\NewDocumentCommand{\Beta}{ r() }{\pr[\operatorname{Beta}](#1)}
\NewDocumentCommand{\Bernoulli}{ r() }{\pr[\operatorname{Bernoulli}](#1)}
\NewDocumentCommand{\Dir}{ r() }{\pr[\operatorname{Dir}](#1)}

\newlength\widthE

\DeclarePairedDelimiterX{\infdivx}[2]{ \big( }{ \big) }{%
  #1\;\delimsize\|\;#2%
}
\newcommand{\infdiv}{D_{KL}\infdivx}

\newtheorem{theorem}{Theorem}

\newtheorem{definition}[theorem]{Definition}

\newlength\myindent
\usepackage[accepted]{icml2021}

\begin{document}


\icmltitlerunning{Efficient Iterative Amortized Inference for Learning Symmetric and Disentangled Multi-Object Representations}

\twocolumn[
\icmltitle{Efficient Iterative Amortized Inference for Learning Symmetric and Disentangled Multi-Object Representations}

\icmlsetsymbol{equal}{*}

\begin{icmlauthorlist}
\icmlauthor{Patrick Emami}{uf}
\icmlauthor{Pan He}{uf}
\icmlauthor{Sanjay Ranka}{uf}
\icmlauthor{Anand Rangarajan}{uf}
\end{icmlauthorlist}

\icmlaffiliation{uf}{University of Florida, Gainesville, FL, USA}

\icmlcorrespondingauthor{Patrick Emami}{pemami@ufl.edu}

\icmlkeywords{Keywords placeholder}

\vskip 0.3in
]

\printAffiliationsAndNotice{} 

\begin{abstract}
Unsupervised multi-object representation learning depends on inductive biases to guide the discovery of object-centric representations that generalize.
However, we observe that methods for learning these representations are either impractical due to long training times and large memory consumption or forego key inductive biases. 
In this work, we introduce EfficientMORL, an efficient framework for the unsupervised learning of object-centric representations.
We show that optimization challenges caused by requiring both symmetry and disentanglement can in fact be addressed by high-cost iterative amortized inference by designing the framework to minimize its dependence on it.
We take a two-stage approach to inference: first, a hierarchical variational autoencoder extracts symmetric and disentangled representations through bottom-up inference, and second, a lightweight network refines the representations with top-down feedback.
The number of refinement steps taken during training is reduced following a curriculum, so that at test time with zero steps the model achieves 99.1\% of the refined decomposition performance.
We demonstrate strong object decomposition and disentanglement on the standard multi-object benchmark while achieving nearly an order of magnitude faster training and test time inference over the previous state-of-the-art model.
\end{abstract}

\section{Introduction}
Deep learning has produced impressive results across multiple domains by taking advantage of enormous amounts of data and compute.
However, it has become clear that the \emph{representations} these models learn have fundamental limitations.
Consider the problem of inferring a representation for a multi-object scene.
Most humans can observe a scene and then manipulate the individual objects in their mind---perhaps imagining that a chair has suddenly flipped upside down.    
This example illustrates that the common approach of summarizing an entire scene as a \emph{single} distributed representation~\cite{rumelhart1986parallel} is likely insufficient. 
It has been shown that this approach fails on simple generalization tasks such as processing novel numbers of objects~\cite{eslami2016attend,watters2019cobra,mao2019neuro}.

Alternatively, a scene can be encoded as a \emph{set} of distributed \emph{object-centric} representations using
object detection~\cite{zhou2019objects} or instance segmentation~\cite{he2017mask}.
While sets can handle arbitrary numbers of objects, these methods require ground truth supervision which limits reusability and generalization.
Unsupervised approaches bring the promise of better generalization at the expense of relying on inductive biases to implicitly define the scene representation.
In this paper, we aim to develop such a method that incorporates three key inductive biases that have been argued previously as being essential for object-centric reasoning and addressing the \emph{binding problem} within deep neural networks~\cite{greff2015binding,greff2017neural,pmlr-v97-greff19a,locatello2020object,huang2020better,greff2020binding}.
\begin{itemize}
    \item \textbf{Symmetry} Multiple distributed representations are inferred for a single scene, each sharing a common format with the others~\cite{pmlr-v97-greff19a,locatello2020object}. This eases relational and compositional reasoning, e.g. for learning dynamics models~\cite{van2018relational,veerapaneni2019entity}.
    \item \textbf{Unordered} Any latent representation can take responsibility for any single object~\cite{pmlr-v97-greff19a}. 
    Typically, a randomized iterative process is used to decide the assignment. 
    \item \textbf{Disentangled} Manipulating one dimension of an object representation changes a single object property and leaves all else invariant~\cite{schmidhuber1992learning,higgins2017beta,higgins2018towards}.
\end{itemize}

Prior attempts to incorporate all three inductive biases have been unsuccessful for a variety of reasons. 
Some do not enforce symmetry to avoid solving for the assignment~\cite{burgess2019monet,engelcke2019genesis} while others learn symmetric yet \emph{entangled} representations which circumvents the challenge of disentangling latent factors~\cite{locatello2020object}.
To the best of our knowledge, IODINE~\cite{pmlr-v97-greff19a} is the only model that has all three inductive biases.
However, IODINE has computational concerns due to its use of iterative amortized inference (IAI)~\cite{Marino2018} to implement the assignment of pixels to symmetric representations.
This translates to relatively long training times of over a week given a reasonable compute budget and slow test time inference. 
\begin{figure*}[t]
    \begin{subfigure}[t]{0.42\textwidth}
        \centering
        \includegraphics[scale=0.42]{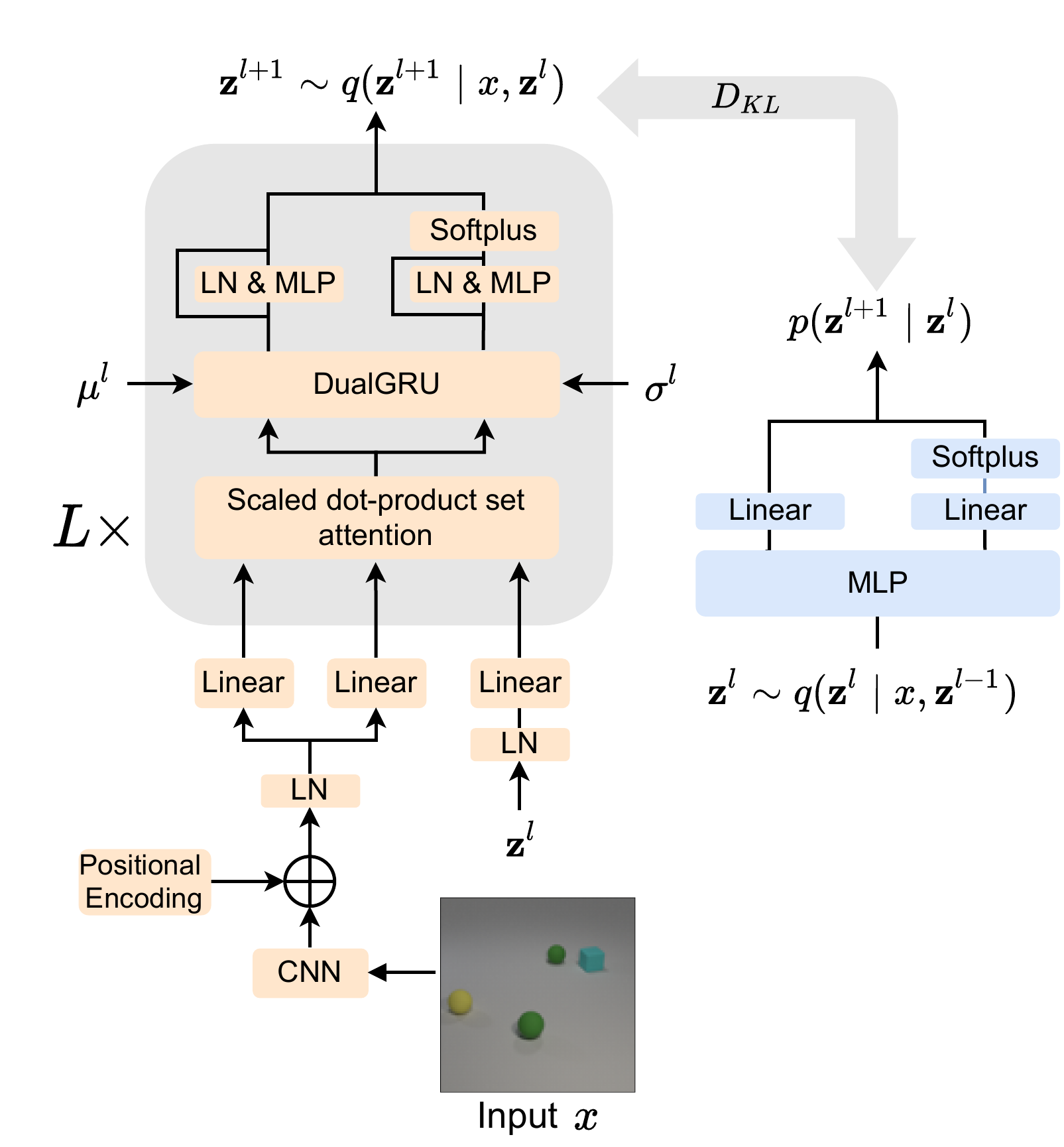}
        \caption{Stage 1\label{fig:mainA}}
    \end{subfigure}%
    \begin{subfigure}[t]{0.32\textwidth}
        \centering
        \includegraphics[scale=0.45]{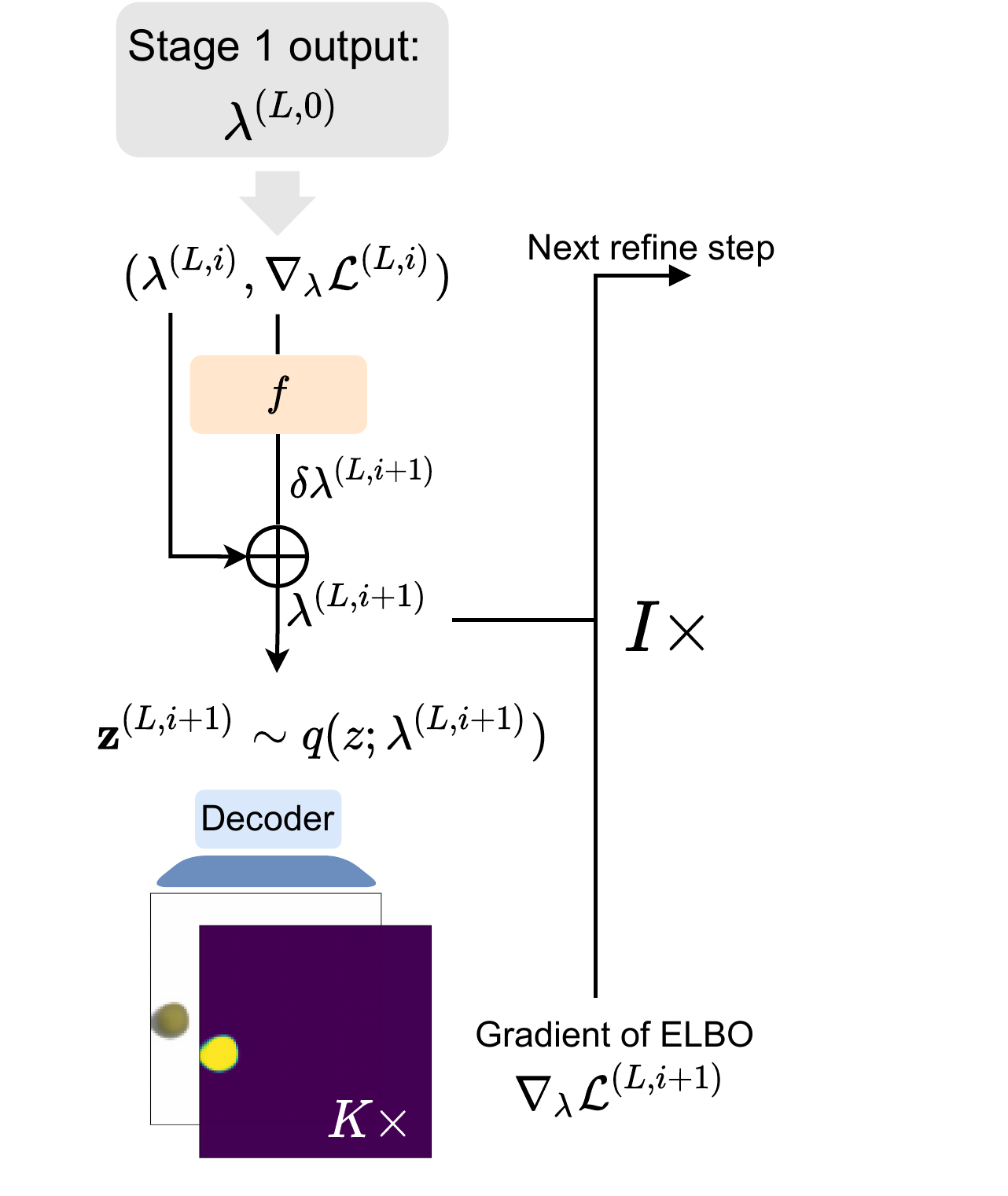}
        \caption{Stage 2\label{fig:mainC}}
    \end{subfigure}
    \begin{subfigure}[t]{0.22\textwidth}
        \includegraphics[trim=0cm 1cm 0cm 0cm,clip=True,scale=0.6]{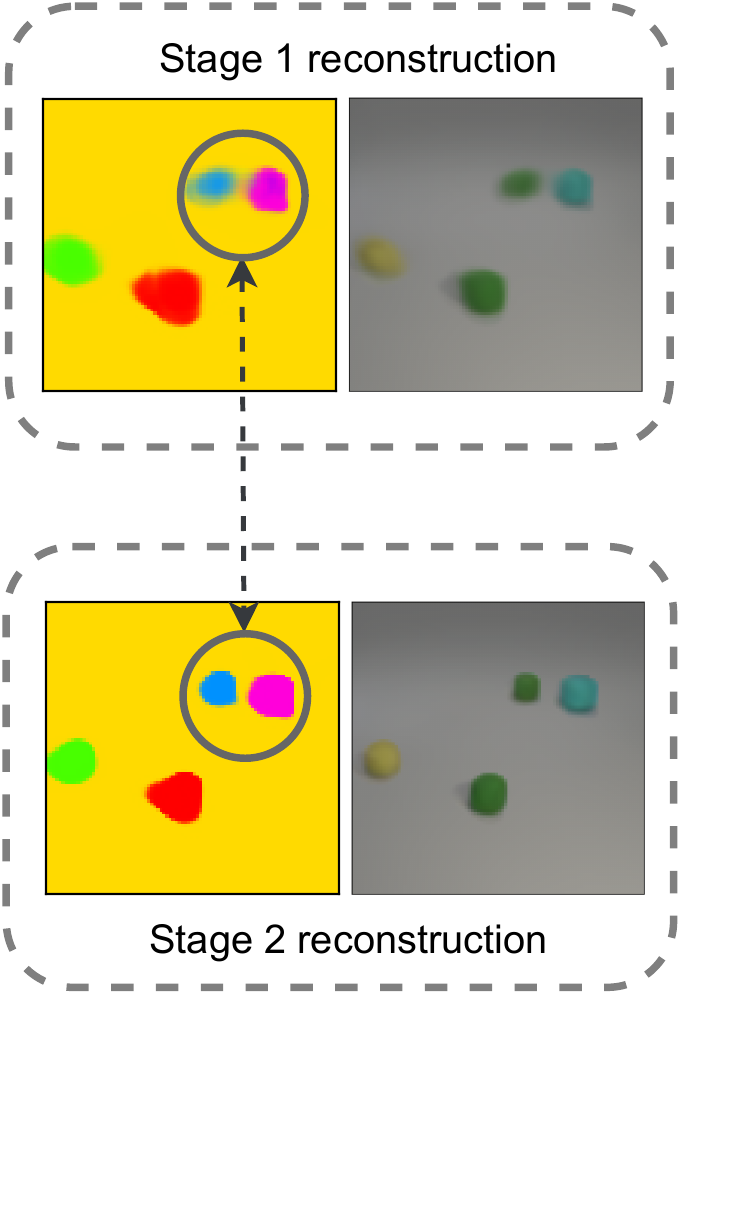}
        \caption{\label{fig:mainAC}}
    \end{subfigure}
    \hfill
    \caption{\textbf{Two-stage inference} a) Bottom-up inference over $L$ stochastic layers is used to iteratively extract $K$ symmetric and disentangled representations from an image $x$. 
    Disentanglement is achieved via hierarchical prior regularization.
    b) A \emph{lightweight} refinement network $f_\phi$ refines the Stage 1 posterior $\bm{\lambda}^{(L,0)}$ for $I$ steps. 
    Although $f_\phi$ has low-dimensional inputs and outputs to make it efficient, the decoding step for computing the loss $\mathcal{L}^{(L,i)}$ is still costly.
    c) Since EfficientMORL learns to use refinement to avoid getting stuck in poor local minima during the early phase of training, we find that we can speed up training by decreasing $I$ once Stage 1 starts converging.
    After training, the refinement stage can be removed at a small drop in decomposition performance for faster test time inference.\label{fig:main}}
    \vspace{-0.14in}
\end{figure*}

In this work, we show that IAI \emph{can} be used to solve multi-object representation learning while being as efficient as competing approaches \emph{and} without sacrificing representation quality.
Our idea is to cast the iterative assignment of inputs to symmetric representations as bottom-up inference in a multi-layer hierarchical variational autoencoder (HVAE).
A hierarchical prior regularizes the bottom-up posterior, disentangling the latent space. 
We use a two-stage inference algorithm to obtain a scene representation; the first stage uses the HVAE, and the second stage uses IAI to simply refine the HVAE posterior.
We find this is crucial for the HVAE to achieve reliable convergence to good local minima, particularly early on during training.
At test time, IAI can optionally be discarded.

\textbf{Contributions}: 
\begin{itemize}
    \item EfficientMORL, a framework for \emph{efficient} multi-object representation learning consisting of a hierarchical VAE and a lightweight network for iterative refinement
    \item Our method learns both \emph{symmetric} and \emph{disentangled} representations while being comparably efficient to state-of-the-art methods whose representations miss on at least one of the key inductive biases 
    \item An order of magnitude faster training and test time inference than the closest comparable method
\end{itemize}
\section{Related Work}
\textbf{Unsupervised image segmentation} Algorithms for unsupervised multi-object representation learning can be broadly differentiated by the use of hand-crafted or learned features. 
Unsupervised image segmentation algorithms~\cite{arbelaez2010contour,6205760} predate modern deep approaches and used perceptual grouping ideas to define features for clustering pixels in human-interpretable ways. 
These algorithms are still used in vision pipelines, and although the community's focus has shifted to learning representations from data, they contain valuable insights for improving neural methods~\cite{bear2020learning}.

\textbf{Spatial attention} How neural approaches decompose scenes into object-centric representations largely segregates the relevant literature.
AIR~\cite{eslami2016attend}, SPAIR~\cite{crawford2019spatially}, and SPACE~\cite{lin2020space} use spatial attention to discover explicit object attributes such as position and scale similarly to unsupervised object detection.
These models excel at decomposing and generating synthetic scenes~\cite{jiang2020generative,deng2021generative}, but the underlying grid-based scene representation and symbolic bounding-box-like object representations are ill-suited for handling complex real-world scenes.

\textbf{Sequential attention} 
MONet~\cite{burgess2019monet} and GENESIS~\cite{engelcke2019genesis} use sequential attention to bind latent variables to the components of a \emph{segmented} image but as a result learn \emph{ordered} representations.
Imposing an ordering on the set of representations for a scene is unnatural, can leak global scene information into the object-centric representations, and biases the decomposition (e.g., the background or large objects are always attended to first).\footnote{See Appendix A.3 of~\citet{pmlr-v97-greff19a} and our Appendix~\ref{sec:app:genesis} for a discussion on unordered vs ordered representations.} 
In a follow-up work, MONet was extended with an in-painting network to improve its spatial disentanglement~\cite{yang2020learning}.

\textbf{Iterative inference} A line of methods~\cite{greff2016tagger,greff2017neural,van2018relational,Yuan2019,pmlr-v97-greff19a} use iterative inference to bind symmetric latent representations with the components of a segmentation mixture model.
IODINE is the state-of-the-art method in this category; our method, described in the next section, presents an efficient alternative without sacrificing representation quality.
Slot Attention~\cite{locatello2020object} is a general method for mapping a distributed representation to a symmetric set representation and has been used within a deterministic autoencoder for unsupervised object discovery. 
However, it tends to learn highly entangled representations, unlike ours, since it can only implicitly encourage disentanglement by adjusting the latent dimension. 
The SRN~\cite{huang2020better} was published concurrently with Slot Attention and appears to offer a similar mechanism for mapping a single distributed representation to a set representation.
\section{EfficientMORL}
\label{sec:method_overview}
Our goal is to infer a set $\mathbf{z}$ := $\{z_1,\dots,z_K\}$, $z_k \in \mathbb{R}^D$ of object-centric representations from a color image $x \in \mathbb{R}^{H \times W \times 3}$.
We assume that $\mathbf{z}$ generates the image $x$ and that each element of $\mathbf{z}$ corresponds to a single object in the scene.
To solve the inverse problem of obtaining $\mathbf{z}$ from $x$, we could compute the posterior distribution $p(\mathbf{z} \mid x)$.
Bayes rule tells us that we also need the joint distribution $p(x, \mathbf{z}) = p(x \mid \mathbf{z}) p(\mathbf{z})$, which describes the image generation process.
Since the latent dimension $D$ can be large (e.g., $64$), computing $p(\mathbf{z} \mid x)$ requires solving an intractable integral.
Instead we use amortized variational inference~\cite{kingma2013auto} and compute an approximate variational posterior $q(\mathbf{z} \mid x)$.
Like IODINE, we make an independence assumption among the $K$ latent variables so that the variational posterior and prior are defined as symmetric products of $K$ multivariate Gaussians with diagonal covariance.
Neural nets with weights $\theta$ are used to obtain the parameters of the generative distribution; for the variational distributions, we use networks with weights $\phi$. Pseudocode for the inference algorithm is provided (Algorithm~\ref{alg:inference}) which we will reference by line number.

\subsection{Stage 1: The hierarchical variational autoencoder}
\label{sec:hvae}
\begin{algorithm}[t]
    \footnotesize
    \caption{\textbf{Two-stage inference} Linear attention maps $k,q,v$ with $D$ output units and LayerNorms (LN) are trainable. SP := Softplus. $\epsilon$ is for numerical stability. $N = HW$.}
    \label{alg:inference}
    \begin{algorithmic}[1]
    \STATE \textbf{Input:} image $x$ 
    \STATE $x = \texttt{image\_encoder}(x)$ 
    \STATE $x = \texttt{LayerNorm}(x + \texttt{pos\_encoding}(x))$
    \STATE \textcolor{gray}{ /* Stage 1: Bottom-up inference */}
    \STATE $\mathbf{z}^0 \sim \mathcal{N}(\mu^0, (\sigma^{0}I)^2)$
    \STATE $\bm{\lambda}^{0} = (\mu^0, \sigma^0)$
    \FOR{stochastic layer $l = 1\dots L$}
        \STATE $\mathbf{z}^{l-1} = \texttt{LayerNorm}(\mathbf{z}^{l-1})$
        \STATE $\bm{\alpha} = \texttt{softmax}_{K}\bigl (\frac{1}{\sqrt{D}} k(x) q(\mathbf{z}^{l-1}) \bigr)$
        \STATE $\bm{\alpha} = (\bm{\alpha} + \epsilon) / \sum_N (\bm{\alpha}+ \epsilon)$
        \STATE $\bm{\Theta} = \sum_N \bm{\alpha} \cdot v(x)$ 
        \STATE $\bm{\lambda}^{l} = \texttt{DualGRU}([\bm{\Theta}; \bm{\Theta}], \bm{\lambda}^{l-1})$
        \STATE $(\bm{\mu}^{l},\bm{\sigma}^{l}) = \bm{\lambda}^l$
        \STATE $\bm{\mu}^{l} \mathrel{+}= \texttt{MLP}(\texttt{LayerNorm}(\bm{\mu}^{l}))$
        \STATE $\bm{\sigma}^{l} \mathrel{+}= \texttt{SP}(\texttt{MLP}(\texttt{LayerNorm}(\bm{\sigma}^{l})))$
        \STATE $\mathbf{z}^{l} \sim \mathcal{N}(\bm{\mu}^{l}, (\bm{\sigma}^{l}I)^2)$ \hspace{2em} \textcolor{gray}{ /* $q(\mathbf{z}^l \mid x, \mathbf{z}^{l-1})$ */ }
    \ENDFOR
    \STATE $\mathcal{L}^{(L,0)} = \mathcal{L}_{\text{NLL}} + \infdiv{q(\mathbf{z}^{1:L} \mid x)}{p(\mathbf{z}^{1:L})}$
    \STATE \textcolor{gray}{ /* Stage 2: Iterative refinement */}
    \STATE $\bm{\lambda}^{(L,0)} = (\bm{\mu}^L, \bm{\sigma}^{L})$
    \FOR{refinement iter $i = 1\dots I$}
        \STATE $\nabla_{\bm{\lambda}} \bar{\mathcal{L}}^{(L,i-1)} = \texttt{LN}(\texttt{stop\_grad}(\nabla_{\bm{\lambda}} \mathcal{L}^{(L,i-1)}))$
        \STATE $\bm{\lambda}^{(L,i)}  = \bm{\lambda}^{(L,i-1)} + f(\bm{\lambda}^{(L,i-1)}, \nabla_{\bm{\lambda}} \bar{\mathcal{L}}^{(L,i-1)})$
        \STATE $\mathbf{z}^{(L,i)} \sim q(\mathbf{z} ; \bm{\lambda}^{(L,i)})$
        \STATE $\bm{\pi},\mathbf{y} = \texttt{decoder}(\mathbf{z}^{(L,i)}
)$
        \STATE $\mathcal{L}^{(L,i)} = \mathcal{L}_{\text{NLL}}^{(L,i)} + \infdiv{q(\mathbf{z} ; \bm{\lambda}^{(L,i)})}{p(\mathbf{z}^{L} \mid \mathbf{z}^{L-1})}$
    \ENDFOR
    \RETURN $\bm{\lambda}^{(L,I)}$  \textcolor{gray}{ /* The image representation */ }
    \end{algorithmic}
\end{algorithm}

\begin{figure}[t]
    \centering
    \includegraphics[scale=0.63]{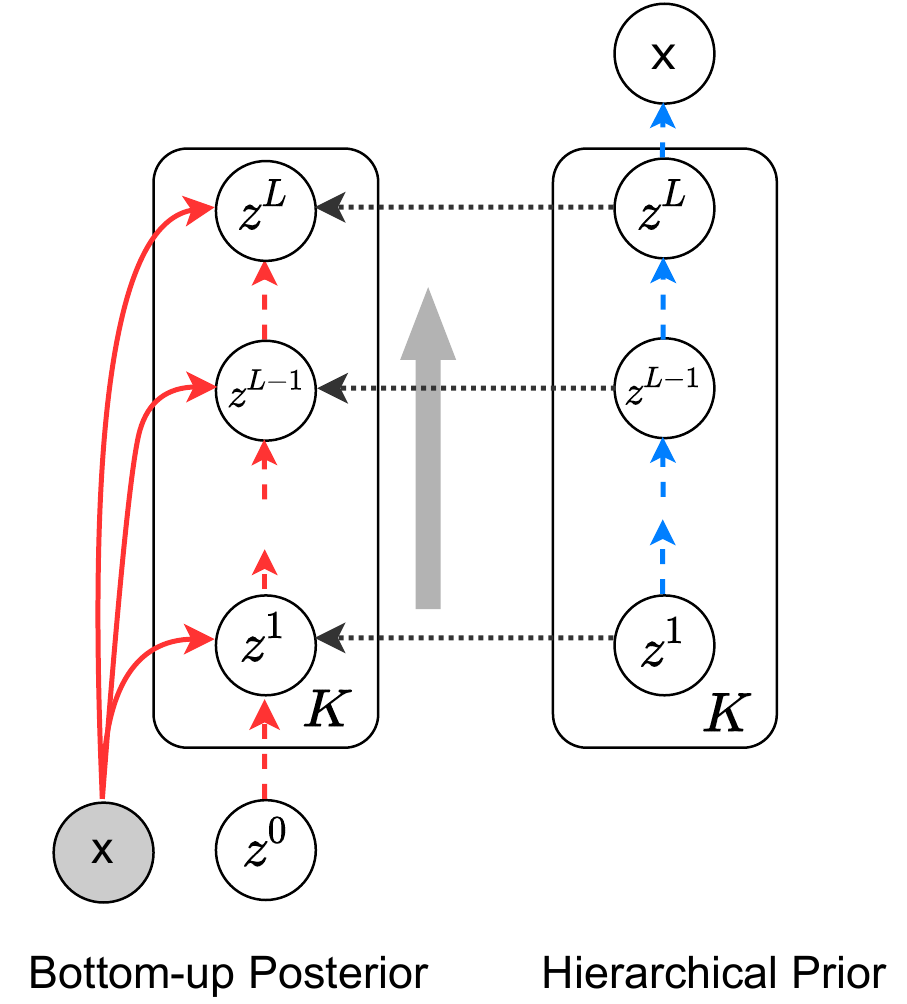}
    \caption{The HVAE's graphical model. Dashed arrows show reparameterized sampling, dotted arrows show prior regularization, and solid arrows show deterministic connections. White circles are random variables and gray are inputs. To try to mitigate posterior collapse, we also consider a variant of the prior where the arrows between $\mathbf{z}^1,\dots,\mathbf{z}^L$ are reversed (see Figure~\ref{fig:app:priors}).\label{fig:graphical}}
\end{figure}
\textbf{Bottom-up inference} The variational posterior is designed to perform an $L$-step iterative assignment of pixels to $K$ latents in a single pass.
As such, we split $\mathbf{z}$ across $L$ stochastic layers, creating a bottom-up pathway (Figure~\ref{fig:graphical}). 
The prior $p_\theta(\mathbf{z}^{1:L})$ regularizes the posterior $q_\phi(\mathbf{z}^{1:L} \mid x)$ at intermediate layers, disentangling the scene representation as a result.
The multi-layer variational posterior distribution is given by
\begin{align}
\label{eq:bottom-up}
     q_\phi(z^0_k) q_\phi(\mathbf{z}^{1:L} \mid x, \mathbf{z}^0) 
     &= q_\phi(z^0_k) \prod_{k=1}^{K} q_\phi({z}_k^{1:L} \mid x, z_k^0)  \\
     &= q_\phi(z^0_k) \prod_{k=1}^K \prod_{l=1}^{L} q_\phi({z}_k^{l} \mid x, {z}_k^{l-1}), \nonumber
\end{align}
where $z^0_k$ is Gaussian with learned mean $\mu \in \mathbb{R}^D$ and variance $\sigma^2 \in \mathbb{R}^D$ has been introduced as a zeroth layer.
Randomness is introduced by sampling $K$ times from $q_\phi(z^0)$ to help initially break symmetry.
\emph{Conditional on each sampled $z_k^0$}, \emph{the set of $K$ marginal distributions $q_\phi({z}_k^{1:L} \mid x, z_k^0)$ are equivariant with respect to permutations applied to their ordering}.
Equivalently, shuffling the order of $\mathbf{z}^0$ will likewise shuffle the order of the set of $K$ Gaussian posterior parameters at the $L$\textsuperscript{th} layer.  
Each layer must be designed to preserve this symmetry while mapping pixels to latents.

We achieve this by adapting the attention-based image-to-set mapping introduced by Slot Attention.
In detail, the $l$\textsuperscript{th} stochastic layer uses scaled dot-product set attention to attend over $N = HW$ tokens derived from a flattened embedding of an image augmented with a positional encoding.
The \textbf{key} $k$ and \textbf{value} $v$ are the embedded image, and the set-structured \textbf{query} $q$ is the \emph{stochastic} sample $\mathbf{z}^{l-1}$ from the previous layer's posterior.
The query is used to output set-structured features $\bm{\Theta} \in \mathbb{R}^{K \times D}$ as a function of the attention $\bm{\alpha} \in [0,1]^{K \times HW}$~(Lines 2-11).
Two GRUs~\citep{cho2014properties}, each with hidden dimension $D$, fuse the previous layer posterior's mean and variance with $\bm{\Theta}$ before an additive update to the predicted mean and variance with separate MLPs (Lines 14-15).
We justify the introduction of a second GRU by noting in an ablation study that the model struggled to learn to map the shared feature $\bm{\Theta}$ to the posterior mean and variance using a single GRU with hidden dimension $2D$. 
For a similar reason the MLPs predicting the Gaussian parameters are not shared, which is standard for VAEs.
In practice, we implement the two GRUs as a single GRU with \emph{block-diagonal} weight matrices that takes in the concatenated features $[\bm{\Theta}; \bm{\Theta}] \in \mathbb{R}^{K \times 2D}$ to parallelize the computation (DualGRU in Algorithm~\ref{alg:inference} and Figure~\ref{fig:main}).
Finally, we sample from the posterior which provides the next query (Line 16).
See Appendix~\ref{sec:app:equivariance} for verification that the permutation equivariance of the $K$ marginal distributions of the posterior are preserved during inference and Appendix~\ref{sec:app:implementation} for more details on the DualGRU implementation.

\textbf{Hierarchical prior} Multi-layer priors $p_\theta(\mathbf{z}^{1:L})$ for mean field HVAEs are often designed so that correlations among the latent variables can be captured to facilitate learning highly-expressive priors and posteriors.
Recently, a particular approach to accomplish this---top-down priors with bidirectional inference---has achieved impressive results for unconditional image generation~\cite{sonderby2016ladder,kingma2016improving,vahdat2020nvae,child2020very}.

However, we found it non-trivial to adapt such top-down priors for our setting.
First, it is not obvious how to re-use our unique bottom-up inference network within a top-down prior to combine their pathways~\citep{sonderby2016ladder}.
Second, it is also unclear how to design a suitable prior that uses global information to generate coherent multi-object scenes without removing the equivariance property.
We leave investigating these complex priors for future work and instead simply take the image likelihood and multi-layer prior to be
\begin{align}
   \label{eq:prior}
   p_\theta(x, \mathbf{z}^{1:L}) &= p_\theta(x \mid \mathbf{z}^L) p_\theta(\mathbf{z}^{1:L})\\
   &= p_\theta(x \mid \mathbf{z}^{L})
   \prod_{k=1}^K p(z_k^{1}) \prod_{l=2}^{L} p_\theta({z}_k^{l} \mid {z}_k^{l-1}). \nonumber 
\end{align} 
See Figure~\ref{fig:graphical} for the graphical model. 
The bottom-level prior $p(z_k^1)$ is a standard Gaussian distribution and we implement each layer $p_\theta({z}_k^{l} \mid {z}_k^{l-1})$ as an MLP with one hidden layer followed by two linear layers for the mean and variance respectively.
Note that our simple image likelihood distribution only depends on $\mathbf{z}^L$.

\textbf{Image likelihoods} 
We implement two image likelihood models.
For both, we use a spatial broadcast decoder~\cite{watters2019spatial} to map samples from the posterior or prior to $K$ assignment masks $\bm{\pi}$ normalized by softmax and $K$ RGB images $\mathbf{y}$.
The first model (\emph{Gaussian}) uses $\bm{\pi}$ to compute a weighted sum over $K$ predicted RGB values for each pixel, then places a Gaussian over the weighted sum with fixed variance $\sigma^2$.
The second is a pixel-wise \emph{Mixture of Gaussians} where $\bm{\pi}$ weighs each Gaussian in the sum~\cite{burgess2019monet,pmlr-v97-greff19a,engelcke2019genesis}. 
See Appendix~\ref{sec:app:implementation} for formal descriptions.
We found the discussion in the literature lacking  on the different inductive biases imbued by each image model so we explored this empirically (Section~\ref{sec:decomposition}).
We observed that the Gaussian model tends to split the background across the $K$ components whereas the mixture model consistently places the background into a single component.

\subsection{Stage 2: Iterative amortized inference}
\label{sec:refinement}
Given the bottom-up posterior, prior, and decoder, it is possible to train the HVAE with just the single-sample approximation of the Evidence Lower Bound (ELBO) without IAI.
We define the negative log-likelihood as
\begin{equation}
    \mathcal{L}_{\text{NLL}} = -\mathbb{E}_{\mathbf{z}^{L} \sim q_\phi(\mathbf{z}^{L} \mid x)} \bigl[ \log p_\theta(x \mid \mathbf{z}^{L}) \bigr],
\end{equation}
where the expectation is computed using ancestral sampling. 
The KL divergence factorized by layer between the prior and posterior is
\begin{align}
    \label{eq:KL}
    &\phantom{= }\infdiv{q_\phi(\mathbf{z}^{1:L} \mid x)}{p_\theta(\mathbf{z}^{1:L})}  \\
    &= \mathbb{E} \bigl [ \infdiv{q_\phi(\mathbf{z}^{1} \mid x, \mathbf{z}^{0})}{p(\mathbf{z}^{1})} \bigr] \nonumber \\
    &\phantom{= }+ \sum_{l=2}^{L} \mathbb{E} \bigl [ \infdiv{q_\phi(\mathbf{z}^{l} \mid x, \mathbf{z}^{l-1})}{p_\theta(\mathbf{z}^{l} \mid \mathbf{z}^{l-1})} \bigr], \nonumber 
\end{align}
then we can minimize the negative ELBO 
\begin{equation}
    \label{eq:elboloss}
    \mathcal{L}^{(L,0)} = \mathcal{L}_{\text{NLL}} + \infdiv{q_\phi(\mathbf{z}^{1:L} \mid x)}{p_\theta(\mathbf{z}^{1:L})}.
\end{equation}
\begin{figure}[t]
    \centering
    \includegraphics[scale=0.45]{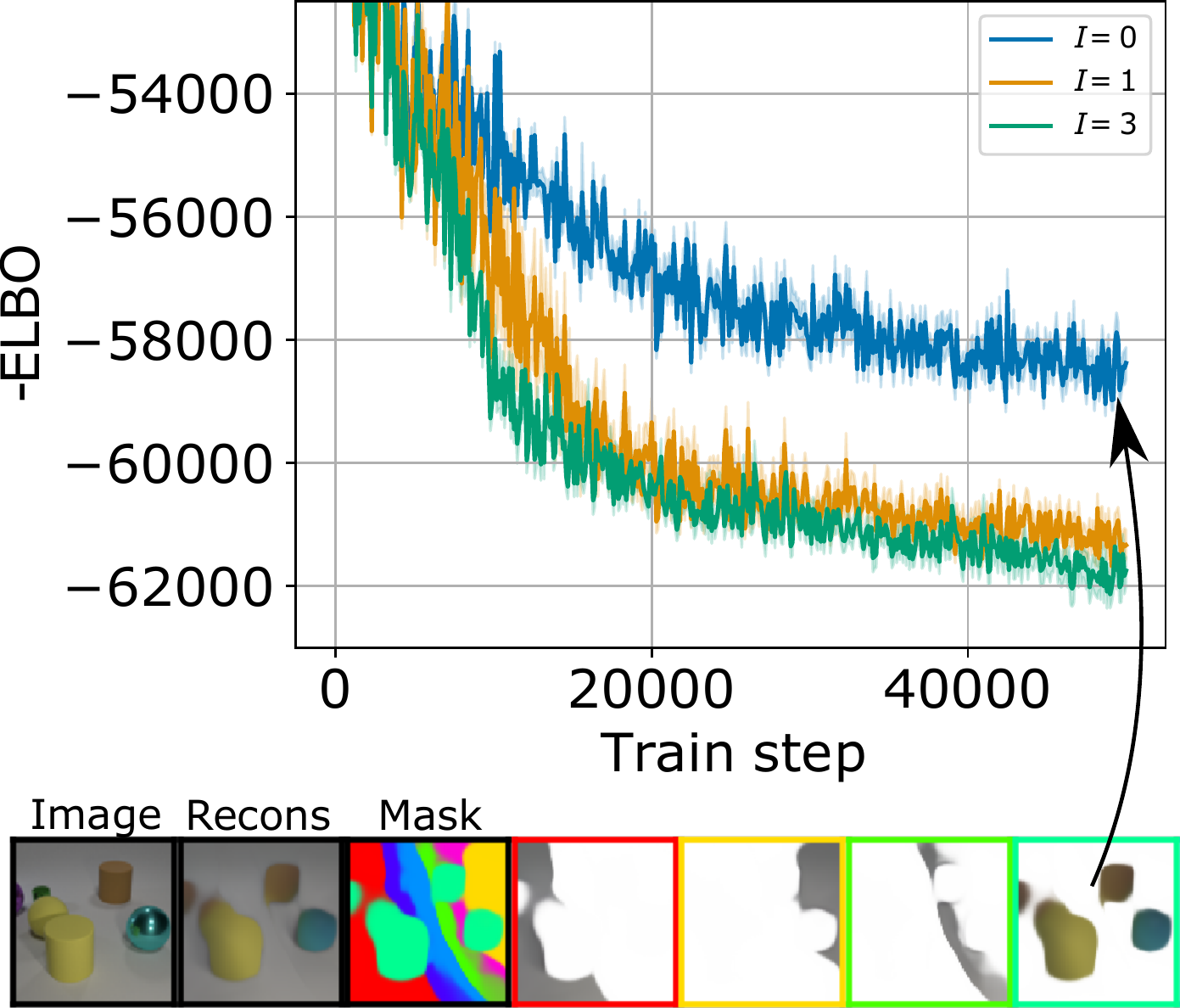}
    \caption{\textbf{Without iterative amortized inference the HVAE achieves a poor ELBO.} Training loss using $I=0,1,3$ refinement steps. Results for each $I$ are averaged across 10 random seeds with 95\% C.I. shown. Without refinement, the HVAE consistently gets stuck in poor local minima early on during training and does not recover. 
    For example, all foreground objects get placed in a single component.
    With just one refinement step we find that the HVAE can better avoid such minima.
    \label{fig:num_refine_iters}}
\end{figure}
Finding good local minima is challenging because the posterior is highly multi-modal due to the symmetric latent structure. Penalizing the model for learning entangled representations through prior regularization makes optimization more challenging as well.
We hypothesize that IAI can address this since it uses top-down feedback---the negative ELBO---to refine the posterior.
In Figure~\ref{fig:num_refine_iters}, we show that without IAI the HVAE consistently converges to a poor ELBO.
It also learns the optimal step size for the refinement update making it highly effective.
\emph{The question remains of how to use it without incurring a large increase in computation since evaluating the negative ELBO requires decoding $K$ image-sized masks and RGB components.}

\textbf{Stage 2 efficiency} As shown in Figure~\ref{fig:main} we use a two-stage approach to inference.
First we use ancestral sampling and the HVAE's bottom-up inference pathway to compute $\bm{\lambda}^{L} := \{\bm{\mu^{L}, \bm{\sigma}^{L}} \}$---the parameters of the marginal distribution $q_\phi(\mathbf{z}^L \mid x)$.
The goal of the second stage is to use a \emph{lightweight} refinement network $f_\phi$ for $I$ steps of IAI (Lines 21-27) to refine $\bm{\lambda}^{L}$ with top-down feedback.
As Figure~\ref{fig:main} shows, the HVAE primarily uses IAI to make small refinements to $\bm{\lambda}^{L}$, particularly during the early training stages.
This means that with only a small number of steps we can see a large improvement in final performance (Figure~\ref{fig:num_refine_iters}) and suggests employing training strategies such as reducing the number of steps after the HVAE starts to converge to reduce overall training time (Section~\ref{sec:analysis}).
Unlike IODINE, our refinement network $f_\phi$ does not take in image-sized inputs, which greatly reduces the number of model parameters and makes refinement even faster during training.
 
\textbf{Stage 2 refinement network} At refinement step $i$, we use a simple network $f_\phi$ that encodes the concatenated Gaussian parameters $\bm{\lambda}^{(L,i-1)}$ and the gradient of the refinement loss $\nabla_{\bm{\lambda}}\mathcal{L}^{(L,i-1)}$.
For the latter, we use layer normalization~\cite{ba2016layer} and stop gradients from passing through (Line 22)~\cite{Marino2018}.
The network $f_\phi$ is an MLP that encodes the input, a GRU with hidden dimension $D$, and two linear layers to compute an additive update:
  \begin{align}
     \delta \bm{\lambda}^{(L,i-1)} &= f_\phi(\bm{\lambda}^{(L,i-1)}, \nabla_{\bm{\lambda}}\mathcal{L}^{(L,i-1)}) \\
     \bm{\lambda}^{(L,i)} &= \bm{\lambda}^{(L,i-1)} + \delta \bm{\lambda}^{(L,i-1)}.
 \end{align}
The refinement loss $\mathcal{L}^{(L,i)}$ (Line 26) is the negative ELBO, defined as the KL divergence between the refined posterior and $p_\theta(\mathbf{z}^L \mid \mathbf{z}^{L-1})$ plus the negative log-likelihood.

\subsection{Training} 

\textbf{Training loss} The loss that gets minimized is
\begin{equation}
    \label{eq:finalloss}
    \mathcal{L} = \mathcal{L}^{(L,0)} + \sum_{i=1}^{I} \frac{I - (i-1)}{I+1} \mathcal{L}^{(L,i)},
\end{equation}
where the discount factor $\frac{I - (i-1)}{I+1}$ emphasizes the loss near $i=0$ to place more weight on the encoder than on the refinement network; this has the opposite effect of the discount factor used by IODINE which places more weight on later refinement terms (large $i$). 

\textbf{Posterior collapse} HVAEs can suffer from posterior collapse~\cite{sonderby2016ladder}, which is when each layer of the approximate posterior collapses to the prior early on during training and never recovers.
We applied mitigation strategies for this in three parts of the framework: the graphical model of the hierarchical prior, the prior used in the refinement loss KL term, and the training objective.

We created a variant of the hierarchical prior where the arrows between $\mathbf{z}^1, \dots, \mathbf{z}^L$ are reversed.
This \emph{reversed prior} now has $p(\mathbf{z}^L)$ as the standard Gaussian and each intermediate layer is given by $p_\theta(\mathbf{z}^l \mid \mathbf{z}^{l+1})$. 
Intuitively, penalizing the posterior at layer $L$ by its deviation from a standard Gaussian imposes a stricter constraint on the posterior's expressiveness. 
By exploring this alternative prior, we can determine whether a better match between the flexibility of the posterior and our simple prior helps address collapse.
We then considered replacing the prior in the refinement loss KL term with $p_\theta(\mathbf{z}^1 \mid \mathbf{z}^2)$ (e.g., \emph{reversed prior++}).
We hypothesized that implicitly pushing the posteriors at lower layers to match the posterior at layer $L$ during refinement should help keep them from collapsing to the prior.
See Figure~\ref{fig:app:priors} for a comparison of the variants.
The loss (Equation~\ref{eq:finalloss}) was modified to use GECO~\cite{rezende2018taming}. 
GECO reformulates the ELBO to initially allow the KL to grow large so that a predefined reconstruction threshold can first be attained.
We chose GECO since tuning its hyperparameters was easier than for deterministic warm-up~\cite{sonderby2016ladder}. 

Our ablation studies analyzing the contribution of each strategy can be found in Appendix~\ref{sec:app:ablations}. 
We note that GECO was needed for CLEVR6 and Tetrominoes but not on Multi-dSprites.
We use the best-performing variant, \emph{reversed prior++}, for the experiments in Section~\ref{sec:experiments}.
\section{Experiments}
\label{sec:experiments}
The evaluation of EfficientMORL is organized as follows.
We first analyze the refinement updates to suggest a principled justification for our training strategy (Section~\ref{sec:analysis}).
Then, we evaluate object decomposition (Section~\ref{sec:decomposition}) and disentanglement performance (Section~\ref{sec:representation}).
Finally, we compare run time and memory costs with competing models (Section~\ref{sec:efficiency}).
Additional qualitative results and the ablation studies on the hierarchical prior, refinement steps $I$, and the DualGRU are in Appendix~\ref{sec:app:ablations}.

\textbf{Datasets}  We use the Multi-Object Dataset~\cite{multiobjectdatasets19} for all experiments. 
This benchmark has two sprites-based environments (Tetrominoes and Multi-dSprites) and a synthetic 3D environment (CLEVR) with ground truth object segmentation masks. 
We follow the same training and evaluation protocol as \citet{pmlr-v97-greff19a,locatello2020object}.
For both sprites datasets, we split the data by using the first $60$K samples for training and then hold out the next $320$ images for testing.
We filter all CLEVR images with less than seven objects to create a $50$K training set (CLEVR6). For testing, we also use $320$ held-out images.
To evaluate generalization we use a test set of $320$ images containing $7-10$ objects (CLEVR10).
Note that after center cropping the CLEVR images, we resize them to $96 \times 96$ instead of $128 \times 128$, unlike~\citet{locatello2020object,pmlr-v97-greff19a}, to make replicating our CLEVR experiments across multiple random seeds practical.

\textbf{Hyperparameters} 
All models use the Adam~\cite{kingma2014adam} optimizer, a learning rate of $4$e-$4$ with warm-up and exponential decay, gradient norm clipping to $5.0$, and a mini-batch size of $32$.
Following~\cite{pmlr-v97-greff19a,locatello2020object} we use $K = 7$ components for CLEVR6, $K = 6$ for Multi-dSprites, and $K = 4$ for Tetrominoes (note that $K$ can be set to any number for any given image if desired).
Complete model architecture, optimizer, and evaluation details are provided in Appendix~\ref{sec:app:implementation}. 

\subsection{IAI steps analysis}
\label{sec:analysis}
\begin{figure}
    \centering
    \begin{subfigure}[t]{0.48\columnwidth}
        \includegraphics[scale=0.41]{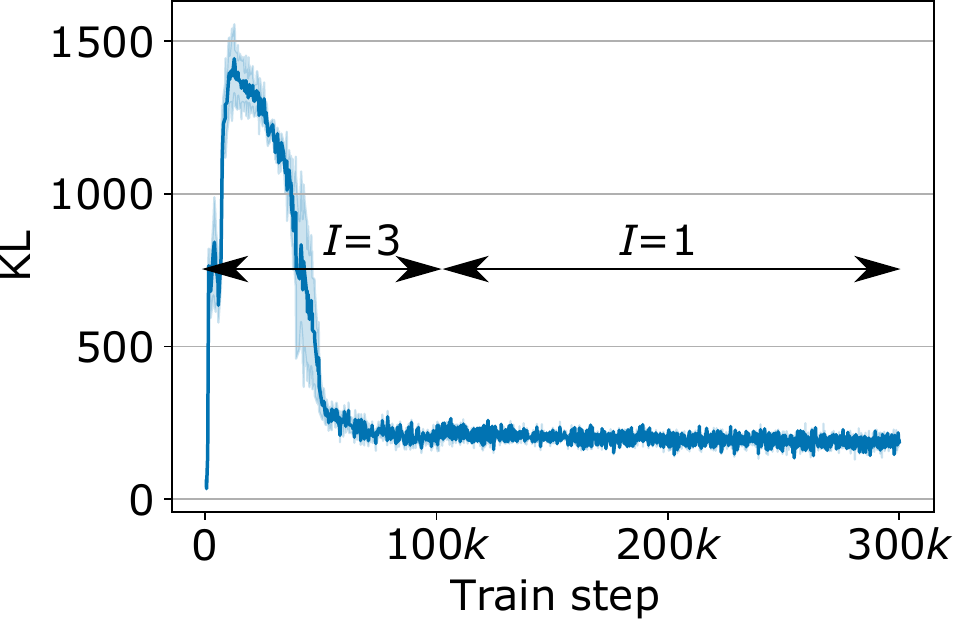}
        \caption{}
    \end{subfigure}
    \begin{subfigure}[t]{0.48\columnwidth}
        \includegraphics[scale=0.41]{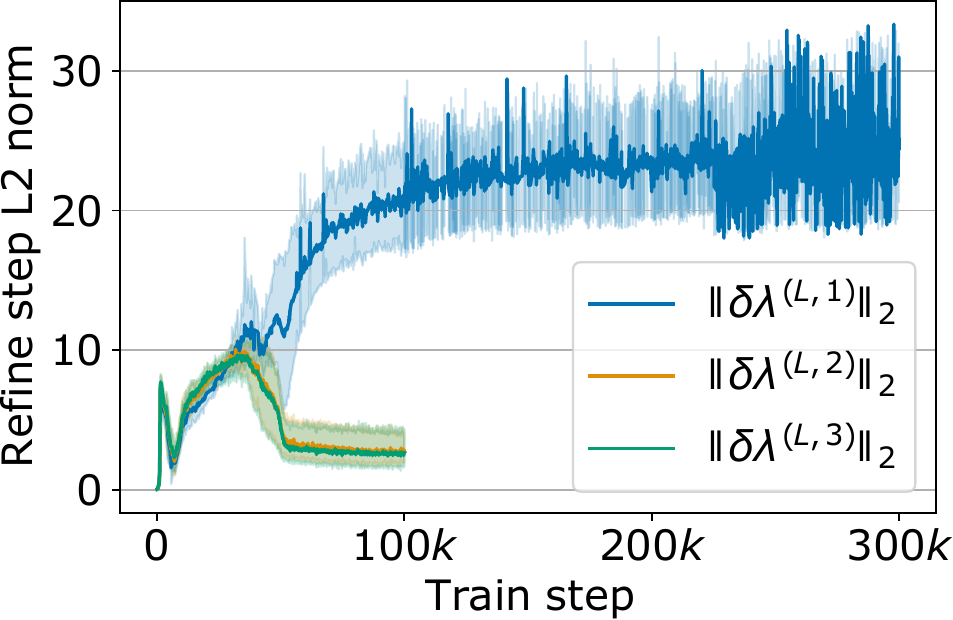}
        \caption{}
    \end{subfigure}
    
    \caption{\textbf{The L2 norm of the refinement updates are good indicators for when we can reduce $I$, speeding up training}. a) Training curves of the final KL across five CLEVR6 runs.
    The early spike in KL is due to GECO. b) As the KL starts to converge around $100$K steps, the L2 norm of the updates for $I >1$ also decreases. This suggests they are no longer contributing much to the final posterior. }
    \label{fig:analysis}
    \vspace{-1em}
\end{figure}
To better understand the role of IAI in EfficientMORL and to justify decreasing $I$ during training to reduce training time, we analyze the KL curves from five training runs on CLEVR6 and plot the L2 norm of each refinement update $\delta \bm{\lambda}^{(L,i)}$ (Figure~\ref{fig:analysis}). 
Across runs, we observe that as the KL starts to converge, the L2 norm of updates for steps $I > 1$ became small.
Our interpretation is that at this point, the bottom-up posterior from stage one is sufficiently good such that more than one refinement step is not needed.
When we decrease $I$ from three to one during training, we notice a slight drop in reconstruction quality that the model quickly recovers from.
We attribute the continued increase in L2 norm for $\delta \bm{\lambda}^{(L,1)}$ to the observation that a single refinement step at test time has a larger effect on the KL than on the reconstruction quality (Figure~\ref{fig:Clevr6-KL-v-iters}, Figure~\ref{fig:app:Tetris-varying-I}).

In our experiments we did not decrease $I$ to one from three on Tetrominoes due to the small image size and fast convergence in $200K$ steps.
On CLEVR6 and Multi-dSprites we train for $100$K steps with $I=3$, then train for another $200$K steps with $I$ decreased to one.
One step is used at test time for evaluating these two environments.

\subsection{Object decomposition}
\label{sec:decomposition}
\begin{table}[t]
    \centering
    \scriptsize
    \caption{Multi-object Benchmark results. Adjusted Rand Index (ARI) scores (mean $\pm$ stddev for five seeds). \textbf{We achieve comparable performance to state-of-the-art baselines.} We outperform IODINE on Multi-dSprites.
    We replicated Slot Attention's CLEVR6 results over five random seeds (**) but one run failed---without it, the ARI improves from $93.3$ to $98.3$. Tetrominoes (*) was reported with only 4 seeds~\cite{locatello2020object}.}
    \begin{tabular}{llll}
        \toprule
         & CLEVR6 & Multi-dSprites & Tetrominoes \\
        \hline
        Slot Attention & $98.8 \pm 0.3$ & $\mathbf{91.3 \pm 0.3}$ & $99.5 \pm 0.2^*$ \\
        Slot Attention (**) & $93.3 \pm 11.1$ & --- & --- \\ 
        IODINE & $\mathbf{98.8 \pm 0.0}$ & $76.7 \pm 5.6$ & $\mathbf{99.2 \pm 0.4}$ \\
        MONet & $96.2 \pm 0.6$ & $90.4 \pm 0.8$ & --- \\
        Slot MLP & $60.4 \pm 6.6$ & $60.3 \pm 1.8$ & $25.1 \pm 34.3$ \\
        \hline
        EfficientMORL & $96.2 \pm 1.6$ & $\mathbf{91.2 \pm 0.4}$  & $98.2 \pm 1.8$ \\
        \hline
    \end{tabular}
    \label{tab:mob}
\end{table}
\begin{figure}[t]
    \centering
    \includegraphics[scale=0.63,trim=10.5cm 0cm 7cm 0cm, clip=True]{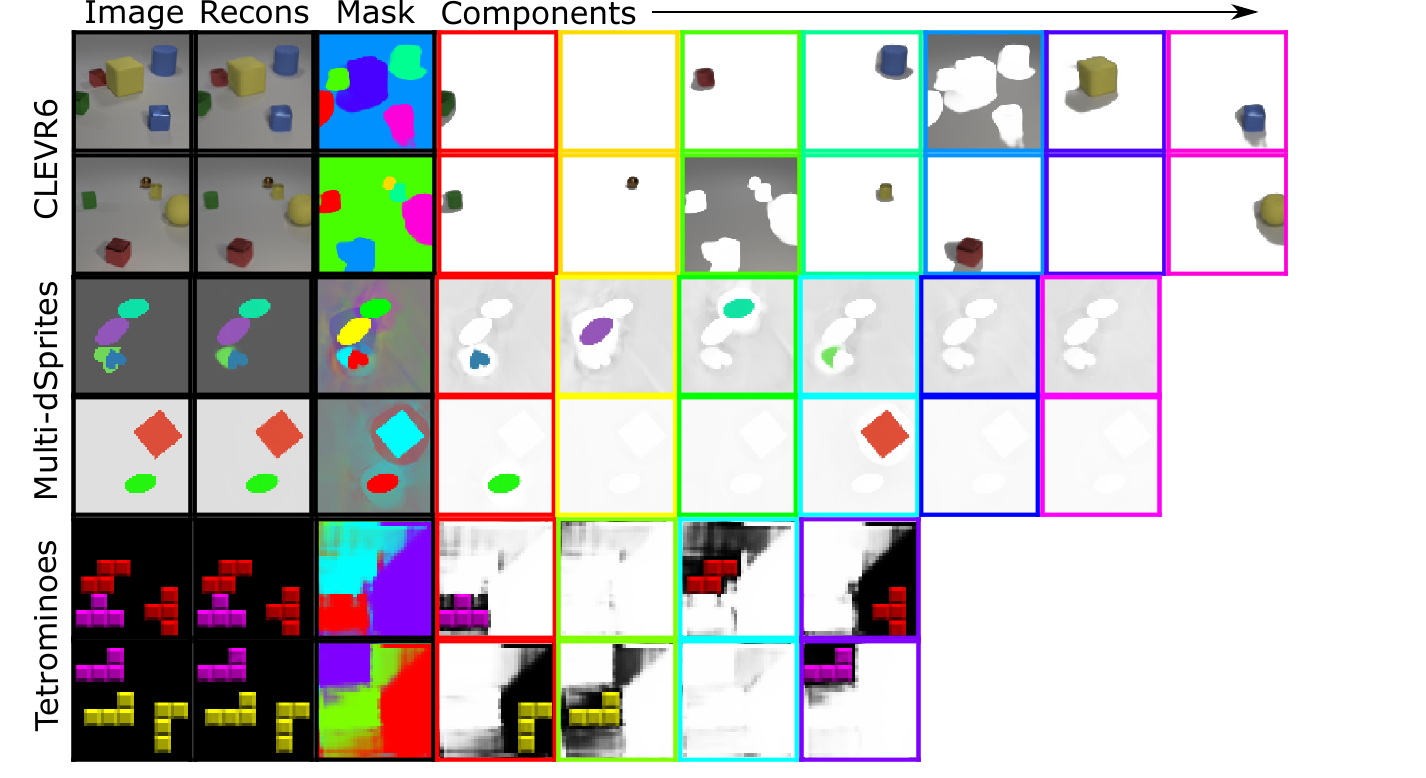}
    \caption{Visualization of scene decompositions for CLEVR6 (top), Multi-dSprites (middle) and Tetrominoes (bottom). The Mixture of Gaussians model used for CLEVR6 places the background into a single component, whereas the Gaussian model splits simple backgrounds across all components.}
    \label{fig:mob_qual}
\end{figure}
\begin{figure}[t]
    \centering
    \includegraphics[scale=.6]{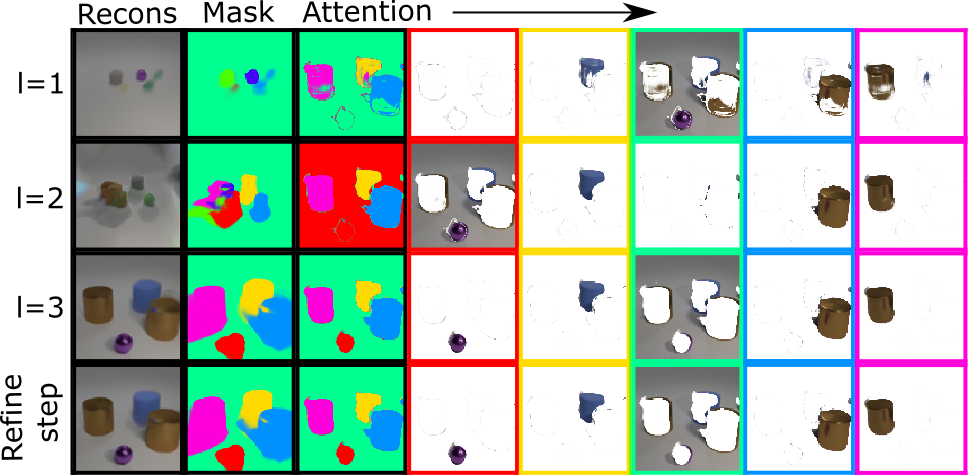}
    \caption{The first two columns show reconstructions/masks from the posteriors, column three shows the hard assignment of the softmax attention $\bm{\alpha}$ (Algorithm~\ref{alg:inference}, Line 9), and columns 4-8 show the same attention drawn over the input image.}
    \label{fig:encode-refine}
\end{figure}
\begin{figure}[t]
    \begin{subfigure}[t]{0.49\columnwidth}
        \centering
        \includegraphics[scale=0.26]{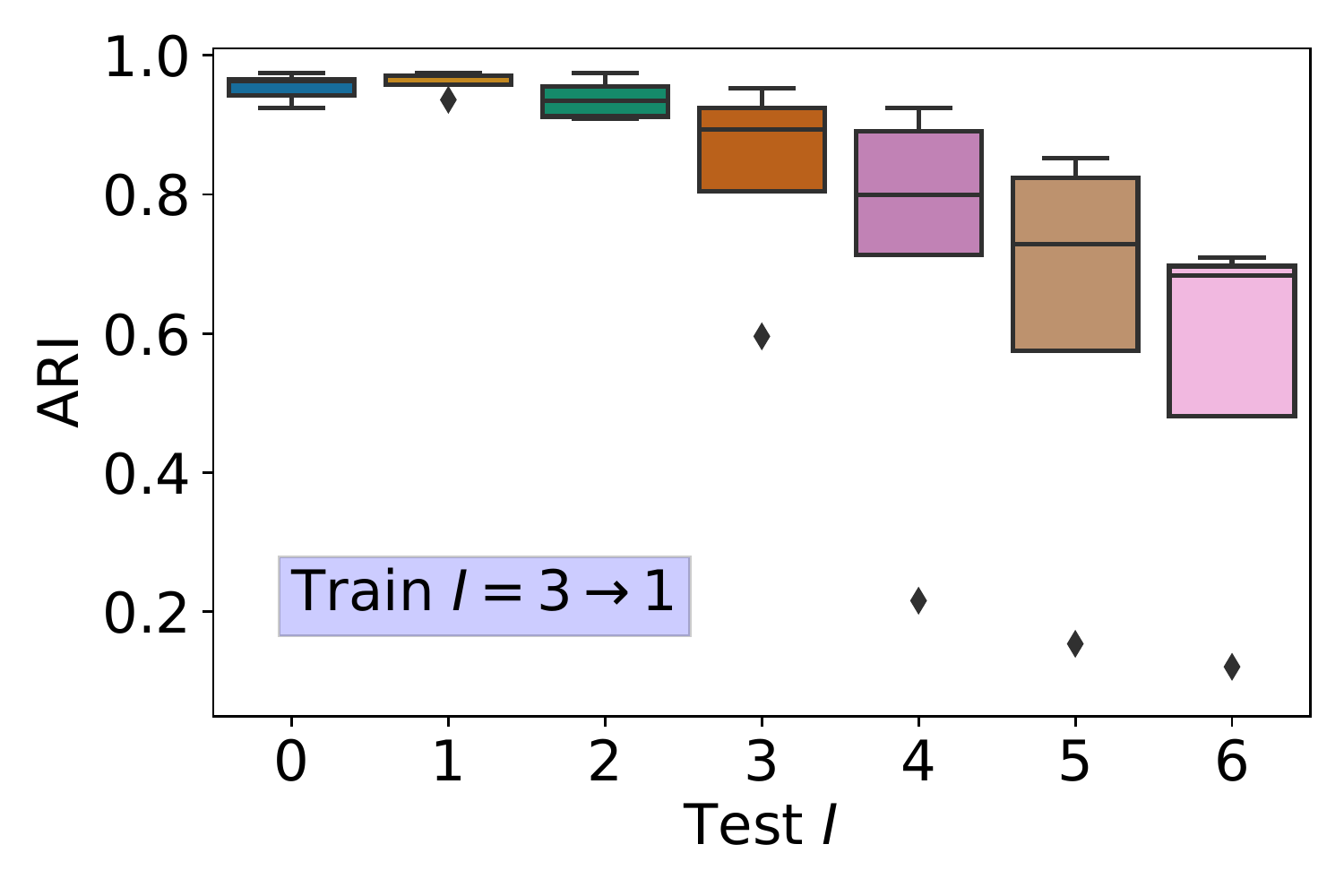}
        \caption{CLEVR6 ARI vs. $I$\label{fig:ARI-v-iters}}
    \end{subfigure}%
    \hfill
    \begin{subfigure}[t]{0.49\columnwidth}
        \centering
        \includegraphics[scale=0.26]{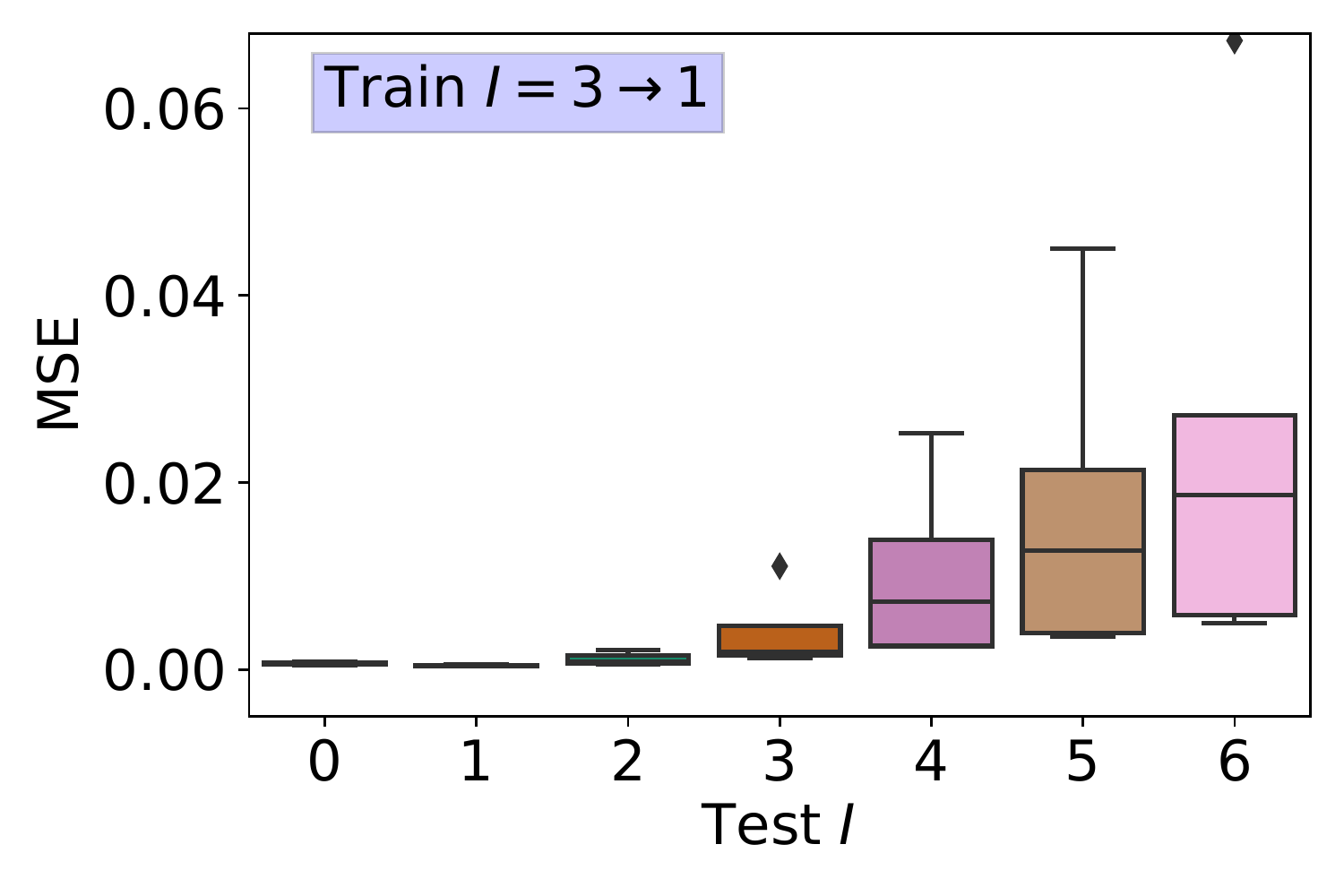}
        \caption{CLEVR6 MSE vs  $I$\label{fig:MSE-v-iters}}
    \end{subfigure}
    \begin{subfigure}[t]{0.49\columnwidth}
        \centering
        \includegraphics[scale=0.26]{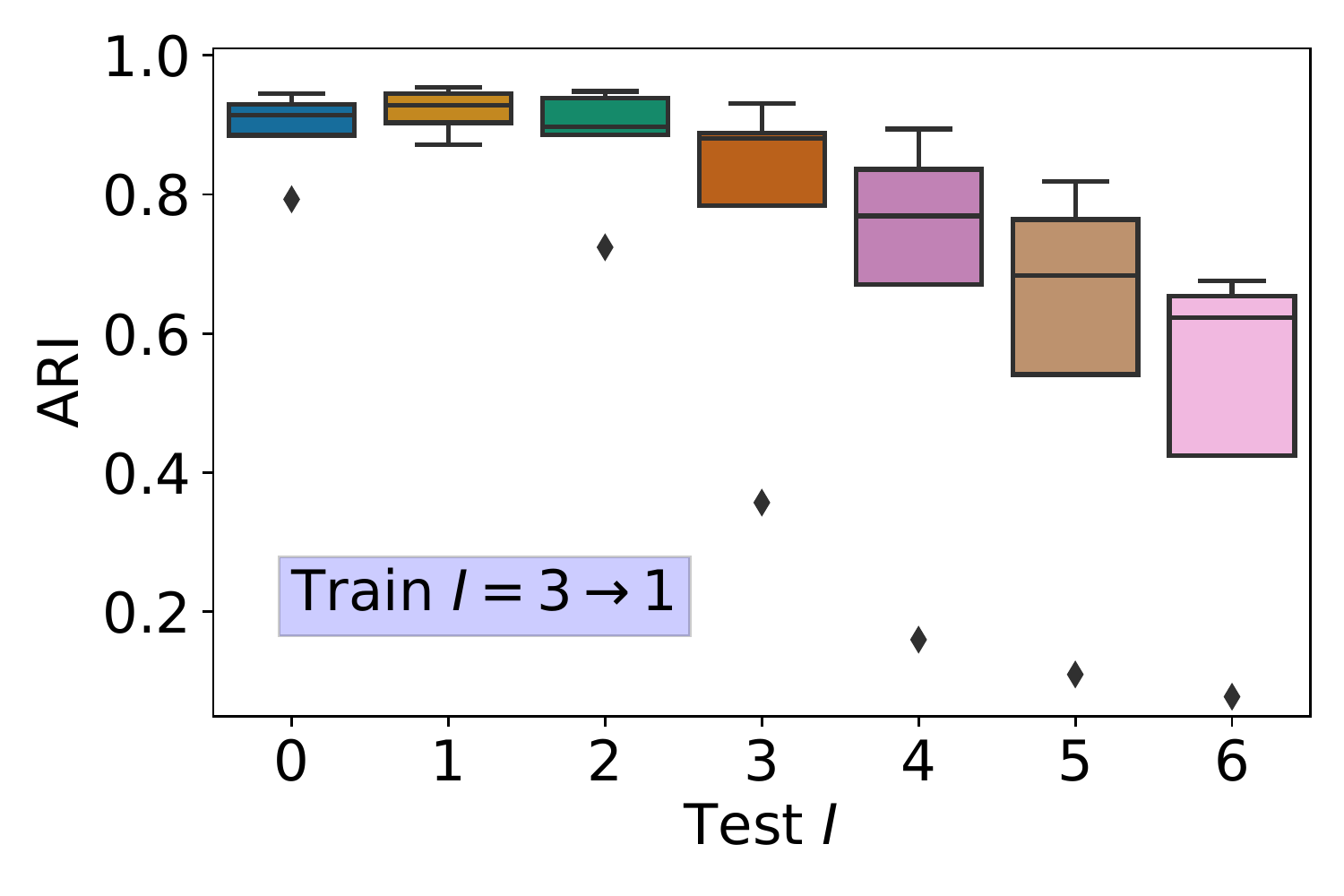}
        \caption{CLEVR10 ARI vs. $I$\label{fig:Clev10-ARI-v-iters}}
    \end{subfigure}%
    \hfill
    \begin{subfigure}[t]{0.49\columnwidth}
        \centering
        \includegraphics[scale=0.26]{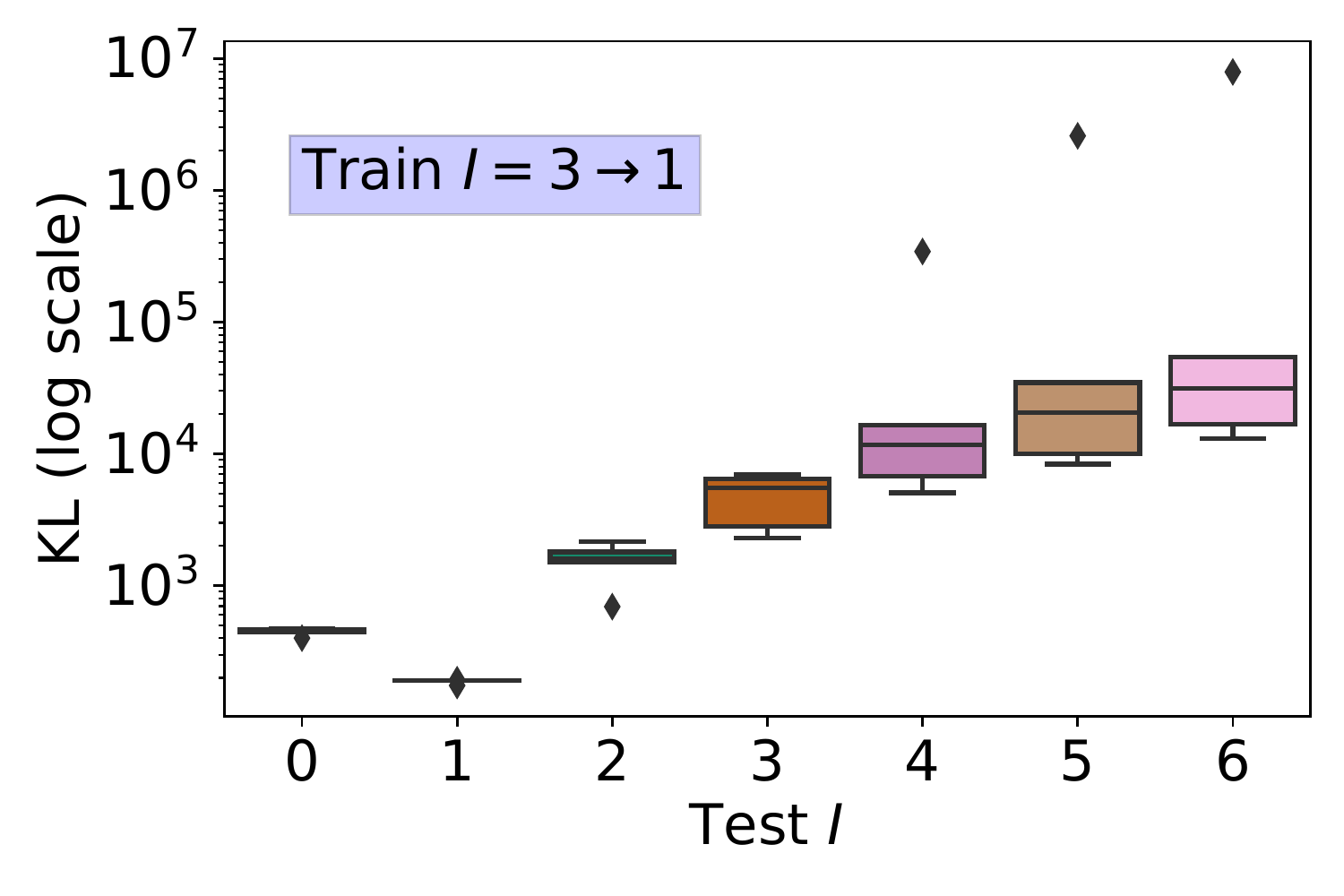}
        \caption{CLEVR6 KL vs. Test $I$\label{fig:Clevr6-KL-v-iters}}
    \end{subfigure}    
    \caption{a-b) \textbf{Refinement marginally improves segmentation and reconstruction at test time}. We show ARI and MSE when varying the number of test refinement steps $I$ on CLEVR6. c) \textbf{The model generalizes to larger numbers of objects at test time}. We increase components $K$ to $11$ for CLEVR10 ($7$-$10$ objects). d) Test time refinement impacts the KL more strongly than segmentation as evidenced by the increased KL at $I=0$.\label{fig:varying_iters}}
\end{figure}
\textbf{Baselines and metrics} The baselines are Slot Attention~\cite{locatello2020object}, IODINE~\cite{pmlr-v97-greff19a}, MONet~\cite{burgess2019monet}, and the Slot MLP baseline from~\cite{locatello2020object} which maps an embedded image to an \emph{ordered} set of $K$ slots.
To measure decomposition quality we use the adjusted rand index (ARI)~\cite{rand1971objective,hubert1985comparing} and do not include the background mask in the ARI computation following standard practice for this benchmark.
We also compute pixel mean squared error (MSE), which takes into account the background.

\textbf{Analysis on the number of stochastic layers} 
\begin{figure}[th]
    \centering
    \includegraphics[scale=0.4]{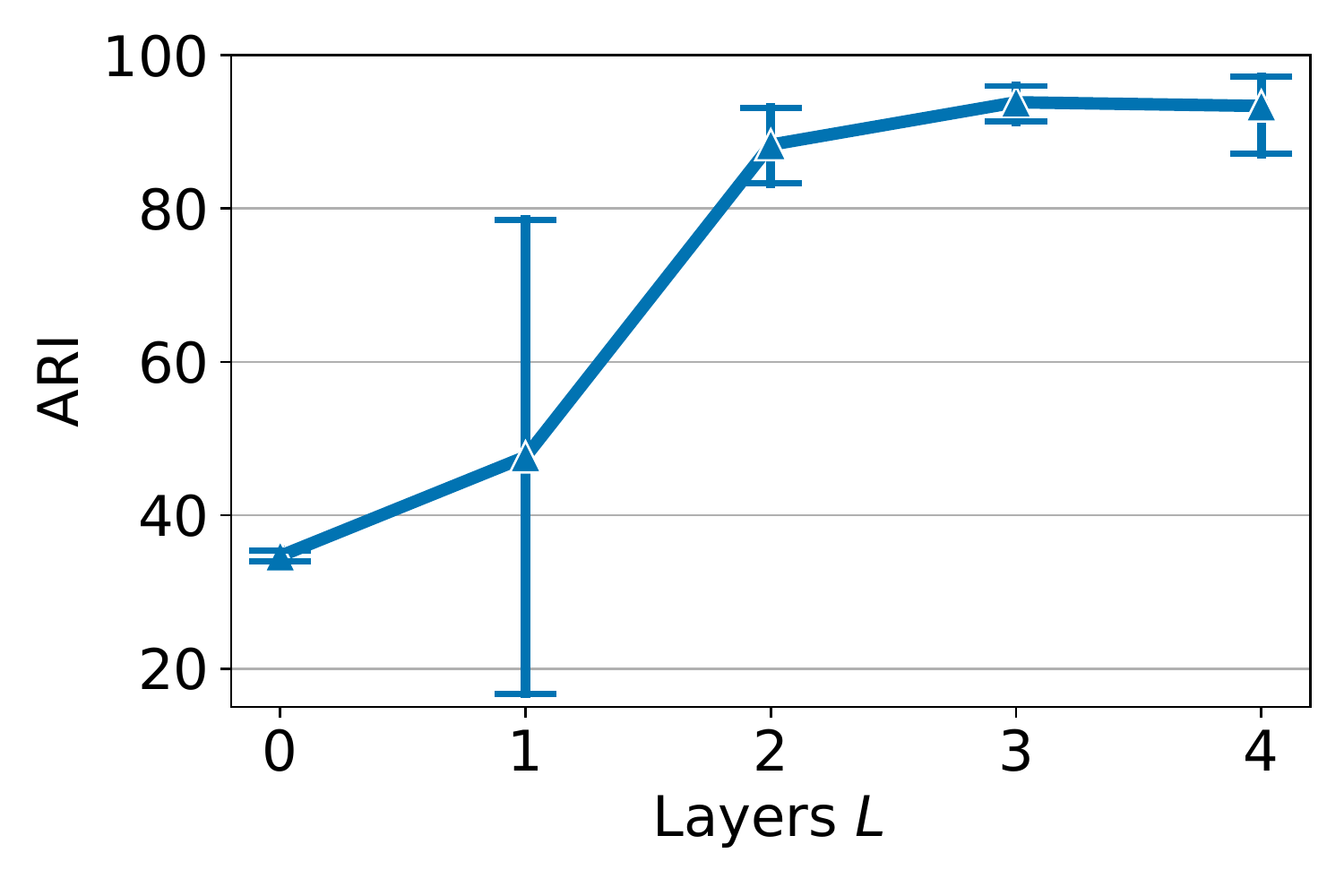}
    \caption{Sensitivity analysis on the number of layers $L$ in the HVAE. Results are averaged over five CLEVR6 training runs. Best performance is achieved at $L=3$.}
    \label{fig:layers}
\end{figure}
We vary the number of stochastic layers $L$ in the prior and posterior (Figure~\ref{fig:layers}) during training and measure the ARI.
Best results are obtained with $L=3$, which we use for all experiments. 
Note that when $L=0$, EfficientMORL extracts the scene representation with only $I$ refinement steps.

\textbf{Main results}
ARI scores are in Table~\ref{tab:mob} and qualitative examples for each environment are in Figure~\ref{fig:mob_qual}.
Overall, EfficientMORL's decomposition performance is comparable to Slot Attention, MONet, and IODINE on all three environments.
EfficientMORL uses the Gaussian image likelihood for the two sprites environments because it quickly and reliably converges, whereas the Mixture of Gaussians had difficulty discovering the sprites. 
The Mixture of Gaussians is used for CLEVR6 and biases the model towards assigning the background to a single component in each training run, which may be desirable. 
Since the Gaussian likelihood biases the model to split the background across all components, it appears to be better suited to handle simple single-color backgrounds (note that IODINE uses the mixture model on all environments, which may explain its lower scores on Multi-dSprites). 

\textbf{Intermediate posteriors} We reconstruct samples drawn from the intermediate posteriors and visualize the layer-wise masks and attention over the input image in Figure~\ref{fig:encode-refine}.
The single refinement step applied to $\bm{\lambda}^{(3,0)}$ imperceptibly changes the reconstructed image and segmentation.
The intermediate posterior reconstructions suggest they have not collapsed to the prior and that the top-level posterior fits an expressive non-Gaussian distribution.
 
\textbf{Varying test time IAI steps} Our earlier analysis (Figure~\ref{fig:num_refine_iters}) demonstrated that during \emph{training}, using $I>0$ stabilizes convergence to achieve a better ELBO, with $I=3$ slightly outperforming $I=1$.
Notably, \emph{zero} IAI steps at test time can achieve $99.1\%$ of the refined ARI and MSE (Figures~\ref{fig:ARI-v-iters},~\ref{fig:MSE-v-iters}).
We see a larger gap in the KL between zero and one step (Figures~\ref{fig:Clevr6-KL-v-iters},~\ref{fig:app:Tetris_KL_BU},~\ref{fig:app:Tetris_KL_TD}), which suggests that refinement plays a larger role in achieving this aspect of the extracted high-quality representation at test time.
The decrease in test ARI as $I$ is increased to six is due to the refinement GRU ignoring sequential information after reducing $I$ to one during training.
If $I$ is held at three, we find this is no longer the case (Figure~\ref{fig:app:Tetris-varying-I} in the appendix).

\textbf{Systematic generalization} We evaluate whether EfficientMORL generalizes when seeing more objects at test time by increasing $K$ to $11$ and varying $I$ on CLEVR10.
As expected due to the equivariance property, we only see a slight drop in ARI (Figures~\ref{fig:Clev10-ARI-v-iters}).

\subsection{Disentanglement}
\label{sec:representation}
\begin{figure*}[t]
    \centering
    \begin{minipage}{0.8\textwidth}
        \begin{subfigure}[t]{0.49\textwidth}
            \centering
            \includegraphics[scale=0.5]{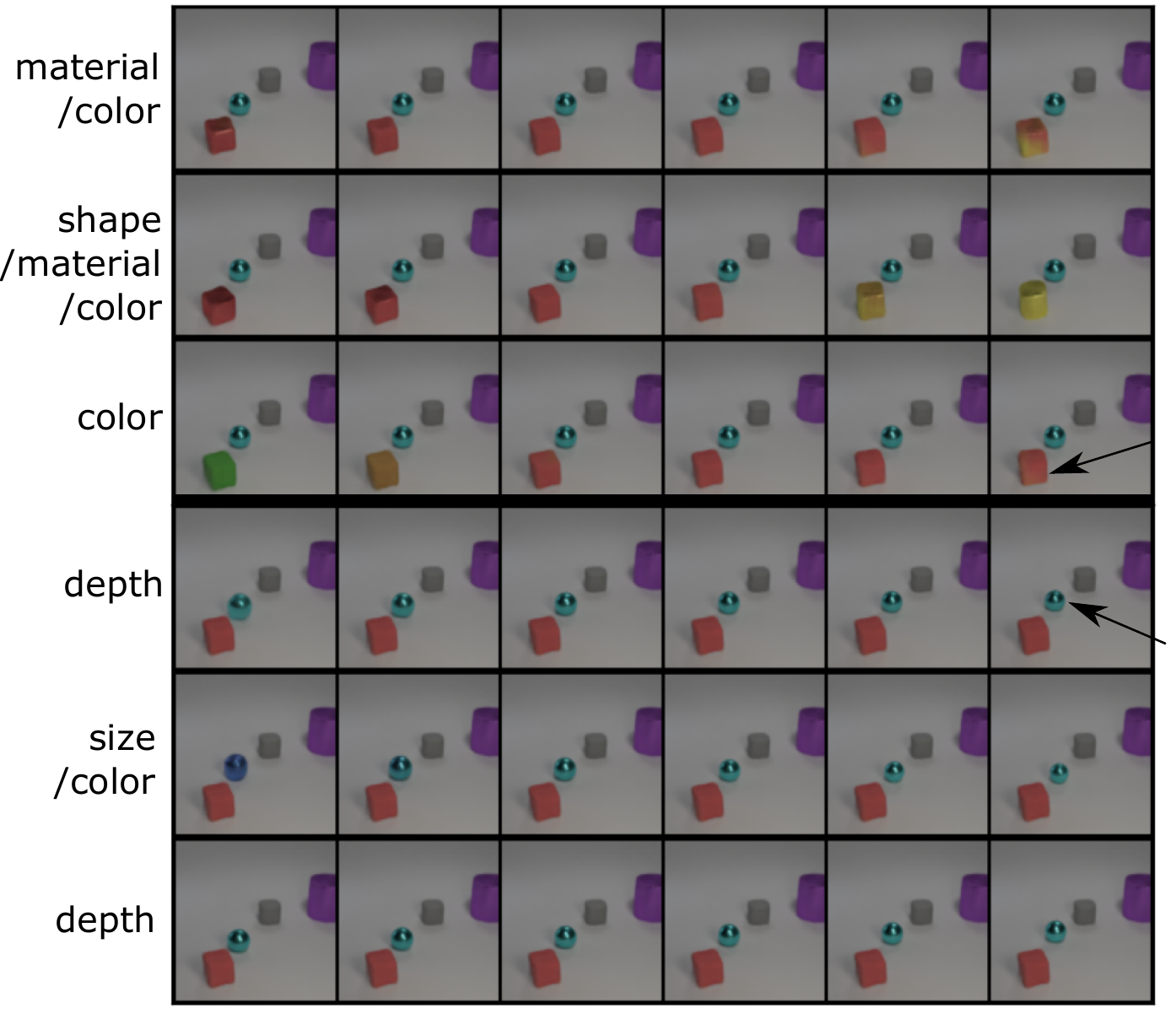}
            \caption{Slot Attention (DCI: $0.46$ / $0.38$ / $0.34$) \label{fig:SA-disentangle}}
        \end{subfigure}
        \hfill
        \begin{subfigure}[t]{0.49\textwidth}
            \centering
            \includegraphics[scale=0.5]{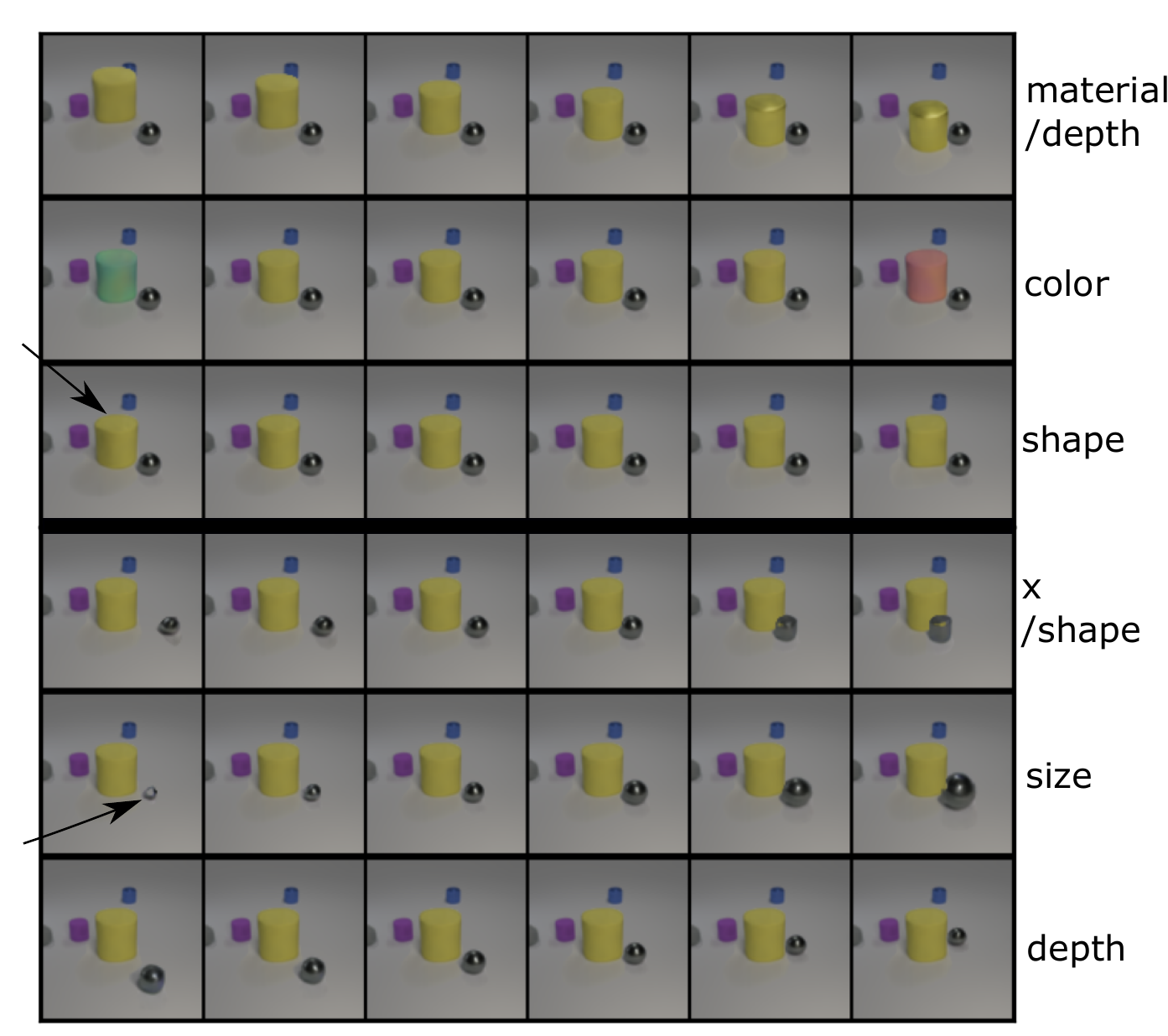}
            \caption{EfficientMORL (DCI: $\mathbf{0.63}$ / $\mathbf{0.63}$ / $\mathbf{0.46}$)\label{fig:EM-disentangle}}
        \end{subfigure}
    \end{minipage}
    \hfill
    \begin{minipage}{0.18\textwidth}
        \begin{subfigure}[b]{1\textwidth}
            \centering
            \includegraphics[scale=0.4]{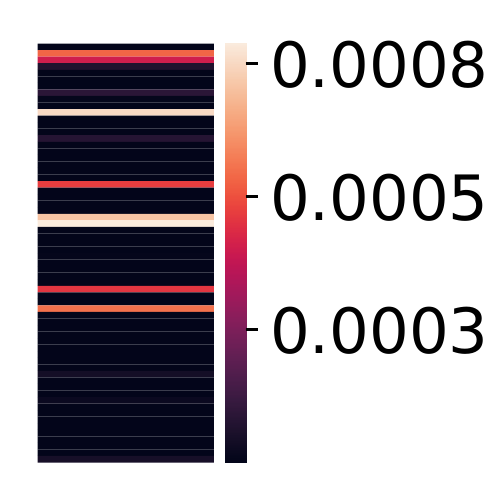}
            \caption{EfficientMORL\label{fig:EMORL_activeness}}
        \end{subfigure}%
        \quad
        \begin{subfigure}[b]{1\textwidth}
            \centering
            \includegraphics[scale=0.4]{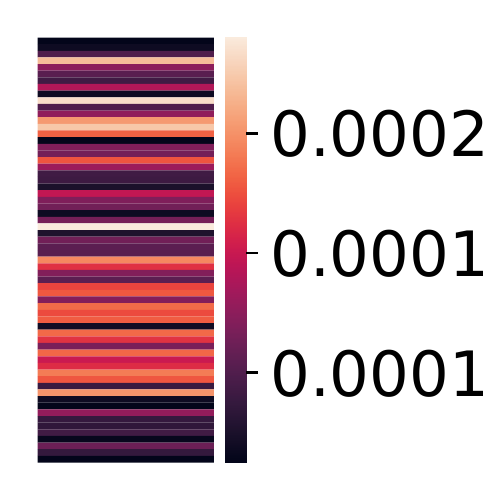}
            \caption{Slot Attention\label{fig:SlotAttention_activeness}}
        \end{subfigure}
    \end{minipage}
    \caption{\textbf{EfficientMORL outperforms Slot Attention at disentangling object attributes.} (a-b) We traverse latent dimensions for two objects in one scene (top three, bottom three rows). Rows are labeled by the attributes we observe to change, with entangled dimensions annotated by multiple attributes. DCI scores are in the captions (higher is better). The latent dimensions of EfficientMORL's representation ($\mathbf{z}^{(l=3,i=1)}$) have less correlation and redundancy (multiple dimensions controlling the same attribute). (c-d) \textbf{EfficientMORL has fewer \emph{active} latent dimensions than Slot Attention}. Perturbing most of the 64 dimensions has no effect on the reconstructed image.}
\end{figure*}
\begin{figure}[th!]
    \centering
    \includegraphics[scale=0.4]{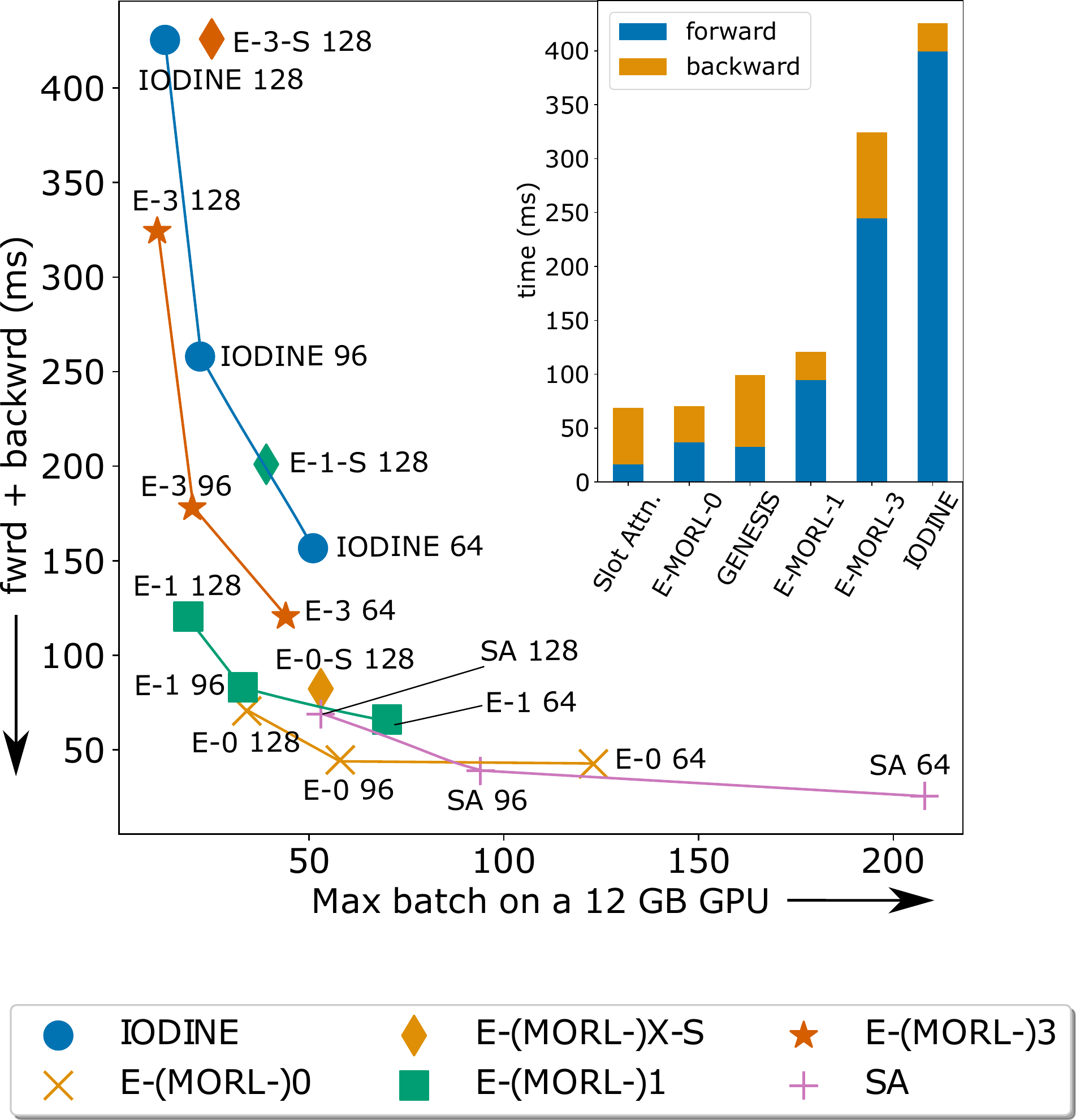}
    \caption{Lines connect results for each image size. E-X-S is ours with the memory-efficient decoder, SA := Slot Attention. Across all settings except E-$3$ we have better memory consumption than IODINE. Top right) \textbf{With $I=0$ we have a $10\times$ faster forward pass than IODINE and with $I \leq 1$ we are comparable to Slot Attention/GENESIS}. Times are for $128\times128$ images.\label{fig:efficiency}}
\end{figure}
\textbf{Baselines and metrics}  
The goal of this experiment is to compare the disentanglement quality of EfficientMORL's representations against the current state-of-the-art \emph{efficient} and \emph{equivariant} model Slot Attention on CLEVR6.
We recall that Slot Attention is a deterministic autoencoder without regularization, so we expect that it learns entangled scene representations.
Quantifying disentanglement for multi-object scenes is challenging since widely accepted metrics like DCI~\citep{eastwood2018framework} require access to the oracle matching---which is unknown--- between the $K$ inferred representations and the set of ground truth factors.
Apart from heuristically estimating DCI scores (see Appendix~\ref{sec:app:disentanglement} for details), we visualize latent dimension interpolations and compute the \emph{activeness}~\cite{peebles2020hessian} of the latent dimensions.
Activeness is the mean image variance when changing the $i$\textsuperscript{th} dimension of one uniformly sampled latent $z_k$ across $100$ test images.
Intuitively, a disentangled model should have many deactivated latent dimensions that do not change the image when perturbed.
While a similar comparison against other models like IODINE and GENESIS would be interesting, their authors did not provide disentanglement scores such as DCI, and we had just enough resources to train Slot Attention with multiple random seeds.
Although we leave a broader comparison future work, see Appendix~\ref{sec:app:multicolored} for a specific case study comparing GENESIS and our model.

\textbf{Results} Figure~\ref{fig:EM-disentangle} contains the DCI scores and examples of varying latent dimensions for two different objects in a single scene.
Our method's better disentanglement is verified by much higher DCI scores.
Many of Slot Attention's latent dimensions change multiple factors (for example, one changes the shape, material, and color of a single object), while we observe this in ours for only a small number. 
Moreover, by examining Slot Attention's activeness heatmap (Figure~\ref{fig:SlotAttention_activeness}), we see that the majority of latent dimensions contribute to the mean variance, whereas the majority of latent dimensions in ours are deactivated (Figure~\ref{fig:EMORL_activeness}).

\subsection{Efficiency}
\label{sec:efficiency}
\textbf{Setup} 
We measure the time taken for the forward and backward passes on $1$ 2080 Ti GPU with a mini-batch size of $4$, images sizes $64 \times 64$, $96 \times 96$, and $128\times 128$, and $I \in \{0,1,3\}$.
For comparison we use our own implementation of IODINE ($K=7$ and $5$ inference steps), GENESIS with $K=7$, and Slot Attention with $K=7$ and $L=3$ in PyTorch~\cite{torch}.
We checked implementation details against the official releases to ensure a fair comparison.
Also, to compare memory consumption we record the largest mini-batch size able to fit on $1$ 12 GB 2080 Ti GPU.

\textbf{Results} Figure~\ref{fig:efficiency} shows the key results with the remainder in the appendix. 
We show that our model with $I=0$ has $10\times$ faster forward pass (e.g., at test time) and a $6\times$ faster forward + backward pass than IODINE.
In general, our model has better memory consumption than IODINE up until $I=3$. 
However, if we replace the decoder with Slot Attention's deconvolutional decoder, we can double the maximum mini-batch size that fits on a single GPU (E-X-S in Figure~\ref{fig:efficiency}).
The impact of increased run time and memory between $I=1$ and $I=3$ on wall clock training time is mostly offset by decreasing $I$ after the model starts to converge early on, a unique aspect of our model.

The wall clock time to train EfficientMORL on CLEVR6 with a mini-batch size of $32$ split evenly across $8$ GPUs is about $17$ hours.
Slot Attention reports a wall clock training time of $24$ hours on CLEVR6, but they train using the full $128\times128$ images.
Controlling for that, our model takes slightly longer than Slot Attention to train on CLEVR.
IODINE reportedly trains for $1$M steps, which would take $10\times$ longer than our model took to converge.
\section{Discussion}
We introduced EfficientMORL, a generative modeling framework for learning object-centric representations that are symmetric and disentangled without the intense computation typically required when using iterative amortized inference and without sacrificing important aspects of the scene representation. 
EfficientMORL's relatively fast training times and test time inference make it useful for exploring topics such as object-centric representation learning for video.
While these models still require some engineering to work well on new environments, further study will likely lead to more general approaches.

\section*{Acknowledgements}
We thank Hadi Abdullah for providing valuable comments and suggested revisions on an early draft. 
This work is supported in part by NSF Awards 1446813 and 1922782. 
Patrick Emami is also supported in part by a FEF McKnight fellowship and a UF CISE graduate research fellowship.
Any opinions, findings, and conclusions expressed in this material are those of the authors and do not necessarily reflect the views of the National Science Foundation.

{\normalsize
\bibliographystyle{icml2021}
\bibliography{main}
}
\clearpage
\appendix 
\onecolumn

\section{Equivariance and invariance properties}
\label{sec:app:equivariance}
In this section, we verify that the implementations of the marginal distributions $q_\phi(z_k^{1:L} \mid x, z_k^0)$ and $p_\theta(z_k^{1:L})$, the decoder network, and the refinement network all achieve the desired equivariance property.
Additionally, we demonstrate that EfficientMORL is also \emph{invariant} to permutations applied to the order of the $N$ inputs $x_n \in \mathbb{R}^{D}, n=1,\dots,N$, where $N = HW$ and $H,W$ are the dimensions of the image.
The verification itself is mostly straightforward, as we mainly rely on weight-tying in the neural network architectures to symmetrically process each element of any set-structured inputs. 

\begin{definition}[Permutation invariance]
Given $\text{X} \in \mathbb{R}^{N \times D_1}$, $f : \mathbb{R}^{N\times D_1} \rightarrow \mathbb{R}^{N \times D_2}$, and any $N \times N$ permutation matrix $\Pi$, we say that $f$ is permutation invariant if
\[
f(\Pi \text{X}) = f(\text{X}).
\]
\end{definition}

\begin{definition}[Permutation equivariance]
Given $\text{X} \in \mathbb{R}^{N \times D_1}$, $f : \mathbb{R}^{N\times D_1} \rightarrow \mathbb{R}^{N \times D_2}$, and any $N \times N$ permutation matrix $\Pi$, we say that $f$ is permutation equivariant if
\[
f(\Pi \text{X}) = \Pi f(\text{X}).
\]
\end{definition}

\subsection{Equivariance} 

\textbf{Bottom-up posterior} Recall that there is an initial \emph{symmetry-breaking} step of sampling $K$ times from $\mathcal{N}(\mu^0, (\sigma^{0}I)^2)$ and consider the first layer of the marginal, $q_\phi(\mathbf{z}^1 \mid x, \mathbf{z}^0)$, which is conditioned on samples $\mathbf{z}^0$.
First, the samples are used to predict set-structured features $\bm{\Theta} \in \mathbb{R}^{K \times D}$ using scaled dot-product set attention~\citep{locatello2020object}.
By appealing to the proof in Section D.2 of~\citet{locatello2020object}, all operations within the scaled dot-product set attention (Lines 8-11 of Algorithm~\ref{alg:inference}) preserve permutation equivariance of the features $\bm{\Theta}$.
Following this, we concatenate $\bm{\Theta}$ with itself along the $D$-dimension and pass it to the DualGRU. 
The DualGRU uses weight-tying across $K$ to symmetrically process inputs $[\bm{\Theta},\bm{\Theta}]$ and previous state $[\bm{\mu}^0,\bm{\sigma}^0]$.
Next, the two MLPs also use weight-tying across $K$ to output $K$ means and variances.
Finally, we sample $\mathbf{z}^1$ from $q_\phi(\mathbf{z}^1 \mid x, \mathbf{z}^0)$. 
We only sample once from each of the $K$ marginals to generate the input with which to condition $q_\phi(\mathbf{z}^2 \mid x, \mathbf{z}^1)$.
Each stochastic layer thereafter is computed identically, and therefore the marginals of the bottom-up posterior are equivariant with respect to permutations applied to their ordering.

The \textbf{hierarchical prior} preserves the symmetry of its marginals since it consists of only feed-forward layers applied individually to each element of a sample $\mathbf{z}$ via weight-tying across $K$.
Both considered decoders are permutation invariant since they aggregate the $K$ masks and RGB components with a summation over $K$, which is a permutation invariant operation (see Section~\ref{sec:app:implementation} for decoder details).
This ensures that the reconstructed/generated image does not depend on the ordering of the posterior/prior marginals.

The \textbf{refinement network} simply consists of feed-forward and recurrent layers with weights tied across $K$ that are again applied individually to each element of the set of posterior parameters $\bm{\lambda}$.

\subsection{Invariance}

Due to the use of Slot Attention's scaled dot-product set attention mechanism to process the inputs in each stochastic layer of the posterior, EfficientMORL inherits the property that it is invariant to permutations applied to the order of the $x_1,\dots,x_N$, $x_n \in \mathbb{R}^D$.
An inspection of Line 11 of Algorithm~\ref{alg:inference} verifies that the $N$-element input $x$ is aggregated with a  permutation invariant summation over $N$.

\section{EfficientMORL vs. GENESIS}
\label{sec:app:genesis}
In this section, we provide arguments as to why models that learn \emph{unordered} (i.e., permutation equivariant) latents might be preferred over those that learn \emph{ordered} latents.
Many of the discussion points here are adapted from Appendix A.3 of~\citet{pmlr-v97-greff19a}. 
We also empirically show one key advantage using a variant of the Multi-dSprites dataset.

GENESIS~\citep{engelcke2019genesis} is a generative model for multi-object representation learning that, like MONet, uses sequential attention over the image to discover objects. 
Its design is motivated by efficiency and the task of unconditional scene generation.
It uses an autoregressive prior and autoregressive posterior to induce an ordering on its $K$ object-centric latent variables which enables fast inference and generation. 
The authors demonstrate that the autoregressive prior also facilitates coherent generation of multi-object scenes. 

\textbf{On biased decomposition} However, GENESIS is not equivariant to permutations applied to the $K$ latents.
That is, GENESIS infers a \emph{ordered} scene decomposition.
Due to the autoregressive prior and posterior, the pixels assigned to the $i$\textsuperscript{th} latent depends on latents $<i$, which biases the model towards specific decomposition strategies that incorporate global scene information (Figure~\ref{fig:app:emorl-genesis}).
GENESIS uses a deterministic stick-breaking process to implement sequential attention, which encourages it to learn fixed strategies such as always placing the background in the first or last component.
Additionally, this can occasionally lead the model towards learning poor strategies that are overly dependent on color.

\textbf{On ambiguity} Only updating each object-centric latent once makes GENESIS potentially less capable of handling ambiguous and complex cases. 
EfficientMORL assigns pixels to latents across multiple iterations, which can facilitate disambiguating parts of a scene that are difficult to parse.

\textbf{On multi-stability} GENESIS also does not share the multi-stability property enjoyed by EfficientMORL and IODINE.
That is, running inference multiple times with EfficientMORL can produce different scene decompositions, particularly for ambiguous cases, due to randomness in the iterative assignment (see Figure 10 of~\citet{pmlr-v97-greff19a}).  

\textbf{On sequential extensions} Finally, we highlight that IODINE has been demonstrated to be straightforward to extend to sequences, and we expect EfficientMORL could be similarly applied to sequences since it shares with IODINE the use of adaptive top-down refinement.
This has been accomplished in a few different ways. 
One way is to simply pass a new image at each step of iterative inference~\cite{pmlr-v97-greff19a}.
Or, initialize the posterior at each time step with the posterior from the previous time step before performing $I$ refinement steps~\citep{nanbo2020learning}.
In the model-based reinforcement learning setting, a latent dynamics model has been used to predict the initial posterior at each time step using the posterior from the previous time step, followed by $I$ refinement steps~\cite{veerapaneni2019entity}. 
Ordered models have been used for sequential data by adding an extra matching step to align the object latents over time, resembling unsupervised multi-object tracking~\cite{watters2019cobra}.

\textbf{GENESIS-v2} We note that the authors of GENESIS recently released GENESIS-v2~\cite{engelcke2021genesis}, which appeared after the initial submission of this work.
We comment briefly on it here.
GENESIS-v2 appears to remove GENESIS's autoregressive posterior and adds a non-parametric \emph{randomized} clustering algorithm to obtain an ordered set of of attention masks.
These attention masks are then mapped in parallel to $K$ latents (GENESIS-v2 is able to adjust $K$ on-the-fly via early-stopping of the clustering algorithm).
Due to the random initialization of the clustering algorithm, the model cannot learn a fixed sequential decomposition strategy.
We highlight that GENESIS-v2 keeps GENESIS's autoregressive prior and that GENESIS-v2 is not equivariant with respect to permutations applied to the set of cluster seeds used to obtain attention masks.
Like GENESIS, GENESIS-v2 uses a stick-breaking process such that the last ($K$\textsuperscript{th}) component is assigned the remaining scope; this suggests that global information may still be leaking into the representations (see Figure~\ref{fig:app:emorl-genesis}).
The authors note that on some training runs, the model learns to always place the background into the last component.

\subsection{Multi-colored and textured Multi-dSprites}
\label{sec:app:multicolored}
\begin{figure}[hbtp]
    \centering
    \begin{subfigure}[t]{\textwidth}
        \centering
         \includegraphics[scale=0.35]{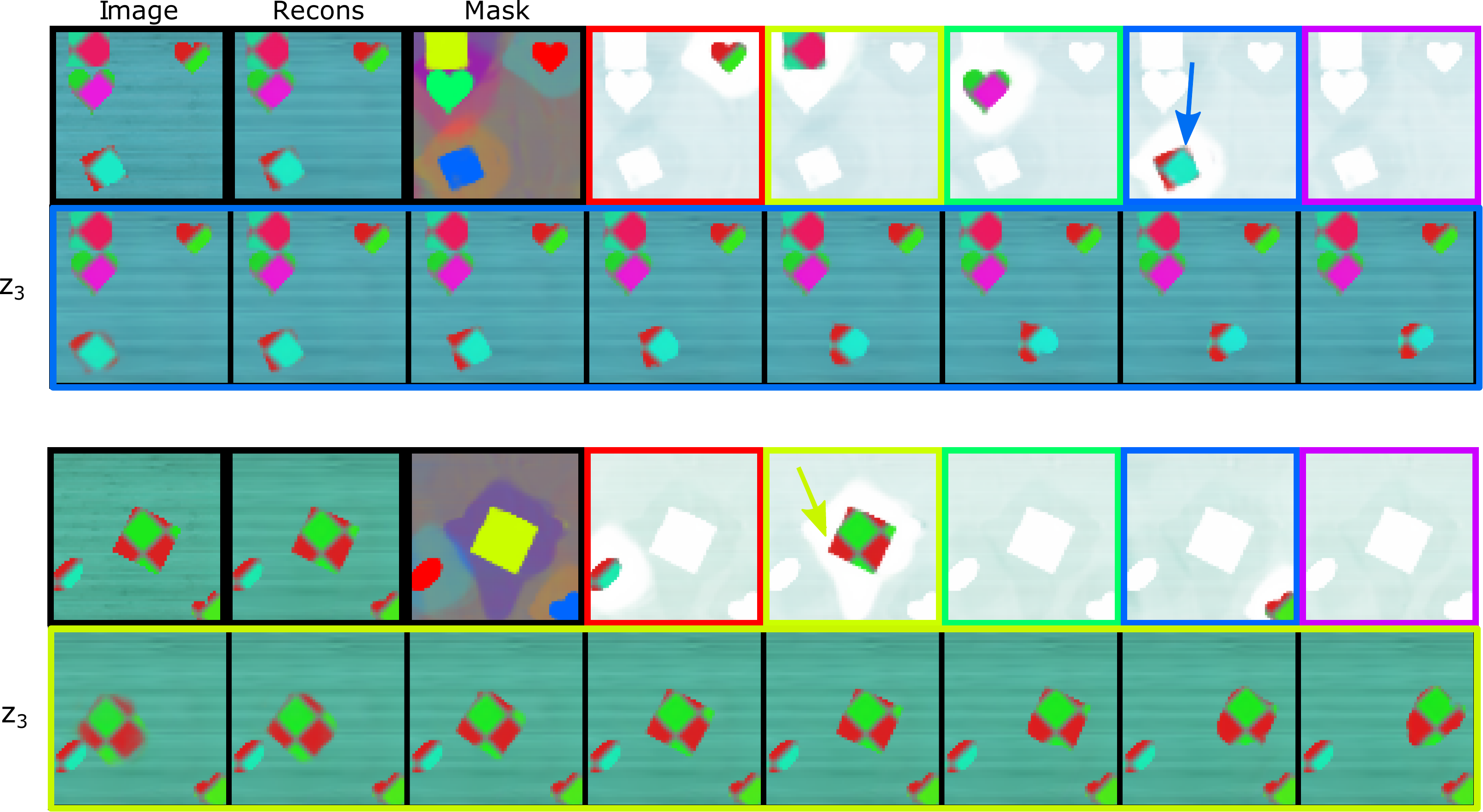}
    \caption{EfficientMORL\label{fig:app:textured-emorl}}
    \end{subfigure}
    \begin{subfigure}[t]{\textwidth}
       \centering
        \includegraphics[scale=0.35]{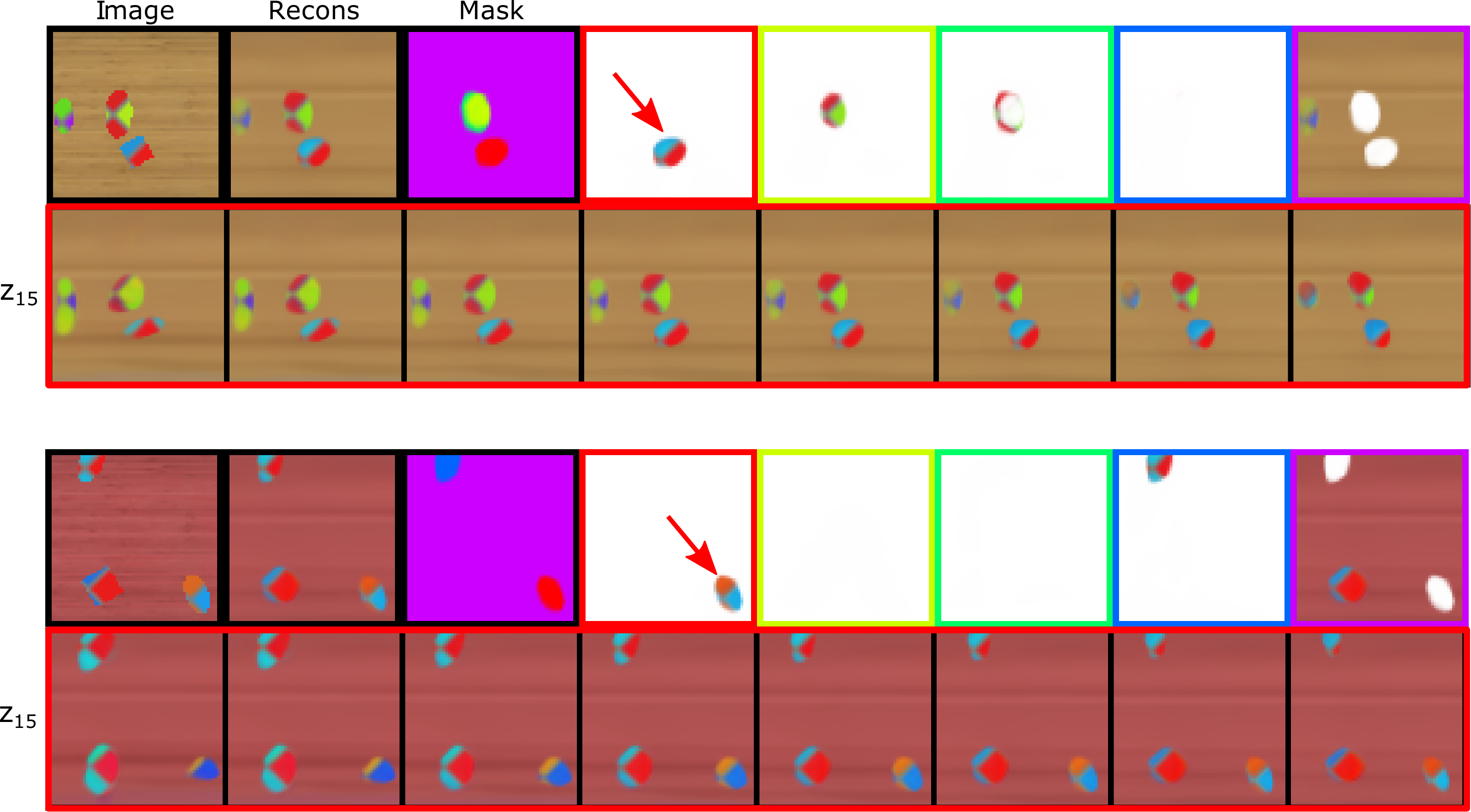}
        \caption{GENESIS-S\label{fig:app:textured-genesis-s}} 
    \end{subfigure}
    \begin{subfigure}[t]{\textwidth}
       \centering
        \includegraphics[scale=0.35]{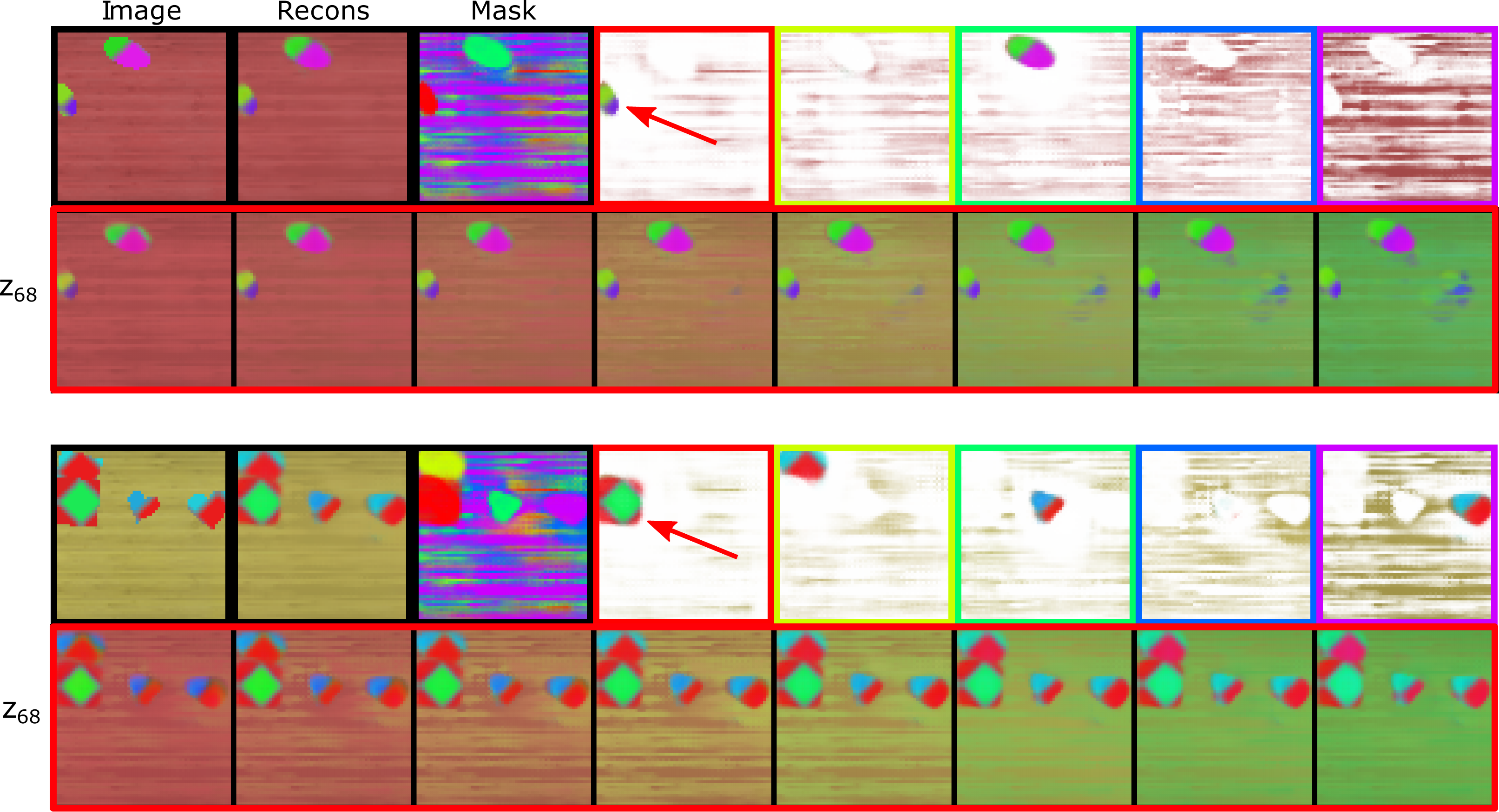}
        \caption{GENESIS\label{fig:app:textured-genesis}} 
    \end{subfigure}
    \caption{\textbf{The \emph{ordered} scene decomposition learned by GENESIS leaks global scene information into the object representations}. The second row of each sub-figure shows the reconstructed images generated by varying a single active latent dimension (chosen arbitrarily) of a single component (indicated by color). We show that for GENESIS-S and GENESIS, which have autoregressive posteriors, \emph{all} objects are affected when the first component is manipulated\label{fig:app:emorl-genesis}.}
\end{figure}
We illustrate how global scene-level information leaks into the object-centric representations in GENESIS in Figure~\ref{fig:app:emorl-genesis}.
For this, we created a variant of the Multi-dSprites dataset that has as background a simple uniform texture and foreground objects that are multi-colored with a distinctive pattern. 
Models that successfully decompose these images cannot do so by only grouping using color cues.

We invested considerable effort into tuning GENESIS to solve this dataset.
What worked was using the Gaussian image likelihood with $\sigma = 0.1$ (EfficientMORL uses the same) instead of the Gaussian mixture model (see Section~\ref{sec:app:implementation}; we tried various values for $\sigma$ for the mixture model likelihood as well) and decreasing GENESIS's GECO parameter update rate to avoid increasing the KL too much early on.
The Gaussian mixture model likelihood consistently caused GENESIS to learn to segment the images purely based on color.
We trained both GENESIS and GENESIS-S models, where GENESIS-S uses a single set of latents instead of splitting them into masks and components $[\mathbf{z}^m, \mathbf{z}^c]$.
GENESIS converged significantly faster than GENESIS-S and obtained the best results, whereas the latter did not finish converging after $500$K steps.

As shown in Figure~\ref{fig:app:emorl-genesis}, the object-centric representations learned by EfficientMORL are able to be independently manipulated.
GENESIS's autoregressive posterior causes it to incorporate global scene-level information into the representations.
We show that manipulating a single dimension of a single object latent can change the entire scene representation.

\section{Limitations}
\label{sec:app:lim}

For a given image, EfficientMORL requires that $K$ is chosen \emph{a priori}, although any value of $K$ can be chosen.
A way to adapt $K$ automatically based on the input could be useful to increase the number of latent variables dynamically if needed.

Related to this, EfficientMORL does not have any explicit mechanism for attending to a subset of objects in a scene.
This could prove useful for handling scenes containing many objects and scenes where what constitutes the foreground and background is ambiguous.

We do not expect EfficientMORL to be able to generate \emph{globally coherent} scenes due to the independence assumption in the prior. 
Identifying how to achieve coherent generation without sacrificing permutation equivariance is an open problem.

EfficientMORL also inherits similar limitations to IODINE related to handling images containing heavily textured \emph{backgrounds} (see Section 5 of~\citet{pmlr-v97-greff19a}), as well as large-scale datasets that do not consist of images that represent a dense sampling of the data generating latent factors, which is important for unsupervised learning.
EfficientMORL \emph{can} handle textured and multi-colored \emph{foreground} objects assuming a simplistic background (Figure~\ref{fig:app:textured-emorl}).

\section{Additional ablation studies} 
\label{sec:app:ablations}
\begin{table*}[th]
    \centering
    \small
    \caption{Ablation study results. All models use reversed prior++ (Figure~\ref{fig:app:reversed_prior_plus}) unless specified.}
    \begin{tabular}{lcccc}
        \toprule
         & Env  & ARI & MSE ($\times 10^{-4}$) & KL \\
        \hline
        EfficientMORL w/out DualGRU & CLEVR6 & $79.7 \pm 22.3 $ & $\mathbf{8.1 \pm 1.7}$ & $1357.4 \pm 321.6$ \\
        EfficientMORL w/out GECO & CLEVR6 & $79.9 \pm 9.9$ & $9.9 \pm 2.5$ & $\mathbf{177.5 \pm 14.7}$\\
        EfficientMORL& CLEVR6 & $\mathbf{82.4 \pm 7.9}$ & $\mathbf{8.3 \pm 1.1}$ & $692.7 \pm 281.4$\\
        \hline
        EfficientMORL w/out GECO & Tetrominoes & $82.2 \pm 17.7$ & $68.8 \pm 42.2$ & $\mathbf{45.9 \pm 9.3}$ \\
        \hline
        EfficientMORL w/ bottom-up prior & Tetrominoes & $85.7 \pm 11.0$ & $17.1 \pm 14.6$ & -\\
        EfficientMORL w/ reversed prior & Tetrominoes & $86.9 \pm 16.2$ &  $22.1 \pm 21.4$ & - \\
        \hline
        EfficientMORL & Tetrominoes &  $\mathbf{97.9 \pm 2.4}$ & $\mathbf{7.0 \pm 6.0}$ & $98.5 \pm 21.4$  \\
        \hline
    \end{tabular}
    \label{tab:ablations}
\end{table*}

 Results are in Table~\ref{tab:ablations}. The CLEVR6 results are computed across $10$ random seeds for $50k$ steps.
 At this many steps the model has begun to converge and large differences in final performance can be easily observed.
 Each of the Tetrominoes results are computed on a validation set of $320$ images across $5$ random seeds.
     
\begin{figure}
    \centering
    \begin{subfigure}[t]{0.33\textwidth}
    \centering
    \includegraphics[scale=0.8]{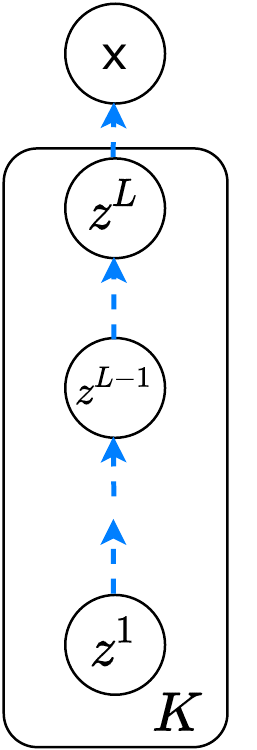} 
    \caption{Uses $p_\theta(\mathbf{z}^L | \mathbf{z}^{L-1})$ in refinement loss KL}
    \end{subfigure}
    \begin{subfigure}[t]{0.33\textwidth}
    \centering
    \includegraphics[scale=0.8]{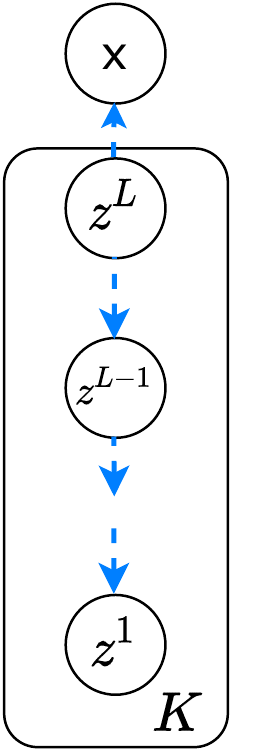} 
    \caption{Uses $p(\mathbf{z}^L)$ in refinement loss  KL}
    \end{subfigure}
    \begin{subfigure}[t]{0.33\textwidth}
    \centering
    \includegraphics[scale=0.8]{figures/appendix/EMORL_priorB.pdf} 
    \caption{Uses $p_\theta(\mathbf{z}^1 \mid \mathbf{z}^2)$ in refinement loss  KL\label{fig:app:reversed_prior_plus}}
    \end{subfigure}
    \caption{The three variants of the hierarchical prior we explore. a) \textbf{EfficientMORL w/ bottom-up prior}, where $p(\mathbf{z}^1)$ is a standard Gaussian and the prior gets more expressive in higher layers, b) \textbf{EfficientMORL w/ reversed prior} where $p(\mathbf{z}^L)$, the corresponding prior for the top layer posterior, is a standard Gaussian, and the prior gets more expressive in lower layers, and c) \textbf{EfficientMORL w/ reversed prior++} where $p(\mathbf{z}^L)$ is also standard Gaussian but the posteriors at lower layers of the hierarchy are encouraged to match the top layer posterior via the modified refinement KL term.\label{fig:app:priors}}
\end{figure}

\begin{figure}[hbtp]
    \centering
    \begin{subfigure}[t]{\textwidth}
    \centering
    \includegraphics[scale=0.54]{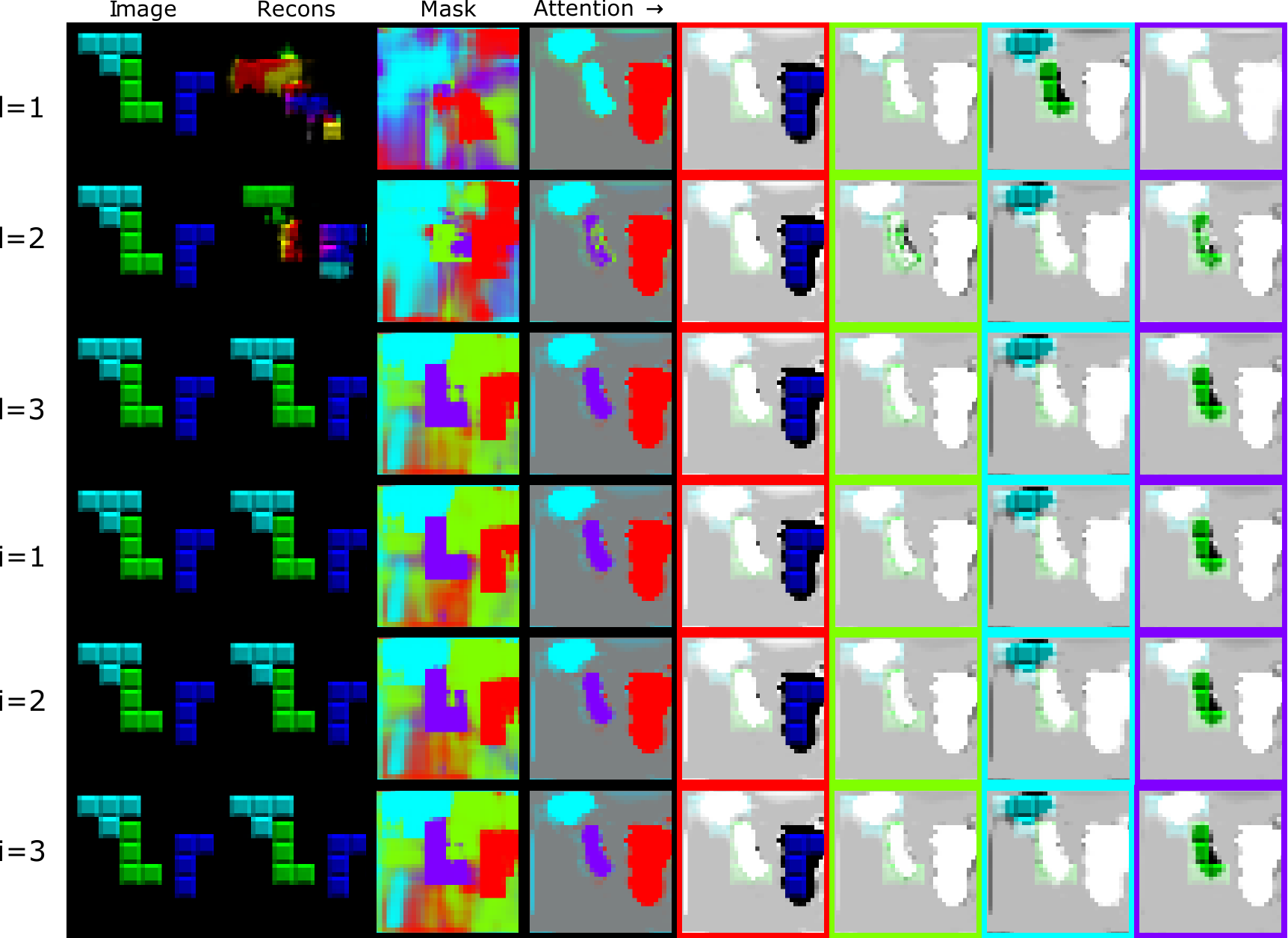} 
    \caption{EfficientMORL w/ bottom-up prior}
    \end{subfigure}
    \begin{subfigure}[t]{\textwidth}
    \centering
    \includegraphics[scale=0.54]{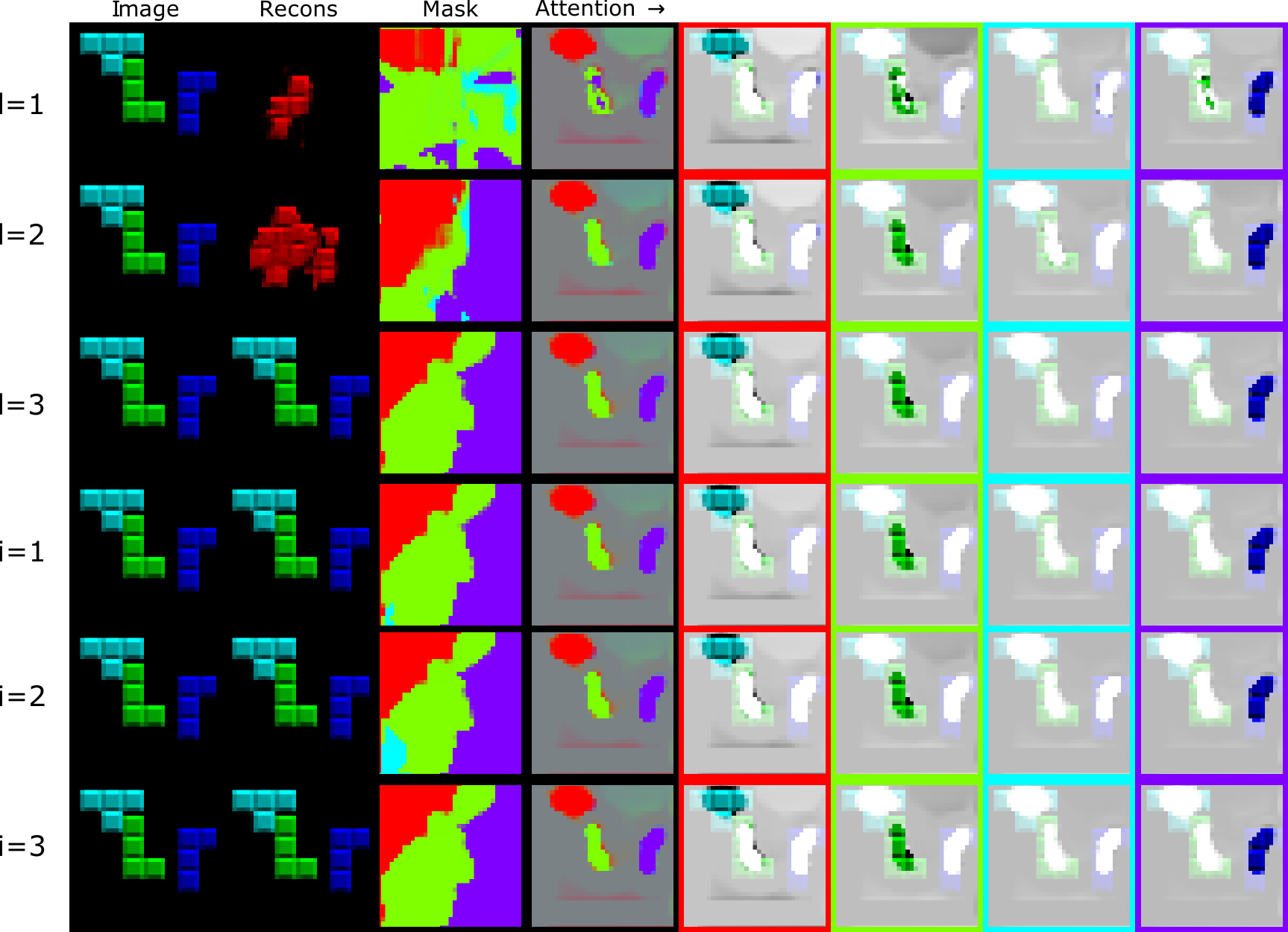} 
    \caption{EfficientMORL w/ reversed prior}
    \end{subfigure}
    \begin{subfigure}[t]{\textwidth}
    \centering
    \includegraphics[scale=0.54]{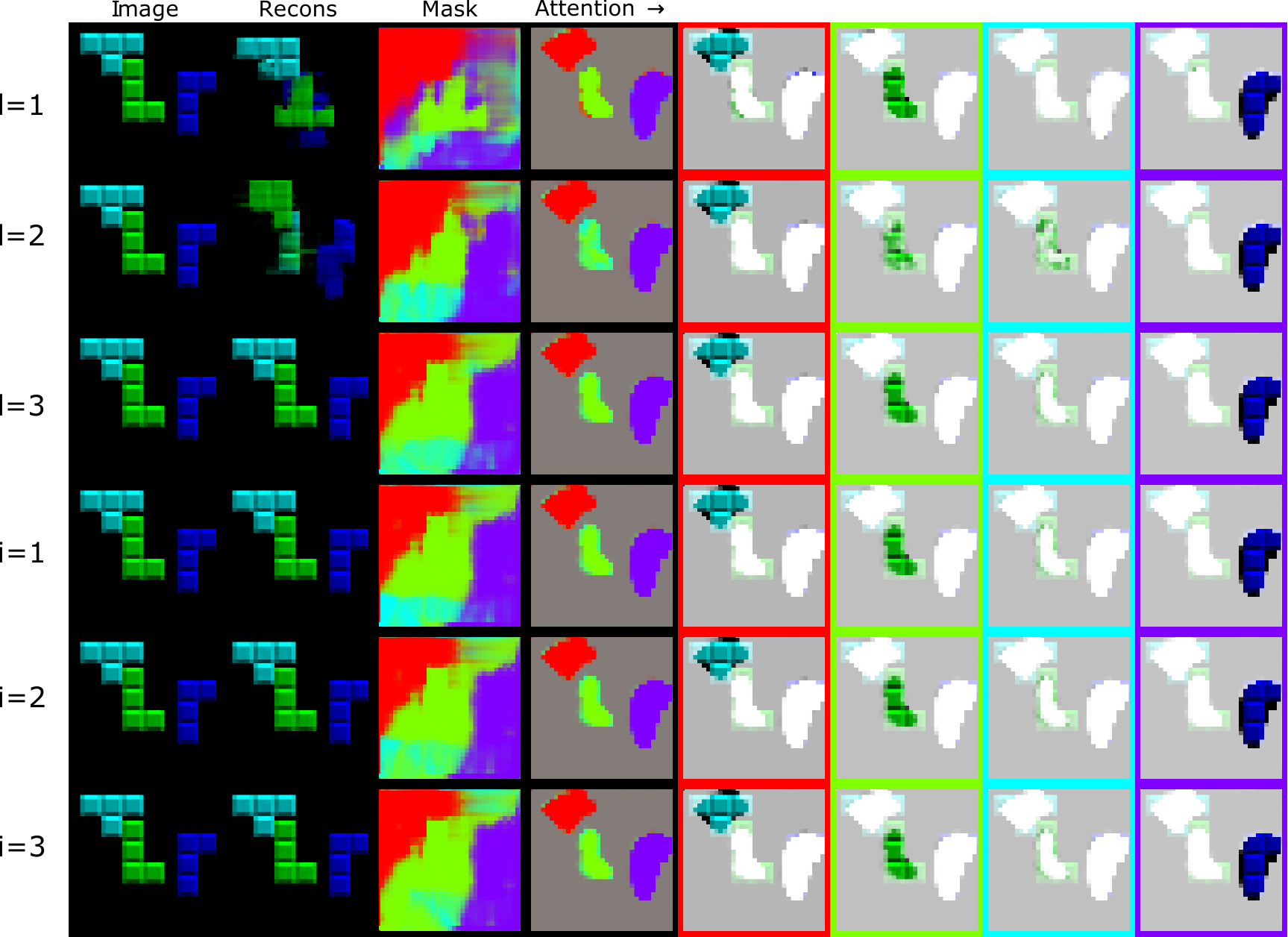} 
    \caption{EfficientMORL w/ reversed prior++}
    \end{subfigure}
    \caption{Samples from the intermediate posteriors and attention from the ablated hierarchical priors. The reconstructions in the first two rows demonstrate the efficacy of the reversed prior++ at preventing the intermediate posteriors from collapsing.\label{fig:app:priors-intermediate-posteriors}}
\end{figure}
\textbf{Hierarchical prior} 
The three hierarchical prior variants are shown in Figure~\ref{fig:app:priors}.
We present the ARI and MSE scores for Tetrominoes in Table~\ref{tab:ablations}, where the reversed prior++ consistently achieves the best results.
By inspecting samples from the intermediate posteriors of each model (Figure~\ref{fig:app:priors-intermediate-posteriors}, second column, first two rows), we can see that the bottom-up prior and reversed prior learn to leave the posterior at layers $1-2$ close to the initial uncorrelated Gaussian $q(\mathbf{z}^0)$, with the reversed prior exhibiting this most strongly.
This limits the expressiveness of the posterior at layer $L$ and on certain runs these models would converge to poor local minima---hence the high standard deviation in ARI and MSE.
On the other hand, the reversed prior++ keeps the posterior at layers $1-2$ close to the layer $L$ posterior.
The large gap in performance suggests that although Tetrominoes appears to be easy to solve, it may be deceptively challenging as it seems to require fitting a complex, non-Gaussian posterior.
Although we do not show the results here, we did try training EfficientMORL w/ reversed prior on CLEVR6 and found that the less expressive posterior was sufficient to achieve comparable results to the reversed prior++.

\textbf{DualGRU} 
We replaced the DualGRU with a standard GRU of hidden dimension $2D$. 
The DualGRU regularizes training due to improved parameter efficiency from the block-diagonal design.
With a single standard GRU, the model tends to achieve a high KL. 
Some of these models converged to poor local minima, which is reflected by the lower ARI score.

\textbf{GECO} We trained EfficientMORL without GECO to examine the severity of posterior collapse on CLEVR6 and Tetrominoes (Table~\ref{tab:ablations}).
We found that posterior collapse did not affect the model on Multi-dSprites and therefore did not use GECO for this environment.
Without GECO, the KL is low due to the early collapse of many of the posterior dimensions.
For most runs, this leads to poor quality reconstructions that achieve higher MSE and lower ARI scores.

\begin{figure}[ht]
    \centering
    \begin{subfigure}[t]{0.45\textwidth}
        \centering
        \includegraphics[scale=0.35]{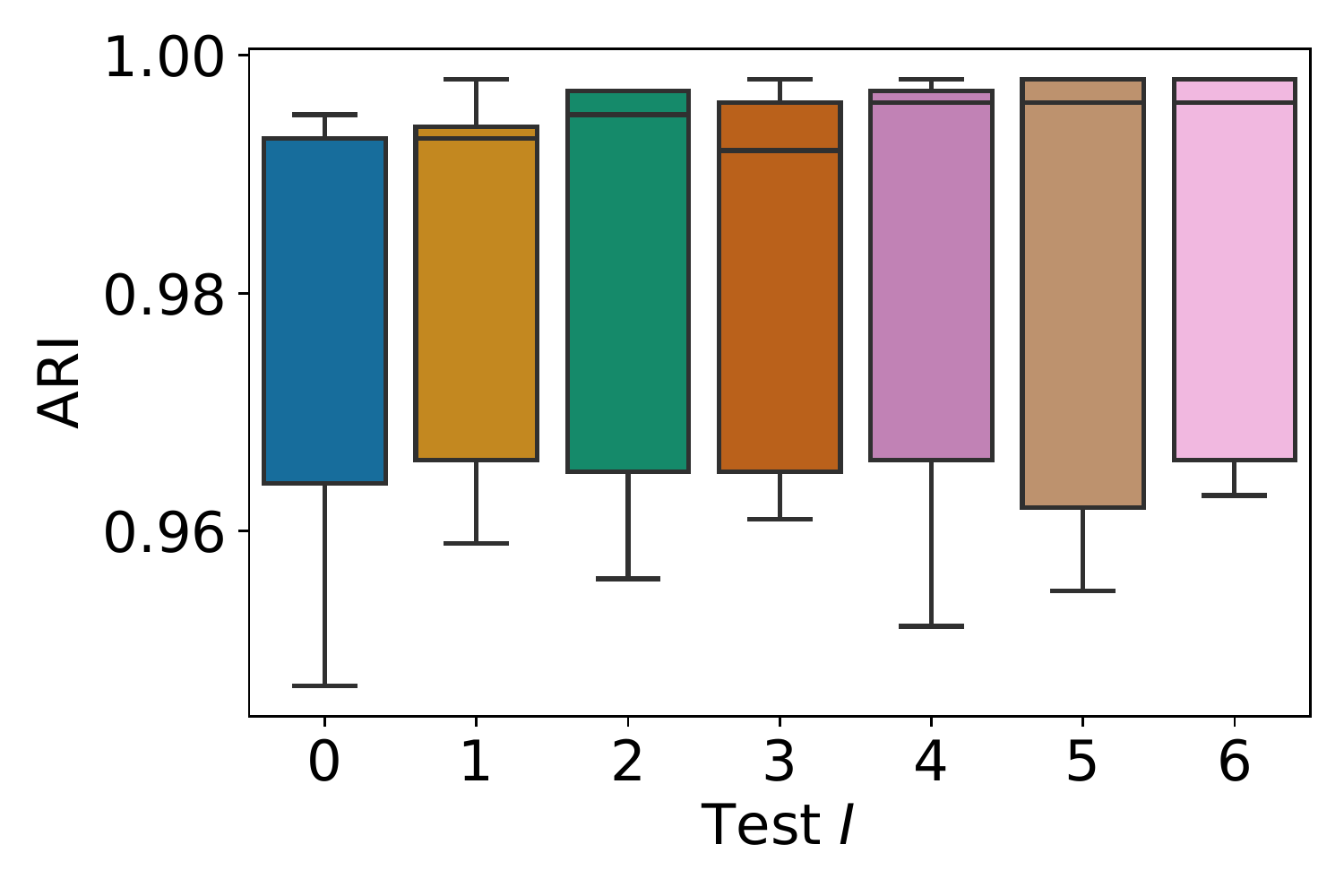}
        \caption{Tetrominoes (w/ reversed prior++) ARI vs. test $I$.\label{fig:app:Tetris_ARI_TD}}
    \end{subfigure}%
    \begin{subfigure}[t]{0.45\textwidth}
        \centering
        \includegraphics[scale=0.35]{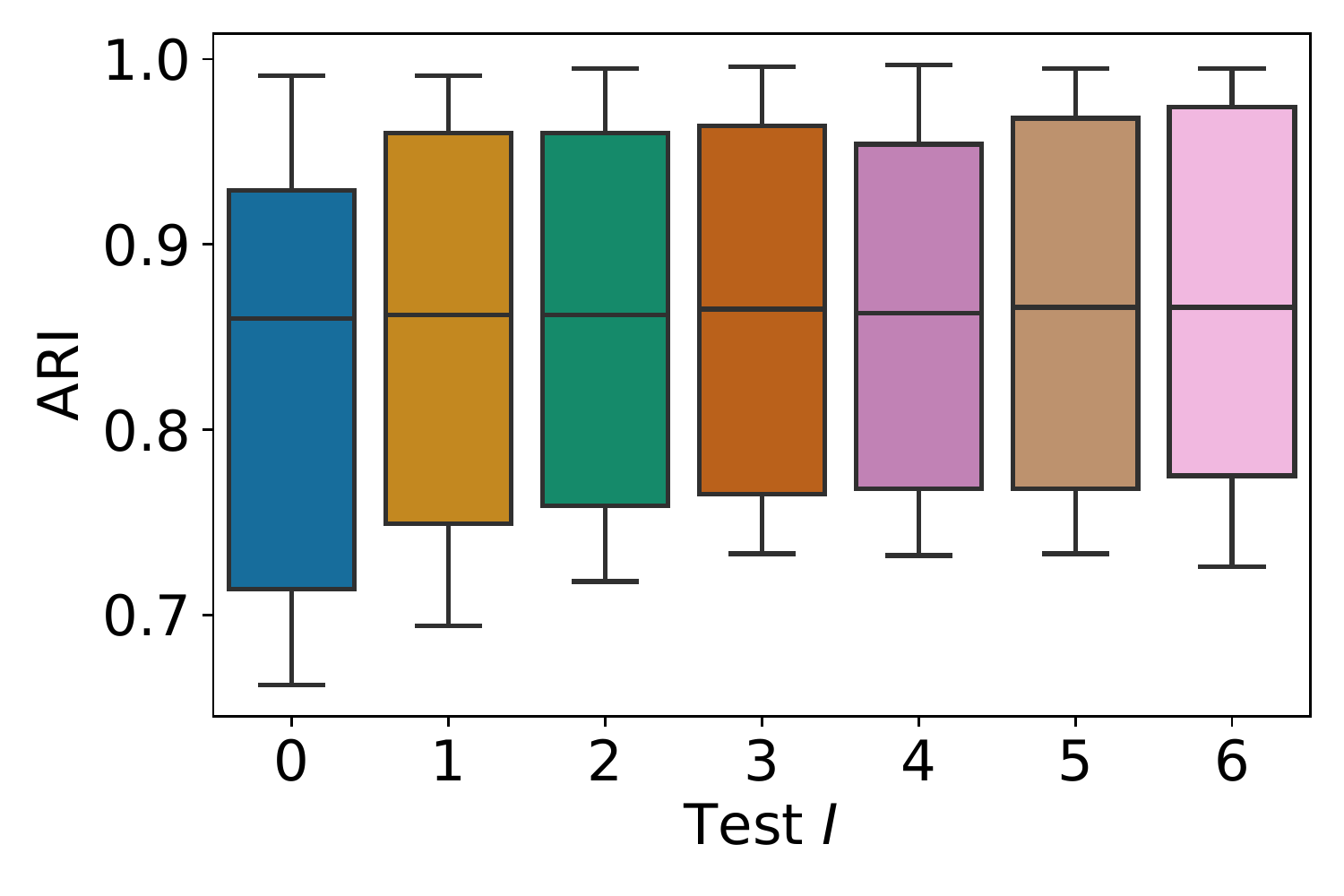}
        \caption{Tetrominoes (w/ bottom-up prior) ARI vs. test $I$.\label{fig:app:Tetris_ARI_BU}}
    \end{subfigure}
    \begin{subfigure}[t]{0.45\textwidth}
        \centering
        \includegraphics[scale=0.35]{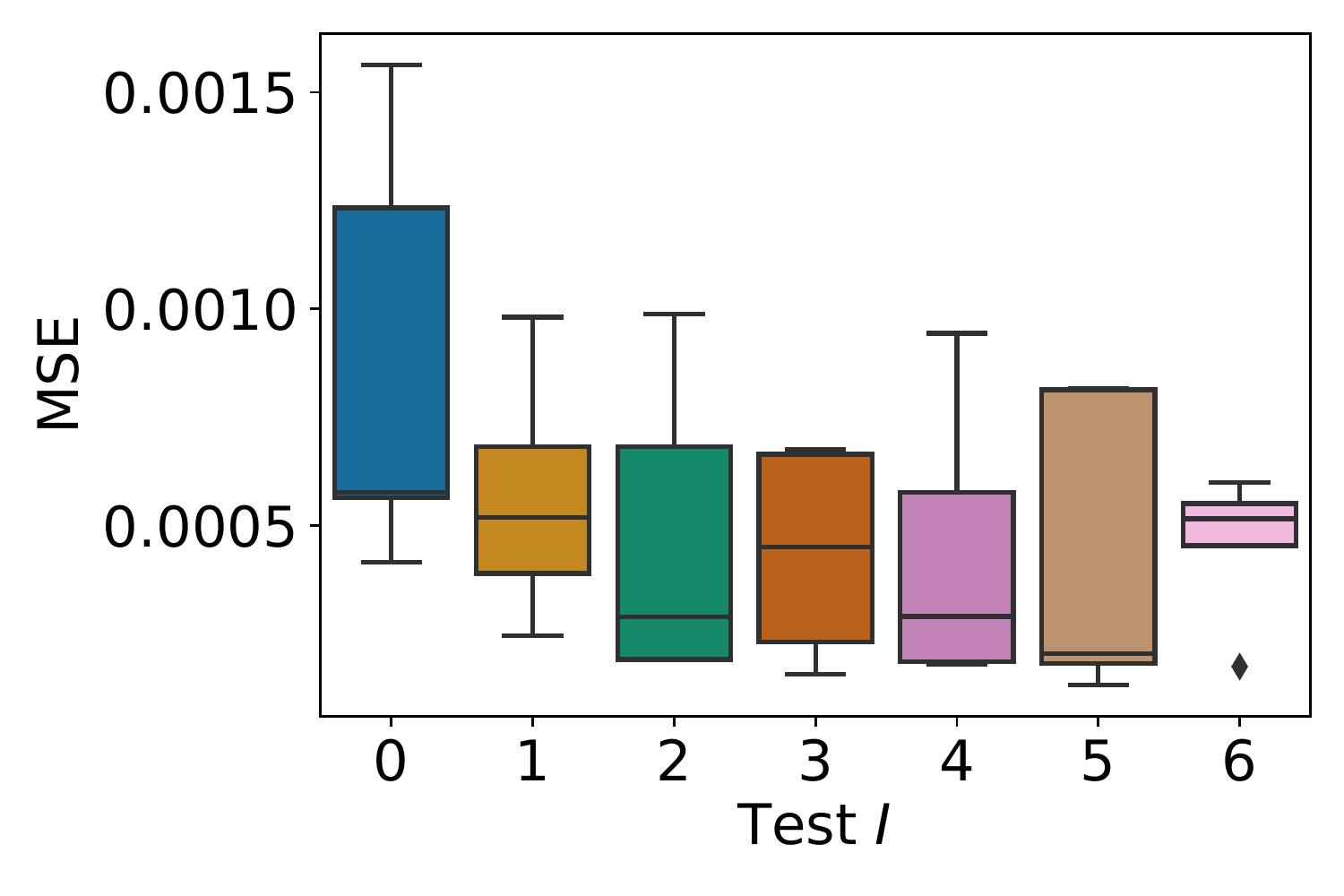}
        \caption{Tetrominoes (w/ reversed prior++) MSE vs. test $I$\label{fig:app:Tetris_MSE_TD}}
    \end{subfigure}%
    \begin{subfigure}[t]{0.45\textwidth}
        \centering
        \includegraphics[scale=0.35]{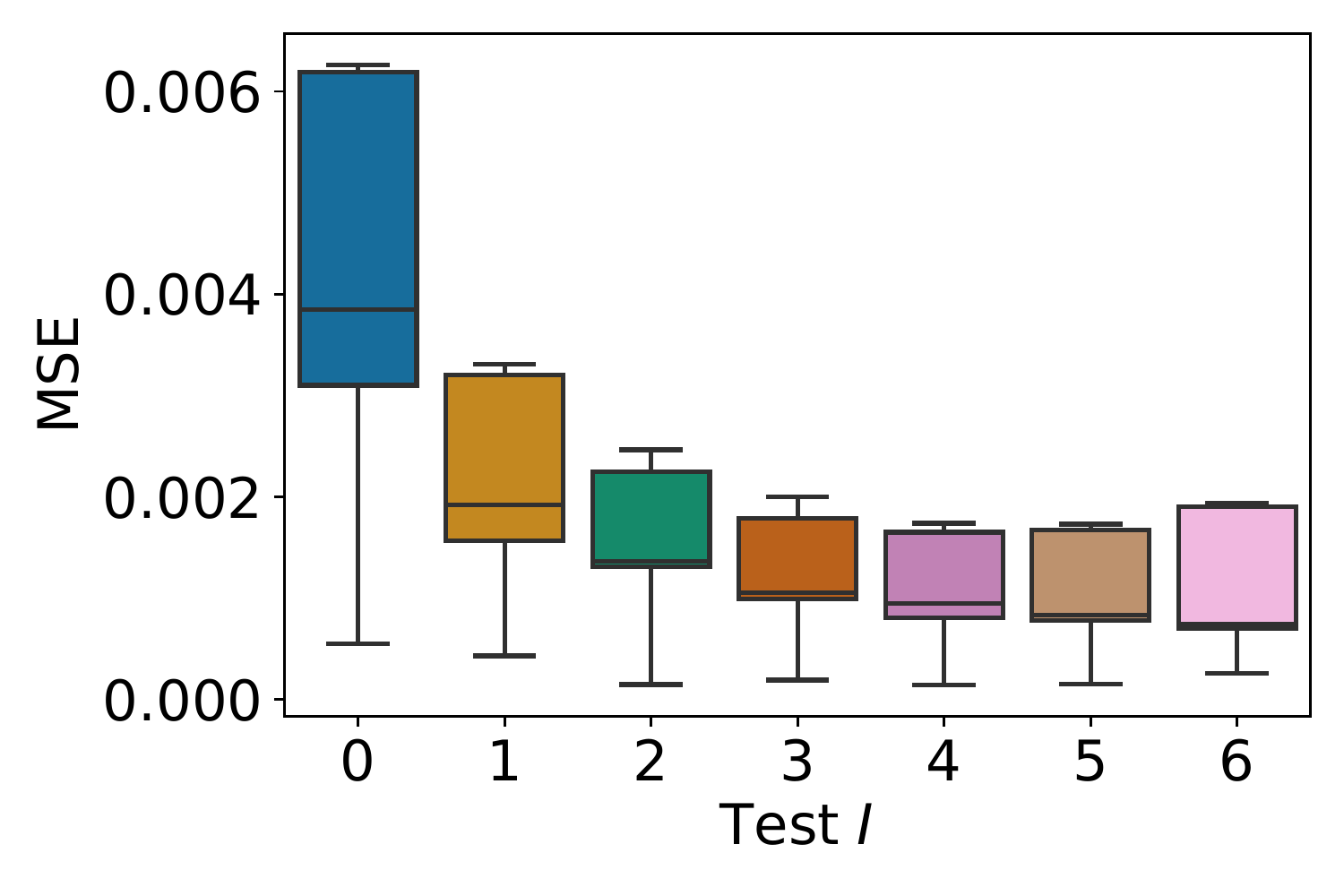}
        \caption{Tetrominoes (w/ bottom-up prior) MSE vs. test $I$\label{fig:app:Tetris_MSE_BU}}
    \end{subfigure}
    \begin{subfigure}[t]{0.45\textwidth}
        \centering
        \includegraphics[scale=0.35]{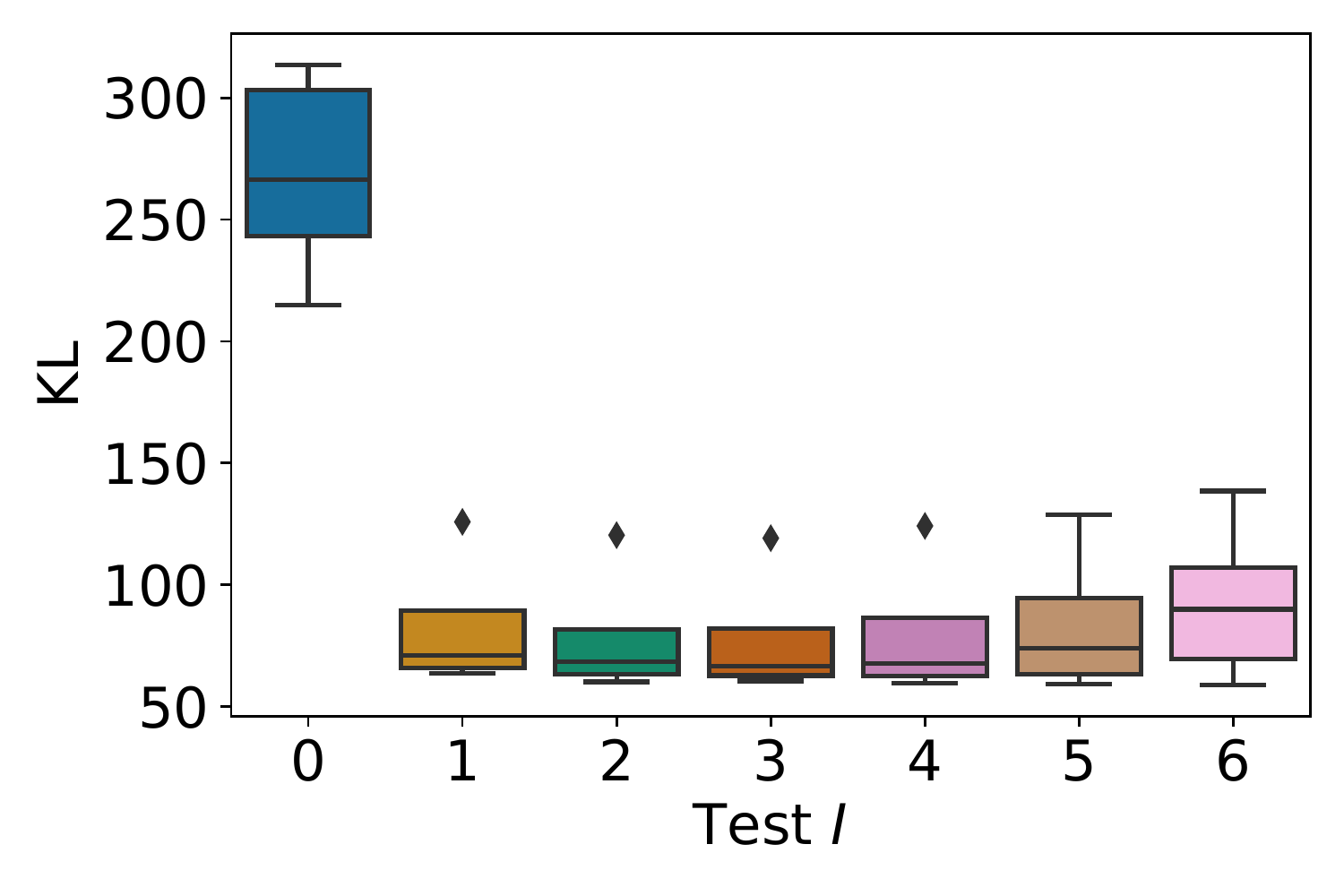}
        \caption{Tetrominoes (w/ reversed prior++) KL vs. test $I$\label{fig:app:Tetris_KL_TD}}
    \end{subfigure}%
    \begin{subfigure}[t]{0.45\textwidth}
        \centering
        \includegraphics[scale=0.35]{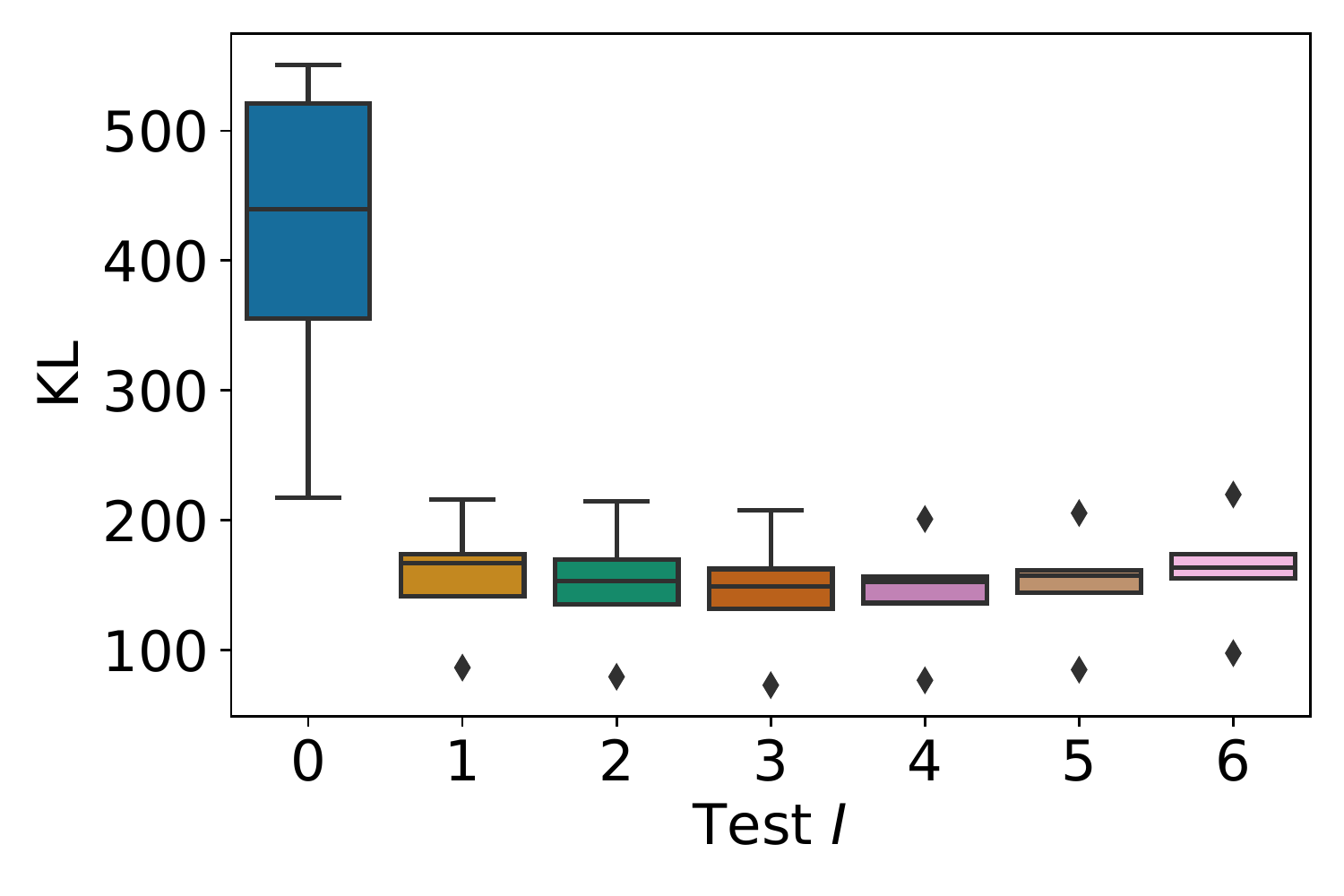}
        \caption{Tetrominoes (w/ bottom-up prior) KL vs. test $I$\label{fig:app:Tetris_KL_BU}}
    \end{subfigure}    
    \caption{a,c) Training with $I=3$ fixed and testing with $I=0$ results in a fractional drop in ARI/MSE with the reversed prior++. A larger gap in KL is again observed between zero and one test refine step, for both the reversed prior++ (e) and bottom-up prior (f). Performance is maintained when increasing $I$ up to 6, as the refinement GRU has learned to exploit sequential information.\label{fig:app:Tetris-varying-I}}
\end{figure}

\textbf{Varying refinement steps} We show the test ARI, MSE, and KL for various values of $I$ using the reversed prior++ and bottom-up prior Tetrominoes models (Figure~\ref{fig:app:Tetris-varying-I}).
Recall that we held $I$ fixed at three when training on Tetrominoes.
The ARI/MSE shows a slight drop when testing with zero refine steps, whereas the KL is noticeably larger at $I=0$.
A single low-dimensional refinement step does not incur a large increase in extra computation (Figure~\ref{fig:efficiency}) if the low KL posterior is desired at test time.

\section{Additional results}
\label{sec:app:results}
\subsection{Object decomposition}
\begin{figure}[ht]
    \centering
    \includegraphics[scale=0.8]{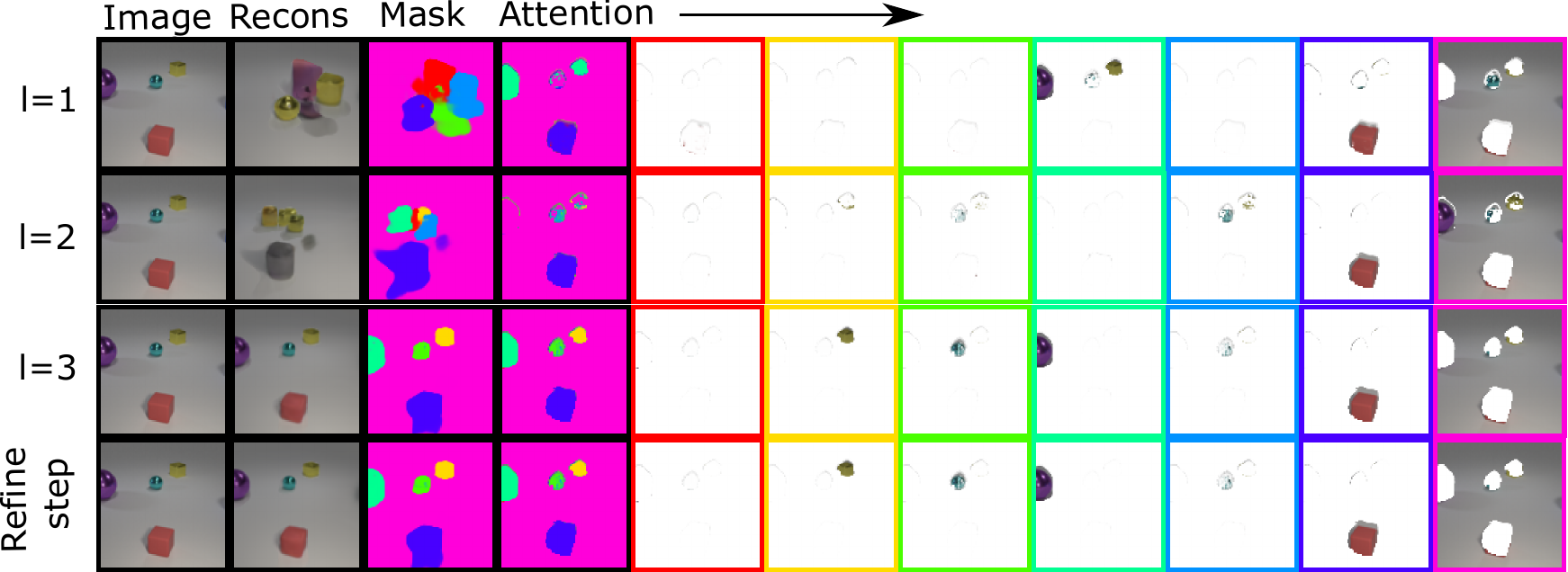}
    \caption{Visualization of reconstructions, masks, and attention from a single sample from the intermediate and final posteriors on CLEVR6.\label{fig:app:intermediate-posteriors-clevr6}}
\end{figure}

An extra visualization of a sample from the intermediate and final posteriors for a single scene, showing reconstruction, masks, and attention, is provided in Figure~\ref{fig:app:intermediate-posteriors-clevr6}.\footnote{Visualizations are made using a modified script provided by~\citet{pmlr-v97-greff19a}.} 

\begin{figure}[hbtp]
    \newcommand{\CLEVRscale}{0.4}
    \centering
    \begin{subfigure}[t]{\textwidth}
        \centering
        \includegraphics[trim=4cm 0cm 4cm 0cm,clip=True,scale=\CLEVRscale]{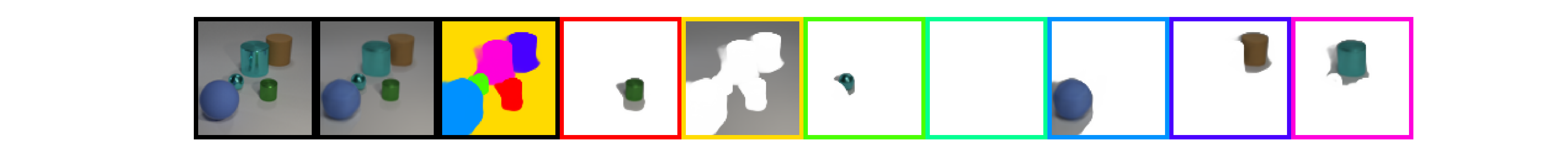}
    \end{subfigure}
    \begin{subfigure}[t]{\textwidth}
        \centering
        \includegraphics[trim=4cm 0cm 4cm 0cm,clip=True,scale=\CLEVRscale]{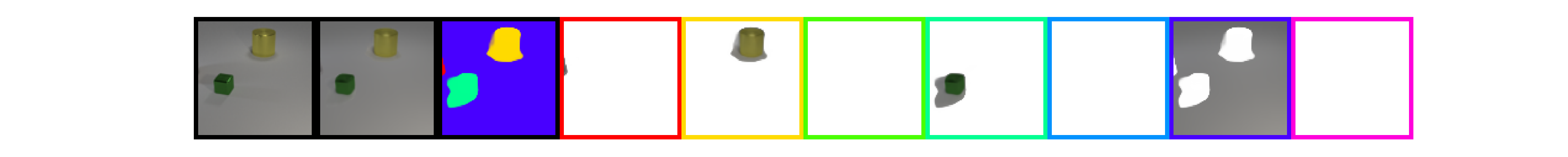}
    \end{subfigure}
    \begin{subfigure}[t]{\textwidth}
        \centering
        \includegraphics[trim=4cm 0cm 4cm 0cm,clip=True,scale=\CLEVRscale]{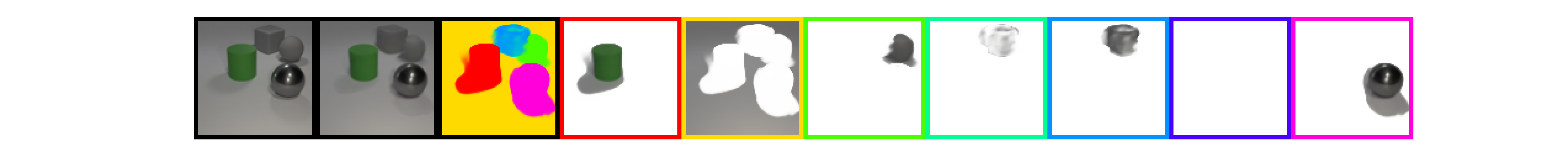}
    \end{subfigure}
    \begin{subfigure}[t]{\textwidth}
        \centering
        \includegraphics[trim=4cm 0cm 4cm 0cm,clip=True,scale=\CLEVRscale]{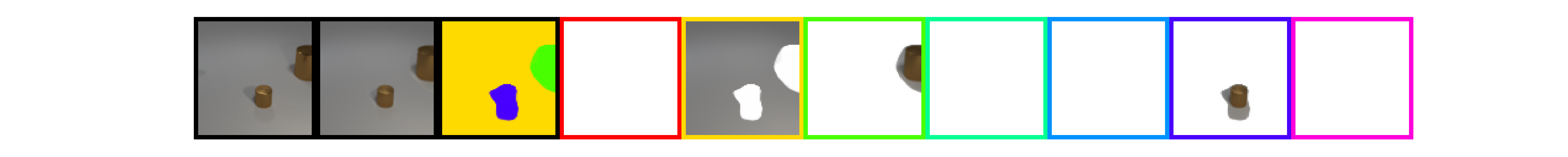}
    \end{subfigure}
    \begin{subfigure}[t]{\textwidth}
        \centering
        \includegraphics[trim=4cm 0cm 4cm 0cm,clip=True,scale=\CLEVRscale]{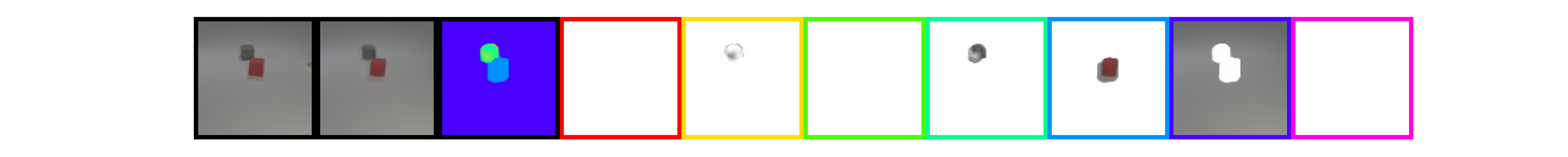}
    \end{subfigure}
    \begin{subfigure}[t]{\textwidth}
        \centering
        \includegraphics[trim=4cm 0cm 4cm 0cm,clip=True,scale=\CLEVRscale]{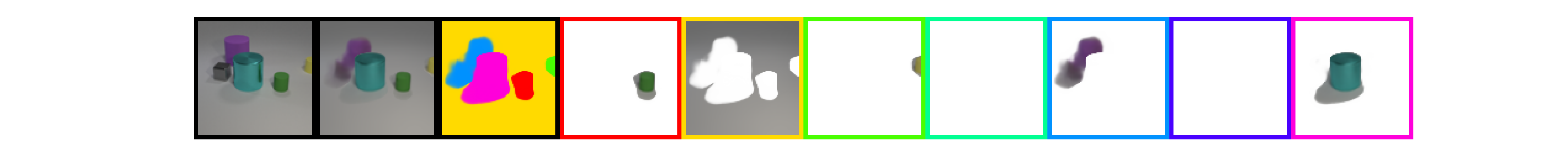}
    \end{subfigure}
    \begin{subfigure}[t]{\textwidth}
        \centering
        \includegraphics[trim=4cm 0cm 4cm 0cm,clip=True,scale=\CLEVRscale]{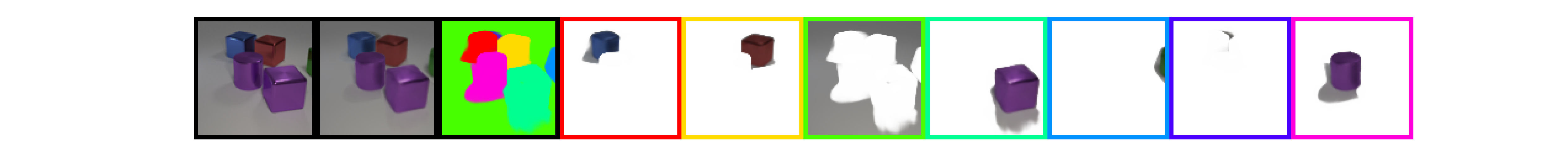}
    \end{subfigure}
    \begin{subfigure}[t]{\textwidth}
        \centering
        \includegraphics[trim=4cm 0cm 4cm 0cm,clip=True,scale=\CLEVRscale]{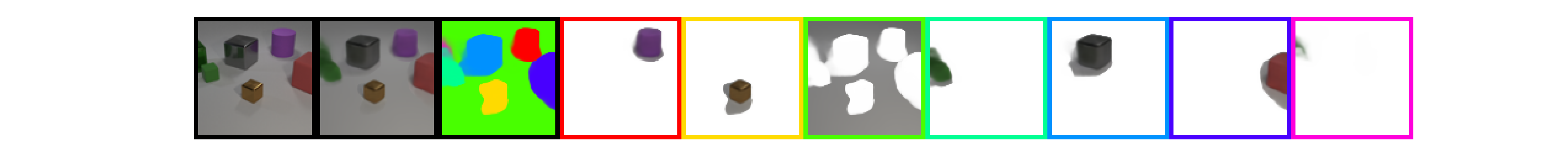}
    \end{subfigure}
    \caption{Additional CLEVR6 scene decompositions with $K=7,L=3$, and $I=1$. First column is the ground truth image, second column is the reconstruction, third column is the mask, and the remaining columns are the masked components. 
    One example of a failure case is shown in row 6 where the model joins a purple cylinder and an adjacent silver cube.\label{fig:app:clevr6_decompositions}}
\end{figure}

\begin{figure}[hbtp]
    \newcommand{\MultidSpritesScale}{0.4}
    \centering
    \begin{subfigure}[t]{\textwidth}
        \centering
        \includegraphics[trim=4cm 0cm 4cm 0cm,clip=True,scale=\MultidSpritesScale]{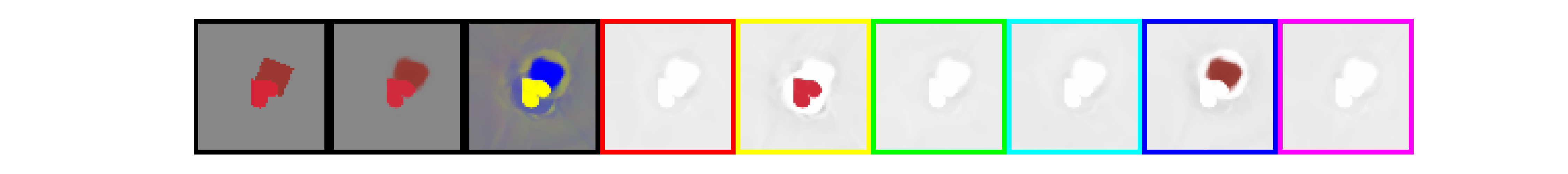}
    \end{subfigure}
    \begin{subfigure}[t]{\textwidth}
        \centering
        \includegraphics[trim=4cm 0cm 4cm 0cm,clip=True,scale=\MultidSpritesScale]{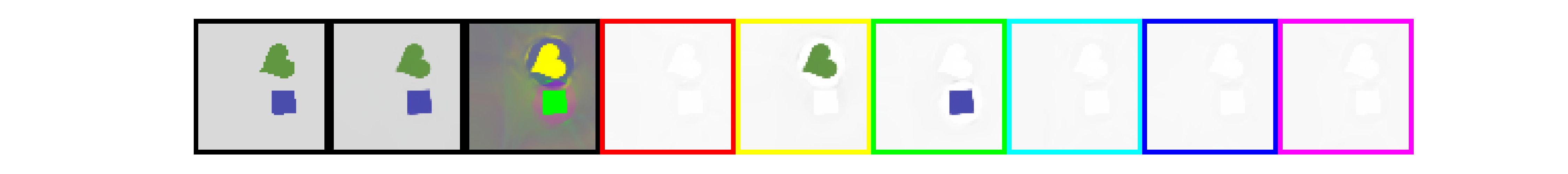}
    \end{subfigure}
    \begin{subfigure}[t]{\textwidth}
        \centering
        \includegraphics[trim=4cm 0cm 4cm 0cm,clip=True,scale=\MultidSpritesScale]{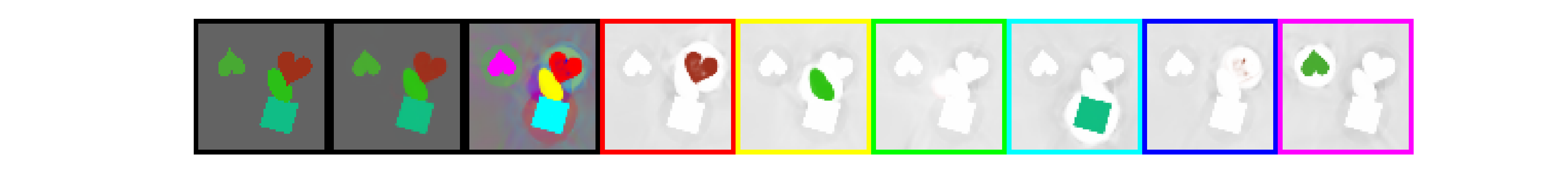}
    \end{subfigure}
    \begin{subfigure}[t]{\textwidth}
        \centering
        \includegraphics[trim=4cm 0cm 4cm 0cm,clip=True,scale=\MultidSpritesScale]{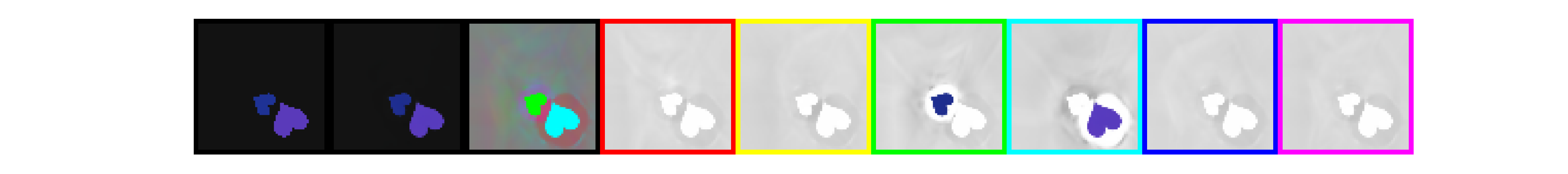}
    \end{subfigure}
    \begin{subfigure}[t]{\textwidth}
        \centering
        \includegraphics[trim=4cm 0cm 4cm 0cm,clip=True,scale=\MultidSpritesScale]{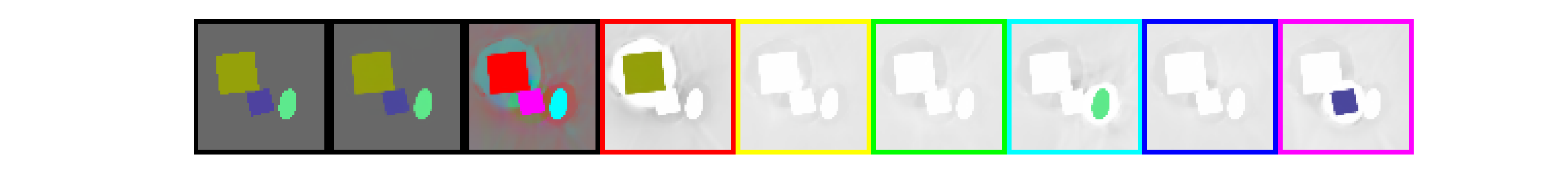}
    \end{subfigure}
    \begin{subfigure}[t]{\textwidth}
        \centering
        \includegraphics[trim=4cm 0cm 4cm 0cm,clip=True,scale=\MultidSpritesScale]{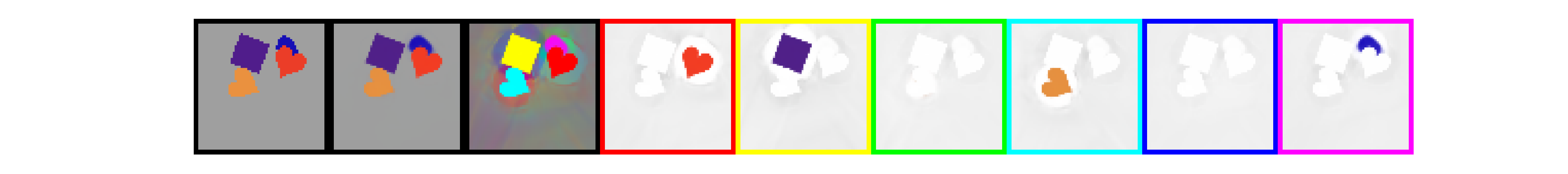}
    \end{subfigure}
    \begin{subfigure}[t]{\textwidth}
        \centering
        \includegraphics[trim=4cm 0cm 4cm 0cm,clip=True,scale=\MultidSpritesScale]{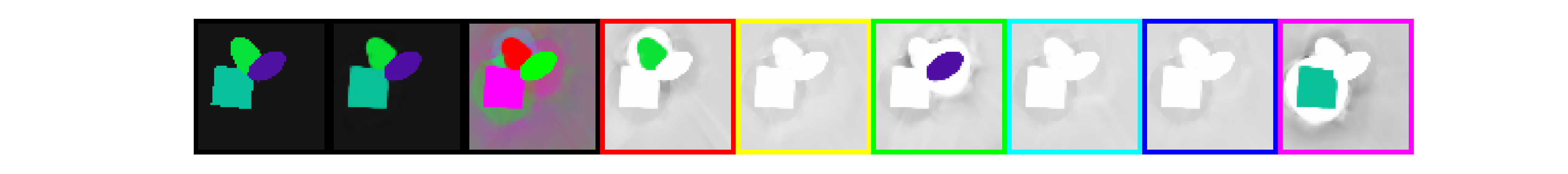}
    \end{subfigure}
    \begin{subfigure}[t]{\textwidth}
        \centering
        \includegraphics[trim=4cm 0cm 4cm 0cm,clip=True,scale=\MultidSpritesScale]{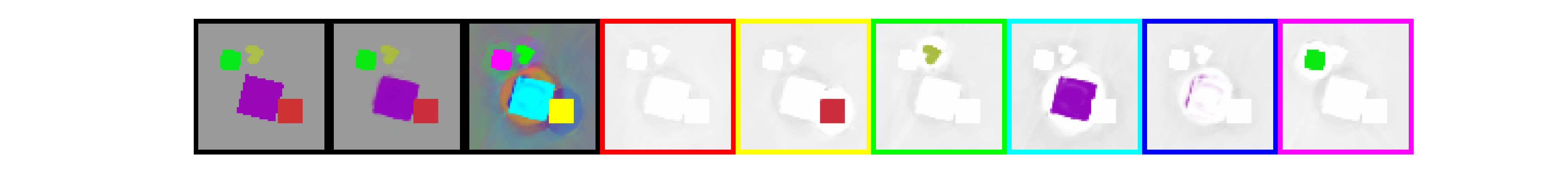}
    \end{subfigure}
    \caption{Additional Multi-dSprites scene decompositions with $K=6,L=3$, and $I=1$. First column is the ground truth image, second column is the reconstruction, third column is the mask, and the remaining columns are the masked components. \label{fig:app:multi_dSprites_decompositions}}
\end{figure}

\begin{figure}[hbtp]
    \centering
    \begin{subfigure}[t]{\textwidth}
        \centering
        \includegraphics[trim=2cm 0cm 2cm 0cm,clip=True,scale=0.45]{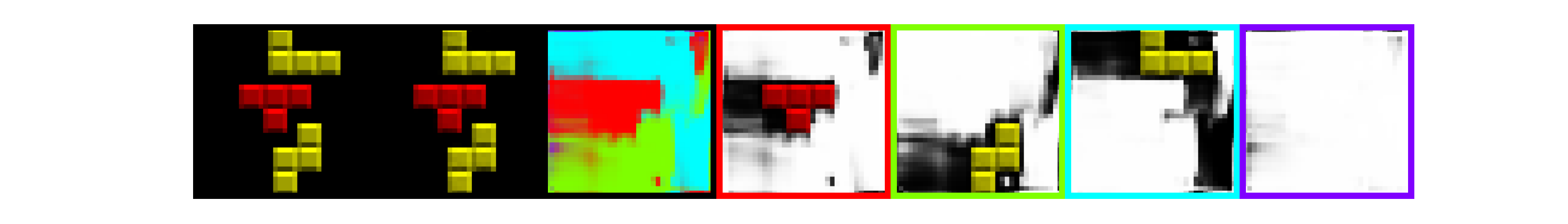}
    \end{subfigure}
    \begin{subfigure}[t]{\textwidth}
        \centering
        \includegraphics[trim=2cm 0cm 2cm 0cm,clip=True,scale=0.45]{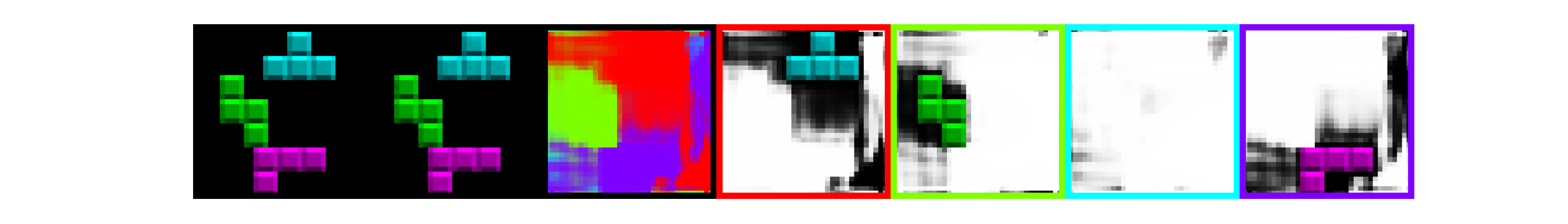}
    \end{subfigure}
    \begin{subfigure}[t]{\textwidth}
        \centering
        \includegraphics[trim=2cm 0cm 2cm 0cm,clip=True,scale=0.45]{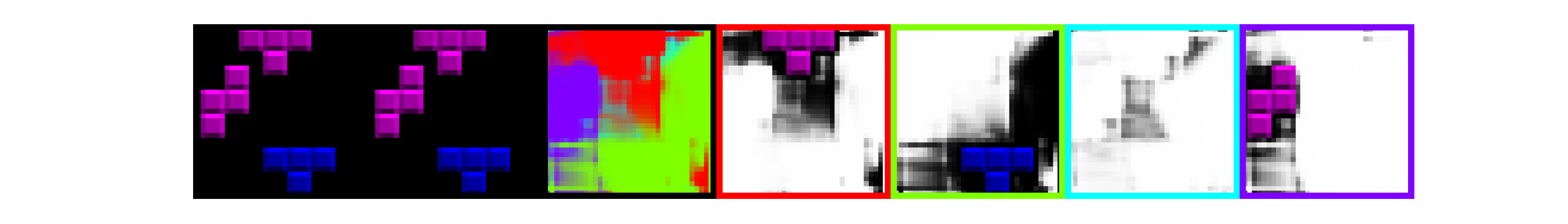}
    \end{subfigure}
    \begin{subfigure}[t]{\textwidth}
        \centering
        \includegraphics[trim=2cm 0cm 2cm 0cm,clip=True,scale=0.45]{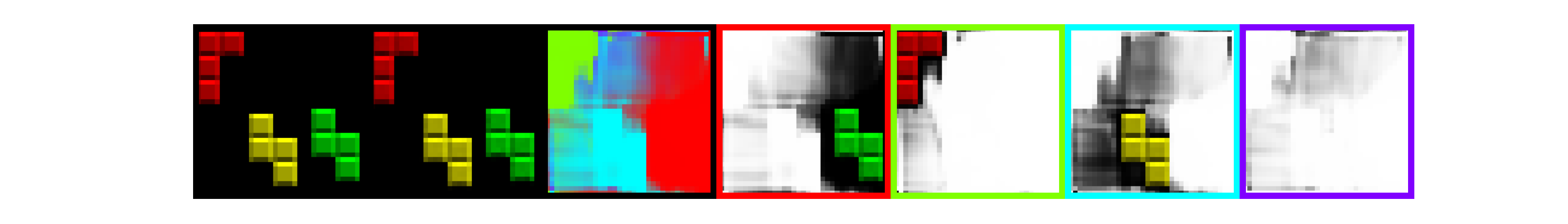}
    \end{subfigure}
    \begin{subfigure}[t]{\textwidth}
        \centering
        \includegraphics[trim=2cm 0cm 2cm 0cm,clip=True,scale=0.45]{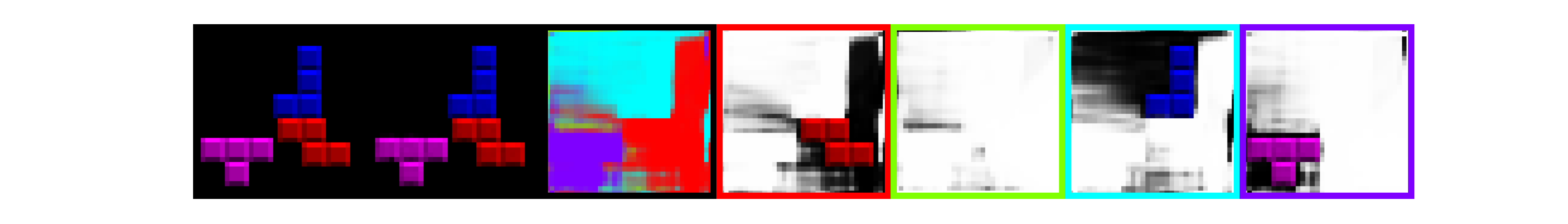}
    \end{subfigure}
    \begin{subfigure}[t]{\textwidth}
        \centering
        \includegraphics[trim=2cm 0cm 2cm 0cm,clip=True,scale=0.45]{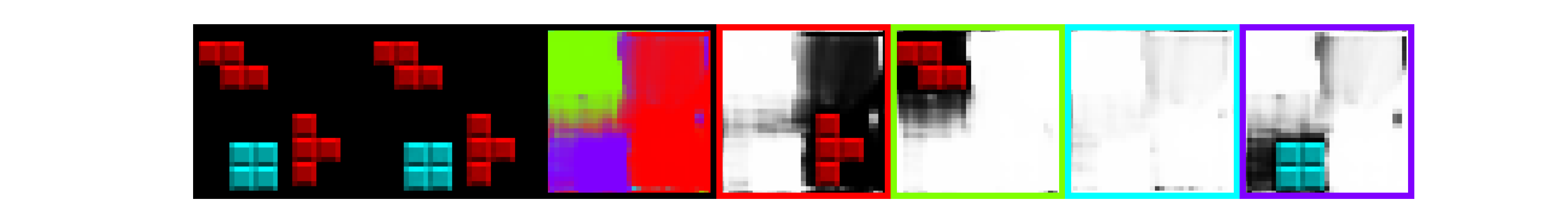}
    \end{subfigure}
    \caption{Additional Tetrominoes scene decompositions with $K=4,L=3$ and $I=3$. First column is the ground truth image, second column is the reconstruction, third column is the mask, and the remaining columns are the masked components.  The Gaussian image likelihood has only a weak inductive bias to encourage the background to be assigned to a single component; the model often learns to split the simple black background across all components. \label{fig:app:tetris_decompositions}}
\end{figure}

We also show extra qualitative scene decompositions for CLEVR6 (Figure~\ref{fig:app:clevr6_decompositions}), Multi-dSprites (Figure~\ref{fig:app:multi_dSprites_decompositions}), and Tetrominoes (Figure~\ref{fig:app:tetris_decompositions}).
Each decomposition is sampled from the final posterior after refinement.

\subsection{Disentanglement}
\label{sec:app:disentanglement}
\begin{figure*}
    \centering
    \includegraphics[trim=0cm 0cm 0cm 2cm,clip=True,scale=0.84]{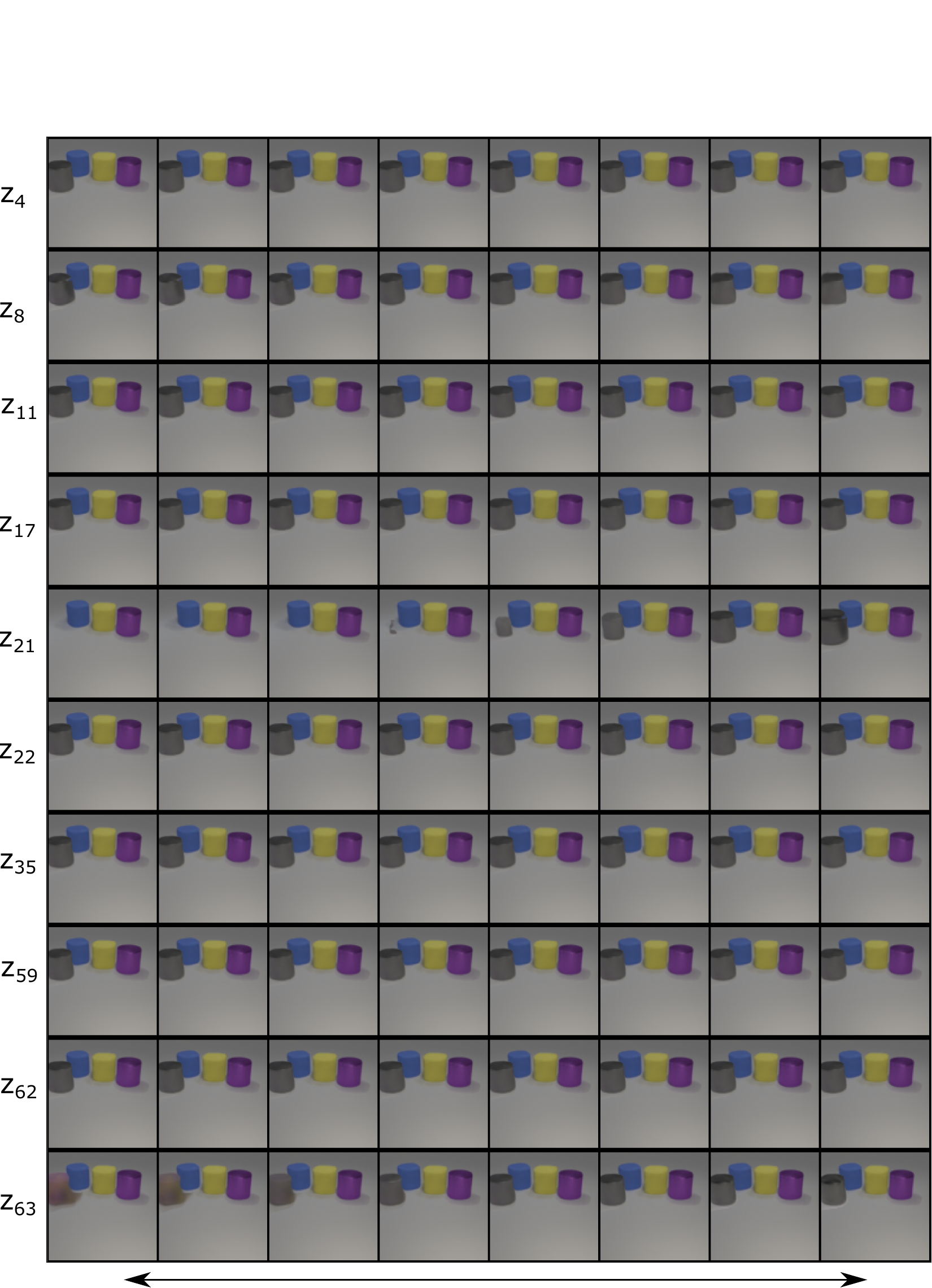}
    \caption{Varying random latent dimensions for CLEVR6. We randomly select latent dimensions from a single object component to vary. Most of the latent dimensions are deactivated and do not change the image.}
    \label{fig:app:disentangle}
\end{figure*}
\begin{figure*}
    \centering
    \includegraphics[trim=0cm 0cm 0cm 0cm,clip=True,scale=0.9]{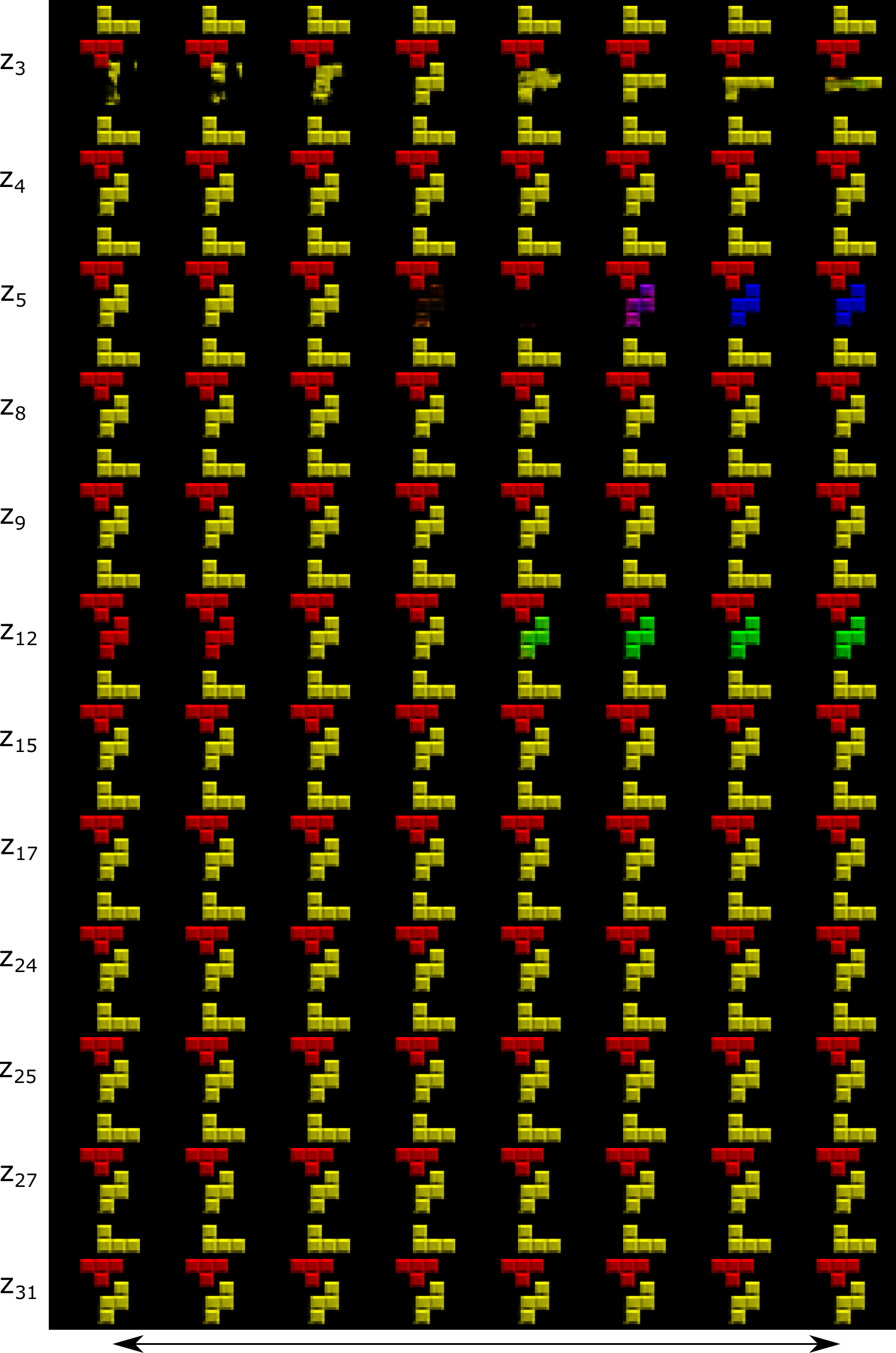}
    \caption{Varying random latent dimensions for Tetrominoes. We randomly select latent dimensions from a single object component to vary. Most of the latent dimensions are deactivated and do not change the image. The model shown here was trained \emph{without} GECO.}
    \label{fig:app:disentangle-tetris}
\end{figure*}
\begin{figure*}
    \centering
    \includegraphics[trim=0cm 0cm 0cm 0cm,clip=True,scale=0.9]{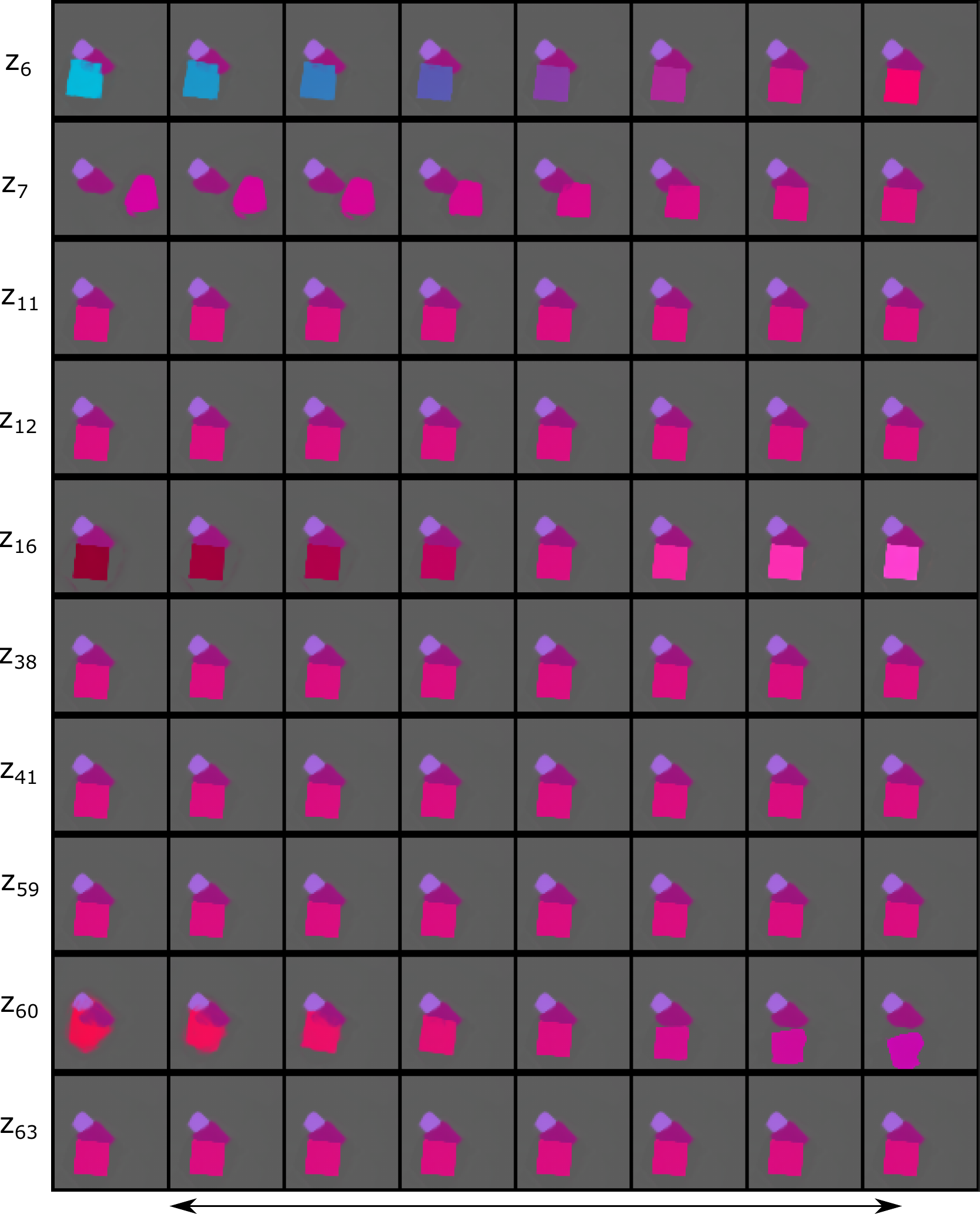}
    \caption{Varying random latent dimensions for Multi-dSprites. We randomly select latent dimensions from a single object component to vary. Most of the latent dimensions are deactivated and do not change the image.}
    \label{fig:app:disentangle-sprites}
\end{figure*}
\textbf{Visualizations} We include additional visualizations of the disentanglement learned by our model.
We randomly select one object component and multiple latent dimensions to vary. 
To determine reasonable ranges to linearly interpolate between for each latent dimension, we computed the maximum and minimum values per latent dimension across 100 images.
Figure~\ref{fig:app:disentangle} shows one model trained on CLEVR6 and Figure~\ref{fig:app:disentangle-sprites} shows one trained on Multi-dSprites.
In Figure~\ref{fig:app:disentangle-tetris}, we visualize one of the ablated Tetrominoes models which did not use GECO and successfully learned to decompose and reconstruct the images, achieving a low final KL.

We emphasize that we did not tune the ELBO to emphasize disentanglement over reconstruction quality in this paper. 
If desired, EfficientMORL can be trained to achieve a more highly disentangled latent representation, for example by modulating the KL term with a $\beta$ hyperparameter.

\textbf{DCI metric} We used the \texttt{disentanglement\_lib}~\citep{locatello2019challenging} to compute DCI scores~\citep{eastwood2018framework}.
DCI measures three quantities, \emph{disentanglement}, \emph{completeness}, and \emph{informativeness} and involves training a regressor (continuous factors) or classifier (discrete factors) to predict the ground truth factors given the extracted representation from the data. 
We repeat the DCI scores here in Table~\ref{tab:dci}.

We take the following steps to compute DCI for the considered multi-object setting for EfficientMORL and Slot Attention on CLEVR6.
For each test image, we extracted $K \times D$ latent codes. For EfficientMORL, these are the means of the final posterior distribution.
For Slot Attention, it is simply the output slots.
We also decode the $K$ masks.
Given the ground truth object segmentation, we compute the linear assignment using IOU as the cost function to find the best matching of ground truth object masks to the $K$ predicted object masks.

Finally, we concatenate all pairs of ground truth latent factors and predicted latent codes for all images into a dataset using a random 50/50 train and test split. 
We use the \texttt{disentanglement\_lib} to compute the DCI scores, which uses gradient boosted trees. A separate predictor is trained per factor. 
For CLEVR6, the factors are $\{$x, y, z, rotation, size, material, shape, color$\}$ where the first four are continuous values and the last four are discrete values.

\begin{table}[t]
    \centering
    \caption{DCI on CLEVR6 (mean $\pm$ std dev across 5 runs). Higher is better.}
    \begin{tabular}{lcccc}
        \toprule
         & Disentanglement & Completeness & Informativeness \\
        \hline
        Slot Attention & $0.46 \pm 0.01$ & $0.38 \pm 0.02$ & $0.34 \pm 0.01$ \\
        EfficientMORL & $\mathbf{0.63} \pm 0.04$ & $\mathbf{0.63} \pm 0.06$  & $\mathbf{0.46} \pm 0.01$ \\
        \hline
    \end{tabular}
    \label{tab:dci}
\end{table}

\subsection{Efficiency} 
\begin{figure}[t!]
\centering
    \begin{subfigure}[t]{0.33\textwidth}
        \includegraphics[scale=0.36]{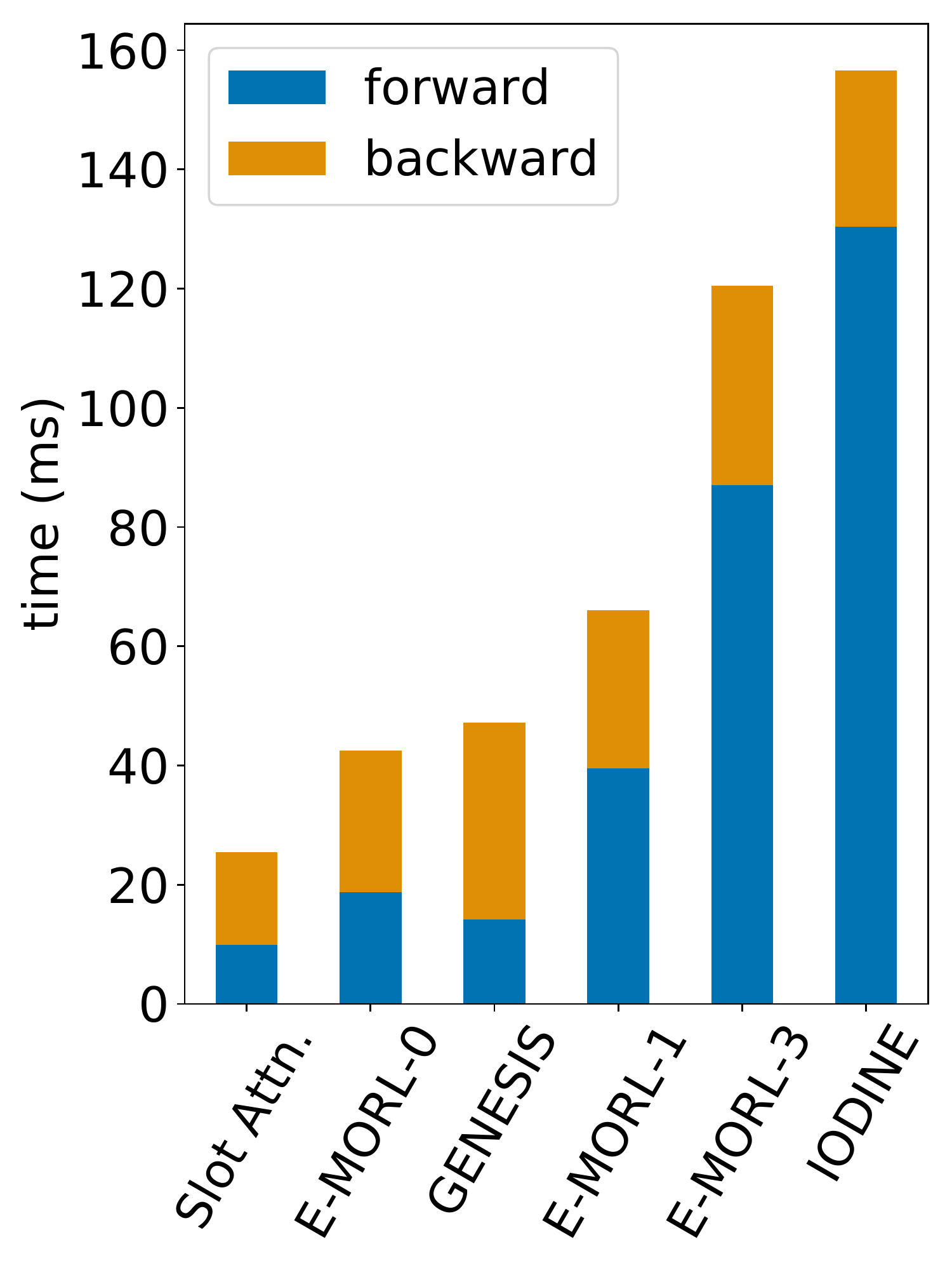}
        \caption{$64 \times 64$ images}
    \end{subfigure}
    \begin{subfigure}[t]{0.33\textwidth}
        \includegraphics[scale=0.36]{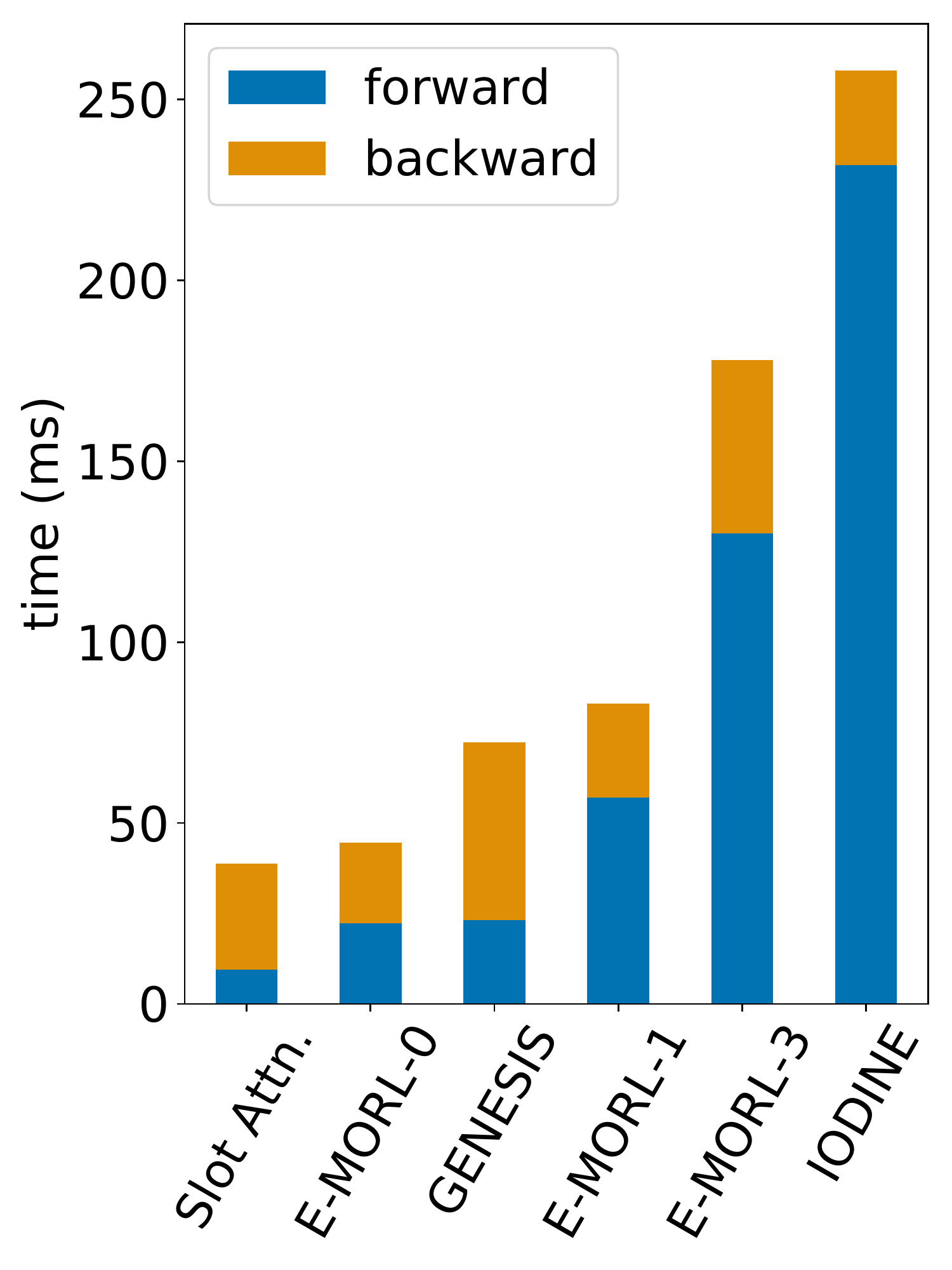}
        \caption{$96 \times 96$ images}
    \end{subfigure}
    \begin{subfigure}[t]{0.33\textwidth}
        \includegraphics[scale=0.36]{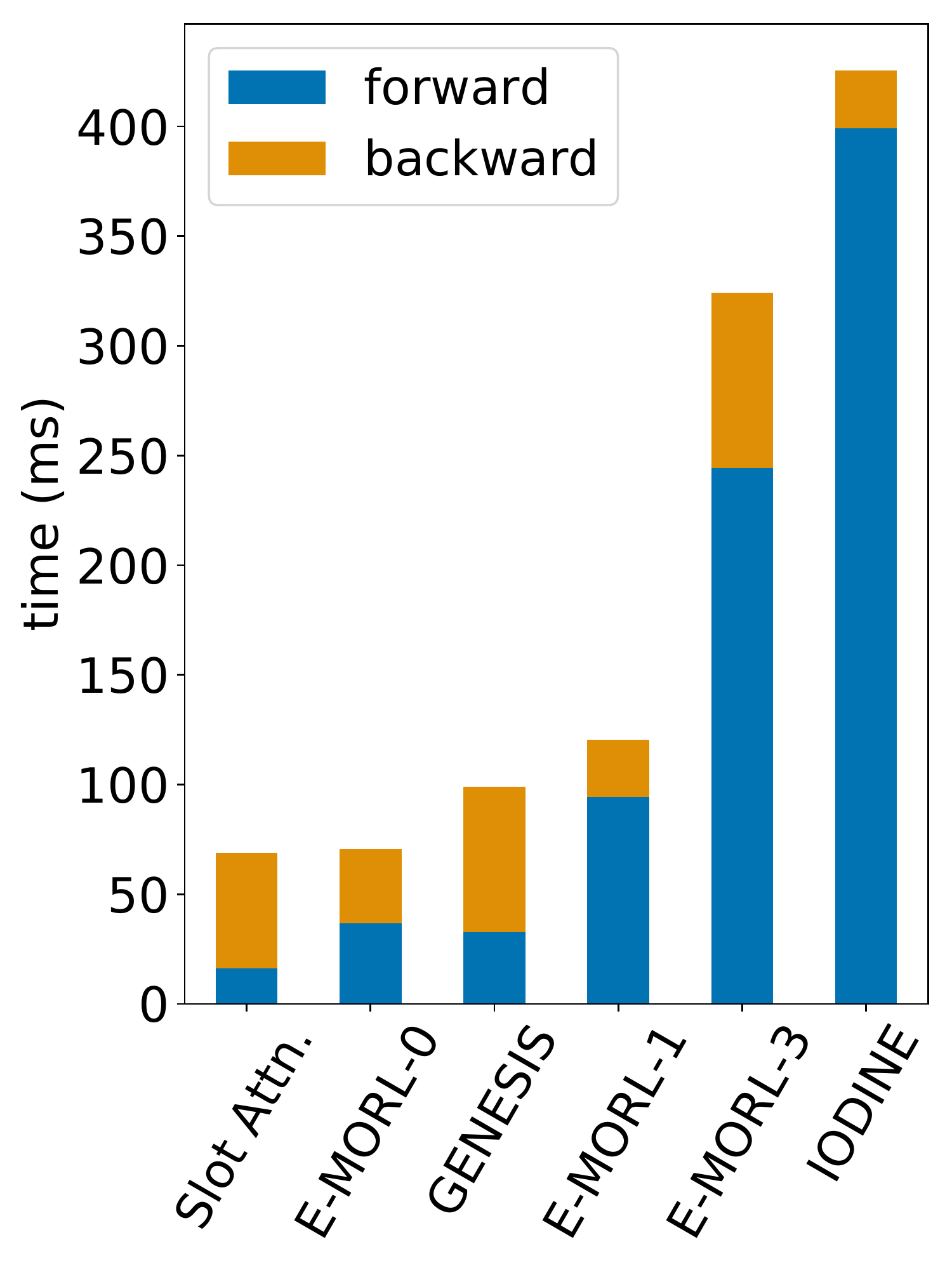}
        \caption{$128 \times 128$ images}
    \end{subfigure}
    \caption{Forward and backward pass timings}
    \label{fig:app:efficiency}
\end{figure}
\paragraph{Trainable parameters} IODINE has approximately $2.75M$ parameters and EfficientMORL has approximately $666K$ parameters, a 75\% decrease.

\paragraph{Timing metric} We computed the forward and backward pass timing for the considered models on the same hardware, using 1 RTX 2080 Ti GPU with mini-batch size of $4$. 
We computed a running average across $10K$ passes after a burn in period of about five minutes.
We show results for multiple images dimensions in Figure~\ref{fig:app:efficiency}.

\paragraph{Refinement curriculum} As stated earlier, we used two approaches in our experiments: fixing $I=3$ throughout training (Tetrominoes), or decreasing $I$ from three to one after $100K$ training steps (CLEVR6 and Multi-dSprites).
When applying EfficientMORL to more challenging datasets for which convergence may take a long time, we recommend trying to simply decrement $I$ by one following a reasonable schedule (e.g., every $100K$ steps).
We directly decreased $I$ from three to one since EfficientMORL converges relatively quickly on all three of the Multi-object Benchmark datasets.

\section{Implementation}
\label{sec:app:implementation}

Code and data are available at \url{https://github.com/pemami4911/EfficientMORL}

\subsection{Architecture details and hyperparameters}

\paragraph{DualGRU} The HVAE encoder has two separate GRUs to predict the mean $\bm{\mu}$ and variance $\bm{\sigma}^2$ for each bottom-up inference layer's posterior.
This reduces the number of parameters in the encoder which regularizes the model and eases learning; see Section~\ref{sec:app:ablations} for an ablation study on its use.

To parallelize the computation of the two GRUs we fuse their weights to create a single GRU---the \emph{DualGRU}---that has sparse block-diagonal weight matrices.
Specifically, the $D$-dim output of the scaled dot-product attention set attention is concatenated with itself $[\bm{\Theta};\bm{\Theta}] \in \mathbb{R}^{K \times 2D}$ and then passed as input to the DualGRU.
As an example, the DualGRU multiplies the input with the following:
\begin{equation}
    \mathbf{W}_{ih} = \left[ \begin{array}{cc} \mathbf{W}_{ih}^{1} & \mathbf{0} \\ \mathbf{0} &  \mathbf{W}_{ih}^{2}\end{array} \right],
\end{equation}
where $\mathbf{W}_{ih}^{1} \in \mathbb{R}^{2D \times D}$ and $\mathbf{W}_{ih}^{2} \in \mathbb{R}^{2D \times D}$. 
The shape of $\mathbf{W}_{ih}$ is $4D \times 2D$; internally, the DualGRU splits the $K \times 4D$ output of its application to the $K \times 2D$ input into $4$ $D$-dim activations for the two reset and update gates.
All other weight matrices $\mathbf{W}_{hh}$, $\mathbf{W}_{in}$, $\mathbf{W}_{hn}$ are similarly defined as block matrices.

In practice, we predict the variance for each Gaussian using the \texttt{softplus} (SP) operation. 
In detail, let $\mathbf{o}$ be a $K \times D$ output of a linear layer. 
Then,
\begin{align}
    \mathbf{o}^{'} &= \texttt{min}( \mathbf{o}, 80*\mathbf{1}) \\
    \bm{\sigma}^2 &= \log (1 + \texttt{exp}( \mathbf{o}' )+ 1e-5 )
\end{align}

\paragraph{Image likelihoods} Let $P = 3HW$ be the number of pixels, $x_p$ be the random variable for pixel $p$'s value, $y_{k,p} \in \mathbb{R}^3$ be the RGB value for the $p$\textsuperscript{th} pixel from the $k \in K$\textsuperscript{th} component, and $\pi_{k,p} \in [0,1]$ be the corresponding assignment. 
The Gaussian image likelihood has $x_p$ distributed according to a Gaussian with mean given by the sum over $K$ of $\pi_{k,p}y_{k,p}$:
\begin{equation}
    p_\theta(x \mid \mathbf{z}^L) = \prod_{p=1}^P \mathcal{N}(x_p; \sum_{k=1}^K \pi_{k,p} y_{k,p}, \sigma^2) \label{eq:gaussian}
\end{equation}
with variance $\sigma^2$ fixed for all $p$. 
We use this formulation for Tetrominoes $(\sigma = 0.3)$ and Multi-dSprites $(\sigma = 0.1)$.

The other image likelihood we consider is a \emph{mixture} of pixel-wise Gaussians~\cite{burgess2019monet,pmlr-v97-greff19a,engelcke2019genesis}:
\begin{equation}
     p_\theta(x \mid \mathbf{z}^L) = \prod_{p=1}^P \sum_{k=1}^K \pi_{k,p} \mathcal{N}(x_p; y_{k,p}, \sigma^2)  \label{eq:MoG}
\end{equation}
with the variance $\sigma^2$ fixed for all $k$ and all $p$. As demonstrated in our experiments, this likelihood attempts to segment the background into a single component, which is (perhaps unintuitively) more challenging for the Tetrominoes and Multi-dSprites environments.
Therefore, we only use it on CLEVR6 ($\sigma = 0.1$).
As evidenced by IODINE, with sufficient hyperparameter tuning it seems possible to use the mixture likelihood for all three environments; however, we wished to minimize hyperparameter optimization to avoid overfitting to these specific environments.
We attempted to use $\sigma = 0.1$ for all environments (like IODINE) but found that slightly increasing $\sigma$ to $0.3$ on Tetrominoes helped the model converge more rapidly.

A third image likelihood that has been previously used~\cite{VANSTEENKISTE2020309} is a \emph{layered} image representation (i.e., alpha compositing).
We did not compare against this one for the following reasons.
The standard way to implement layered rendering requires imposing an \emph{ordering} on the layers. 
This breaks the symmetry of the latent components, which is undesirable. 
Alternatively, one could treat the ordering as a discrete random variable that must be inferred, but 1) this increases the difficulty of the already-challenging inference problem, and 2) this can lead to the generation of implausible scenes due to uncertainty about the true (unknown) depth ordering.

\textbf{Bottom-up inference network} Following Slot Attention and IODINE, we use latent dimensions of $D = 64$ for CLEVR6 and Multi-dSprites and $D = 32$ for Tetrominoes.
Before the first layer, the provided image is embedded using a simple CNN~\cite{locatello2020object}:
\begin{center}
\begin{tabular}{lcc}
    \multicolumn{3}{c}{\textbf{Image encoder}}\\
    \toprule
    Type & Size/\textit{Ch.} & Act. Func. \\ \midrule
    $H \times W$ image & \textit{3}  & \\
    Conv $5\times5$ & \textit{64}  & ReLU \\
    Conv $5\times5$ & \textit{64}  & ReLU \\
    Conv $5\times5$ & \textit{64}  & ReLU \\
    Conv $5\times5$ & \textit{64}  & ReLU \\
    \bottomrule
\end{tabular}
\end{center}
where each 2D convolution uses stride of $1$ and padding of $2$. 
A learned \textbf{positional encoding} of shape $H \times W \times 4$ is projected to match the channel dimension $64$ with a linear layer and then added to the image embedding. 
Like Slot Attention, the four dimensions of the encoding captures the cardinal directions of left, right, top, and bottom respectively. 
This enables extracting spatially-aware image features while processing the image in a permutation invariant manner.
The positionally-aware image embedding is flattened along the spatial dimensions to $HW \times 64$ and then processed sequentially with a LayerNorm, $64$-dimensional linear layer, a ReLU activation, and another $64$-dimensional linear layer.
The result is used as the key and value for scaled dot-product set attention.

Each of the $L$ bottom-up stochastic layers share these parameters:
\begin{center}
    \begin{tabular}{lcc}
        \multicolumn{3}{c}{\textbf{\emph{l}\textsuperscript{th} layer}}\\
        \toprule
        Type & Size/\textit{Ch.} & Act. Func. \\ \midrule
        key $(k)$ & $64 \rightarrow D$ & None \\
        value $(v)$ & $64 \rightarrow D$ & None \\
        query $(q)$ & $D \rightarrow D$ & None \\
        DualGRU & $2D \rightarrow 2D$ & Sigmoid/Tanh \\
        MLP $(\bm{\mu})$ & $D \rightarrow 2D \rightarrow D$ & ReLU \\
        MLP $(\bm{\sigma})$ & $D \rightarrow 2D \rightarrow D$ & ReLU, softplus \\
        \bottomrule
    \end{tabular}
    \end{center}
Each of the above operations are applied symmetrically to each of the $K$ elements of $\mathbf{z}^{l}$.
The two $D$-dimensional outputs of the DualGRU are passed through separate trainable LayerNorm layers before the MLPs. 
The posterior parameters $\mu^0$ and $\sigma^0$ provided to the DualGRU when $l=1$ are trainable and are initialized to $\mathbf{0}$ and $\mathbf{1}$ respectively.

\textbf{Hierarchical prior network}
Each of the $L-1$ data-dependent layers in the prior shares parameters.
These predict the mean and variance of a Gaussian conditional on a $D$-dimensional random sample $\mathbf{z}$.
We use ELU activations~\cite{clevert2015fast} here but use ReLU activations in the inference network in the parts of the architecture similar to Slot Attention.
\begin{center}
    \begin{tabular}{lcc}
        \multicolumn{3}{c}{\textbf{\emph{l}\textsuperscript{th} layer}}\\
        \toprule
        Type & Size/\textit{Ch.} & Act. Func. \\ \midrule
        MLP & $D \rightarrow 128$ & ELU \\
        Linear $(\bm{\mu})$ & $128 \rightarrow D$ & None \\
        Linear $(\bm{\sigma})$ & $128 \rightarrow D$ & softplus \\
        \bottomrule
    \end{tabular}
    \end{center}
This computation is applied symmetrically to each of the $K$ elements of $\mathbf{z}$.

\textbf{Refinement} The output of the refinement network is $\bm{\delta \lambda} = [\bm{\delta\mu},\bm{\delta\sigma}]$ which is used to make the additive update to the posterior.  
\begin{center}
\begin{tabular}{lcc}
    \multicolumn{3}{c}{\textbf{Refinement network}}\\
    \toprule
    Type & Size/\textit{Ch.} & Act. Func. \\ \midrule
    $[\bm{\lambda}, \nabla_{\bm{\lambda}} \mathcal{L}]$ & $4D$ &  \\
    MLP & $4D \rightarrow 128 \rightarrow D$ & ELU \\ 
    GRU & $D \rightarrow D$ & Sigmoid/Tanh \\
    Linear $(\bm{\delta\mu})$ & $D \rightarrow D$ & None \\
    Linear $(\bm{\delta\sigma})$ & $D \rightarrow D$ & SP \\
    \bottomrule
\end{tabular}
\end{center}
The two vector inputs to the refinement network are first processed with trainable LayerNorm layers.
We compute the update for each of the $K$ elements of the set of posterior parameters $\bm{\lambda}$ in parallel.

\textbf{Decoder} The spatial broadcast decoder we use is similar to IODINE's except we adopt Slot Attention's positional encoding to mirror the encoding used during bottom-up inference.
Each of the $K$ elements of $\mathbf{z}$ are decoded independently in parallel.
\begin{center}
\begin{tabular}{lcc}
    \multicolumn{3}{c}{\textbf{Spatial broadcast decoder}}\\ 
    \toprule
    Type & Size/\textit{Ch.} & Act. Func.\\ \midrule
    Input: $\mathbf{z}$ & $D$ &  \\
    Broadcast & $(H+10)\times (W+10) \times D$ & \\
    Pos. Enc. & $4 \rightarrow D$ & Linear \\
    Conv $3\times3$ & \textit{64} & ELU \\
    Conv $3\times3$ & \textit{64} & ELU \\
    Conv $3\times3$ & \textit{64} & ELU \\
    Conv $3\times3$ & \textit{64} & ELU \\
    Conv $3\times3$ & \textit{4} & None \\
    \bottomrule
\end{tabular}
\end{center}
Each convolutional layer uses a stride of $1$ and no padding.
The channel dimension of the positional encoding is again projected to $D$ before being added to the broadcasted $\mathbf{z}$. 
For Tetrominoes, following Slot Attention we use a lighter decoder that has just three convolutional layers with $5\times5$ kernels, stride $1$ and padding $1$, and channel dimension of $32$. 

The deconvolutional decoder we use for E-MORL-X-S in Figure~\ref{fig:efficiency} is identical to Slot Attention's (see Section E.3, Table 6 in their paper).

The four dimensional output of the decoder is split into $K$ masks and RGB images.
The masks are normalized over $K$ by a softmax and the RGB outputs are passed through a sigmoid to squash values between zero and one.
Images under both considered likelihood models are reconstructed by summing the normalized masks multiplied by the RGB components across $K$.

\subsection{GECO~\cite{rezende2018taming}}
Where needed, we mitigate posterior collapse while simultaneously balancing the reconstruction and KL with GECO.
This reformulates the objective as a minimization of the KL subject to a constraint on the reconstruction error.
The training loss $\mathcal{L}$ (Equation~\ref{eq:finalloss}) is modified for GECO:
\begin{align*}
    \mathcal{L}^{(L,0)} &= \infdiv{q_\phi(\mathbf{z}^{1:L} \mid x)}{p_\theta(\mathbf{z}^{1:L})} - \zeta \bigl( C + \mathcal{L}_{\text{NLL}} \bigr) \\
    \mathcal{L}^{(L,i)} &= \infdiv{q(\mathbf{z} ; \bm{\lambda}^{(L,i)})}{p(\mathbf{z}^L)} - \zeta \bigl( C + \mathcal{L}_{\text{NLL}}^{(L,i)} \bigr). \\
\end{align*}
Here, $\zeta$ is a Lagrange parameter that penalizes the model when the reconstruction error is higher than a manually-specified threshold $C$.
Depending on the hierarchical prior variant we replace $p(\mathbf{z}^L)$, e.g., with $p_\theta(\mathbf{z}^{1} | \mathbf{z}^{2})$ for reversed prior++.
We use the recommended exponential moving average $C_{\text{EMA}}$ with parameter $\alpha = 0.99$ to keep track of the difference between the reconstruction error of the mini-batch and $C$.
The Lagrange parameter is updated at every step with $\zeta' = \zeta - \text{1e-6 } C_{\text{EMA}}$.
For numerical stability, we use $\texttt{softplus}(\zeta)$ when computing the GECO update and constrain $\zeta \geq 0.55$ so that $\texttt{softplus}(\zeta)$ is always greater than or equal to one. For CLEVR6, we used $C = -61000$ and for Tetrominoes we used $C = -4500$.
GECO was not needed on Multi-dSprites.

\end{document}